\pdfoutput=1

\documentclass[11pt]{article}

\usepackage[]{emnlp2021}

\usepackage{times}
\usepackage{latexsym}

\usepackage[T1]{fontenc}

\usepackage[utf8]{inputenc}

\usepackage{microtype}

\usepackage[whole]{bxcjkjatype}

\usepackage{subcaption}

\usepackage{multirow}
\usepackage{booktabs}

\usepackage{opensans}

\usepackage{amsmath,amssymb,mathtools}
\usepackage{tikz}
\usepackage{graphicx}

\newcommand{\mathboldface}[1]{\boldsymbol{#1}}
\newcommand{\bm}[1]{\mathboldface{#1}}

\makeatletter
\newcommand*{\encircled}[1]{\relax\ifmmode\mathpalette\@encircled@math{#1}\else\@encircled{#1}\fi}
\newcommand*{\@encircled@math}[2]{\@encircled{$\m@th#1#2$}}
\newcommand*{\@encircled}[1]{%
  \tikz[baseline,anchor=base]{\node[draw,circle,outer sep=0pt,inner sep=.2ex] {#1};}}
\makeatother

\usepackage{scalerel}
\newcommand\mathbox[1]{\mathord{\ThisStyle{%
  \fboxsep3\LMpt\relax\kern1\LMpt\fbox{$\SavedStyle#1$}\kern1\LMpt}}}

\newcommand{\emphmix}[2]{%
  \tikz[baseline]
  \node[fill=blue!15, rounded corners=3pt, anchor=base]
  (#1)
  {\ensuremath{#2}};}
\newcommand{\emphpreserve}[2]{%
  \tikz[baseline]
  \node[fill=lightgray!45, rounded corners=3pt, anchor=base]
  (#1)
  {\ensuremath{#2}};}

\newcommand{\norm}[1]{\lVert #1 \rVert}

\title{Incorporating Residual and Normalization Layers \\ into Analysis of Masked Language Models} %

\author{
Goro\,Kobayashi$^{1}$\hspace{1em}
Tatsuki\,Kuribayashi$^{1,2}$\hspace{1em}
Sho\,Yokoi$^{1,3}$\hspace{1em}
Kentaro\,Inui$^{1,3}$\\[2pt]
$^{1}$ Tohoku University\hspace{1em}
$^{2}$ Langsmith Inc.\hspace{1em}
$^{3}$ RIKEN\hspace{1em}
\\
\texttt{goro.koba@dc.tohoku.ac.jp} \\
\texttt{\{kuribayashi, yokoi, inui\}@tohoku.ac.jp} \\
}

\begin{document}
\maketitle

\begin{abstract}
Transformer architecture has become ubiquitous in the natural language processing field.
To interpret the Transformer-based models, their attention patterns have been extensively analyzed.
However, the Transformer architecture is not only composed of the multi-head attention; other components can also contribute to Transformers' progressive performance. 
In this study, we extended the scope of the analysis of Transformers from solely the attention patterns to the whole attention block, i.e., multi-head attention, residual connection, and layer normalization.
Our analysis of Transformer-based masked language models shows that the token-to-token interaction performed via attention has less impact on the intermediate representations than previously assumed.
These results provide new intuitive explanations of existing reports; for example, discarding the learned attention patterns tends not to adversely affect the performance.
The codes of our experiments are publicly available.\footnote{
    \url{https://github.com/gorokoba560/norm-analysis-of-transformer}
}
\end{abstract}

\section{Introduction}
\label{sec:intro}
Transformer architecture~\citep{vaswani17} has advanced the state of the art in a wide range of natural language processing (NLP) tasks~\cite{devlin2018bert,liu19,lan2020albert}.
Along with this, Transformers have become a major subject of research from the viewpoints of engineering~\cite{rogers-etal-2020-primer} and scientific studies~\cite{Merkx2020ComparingData,Manning30046}.

\begin{figure}[t]
\centering
    \begin{minipage}[t]{.47\hsize}
        \centering
        \includegraphics[height=3.9cm]{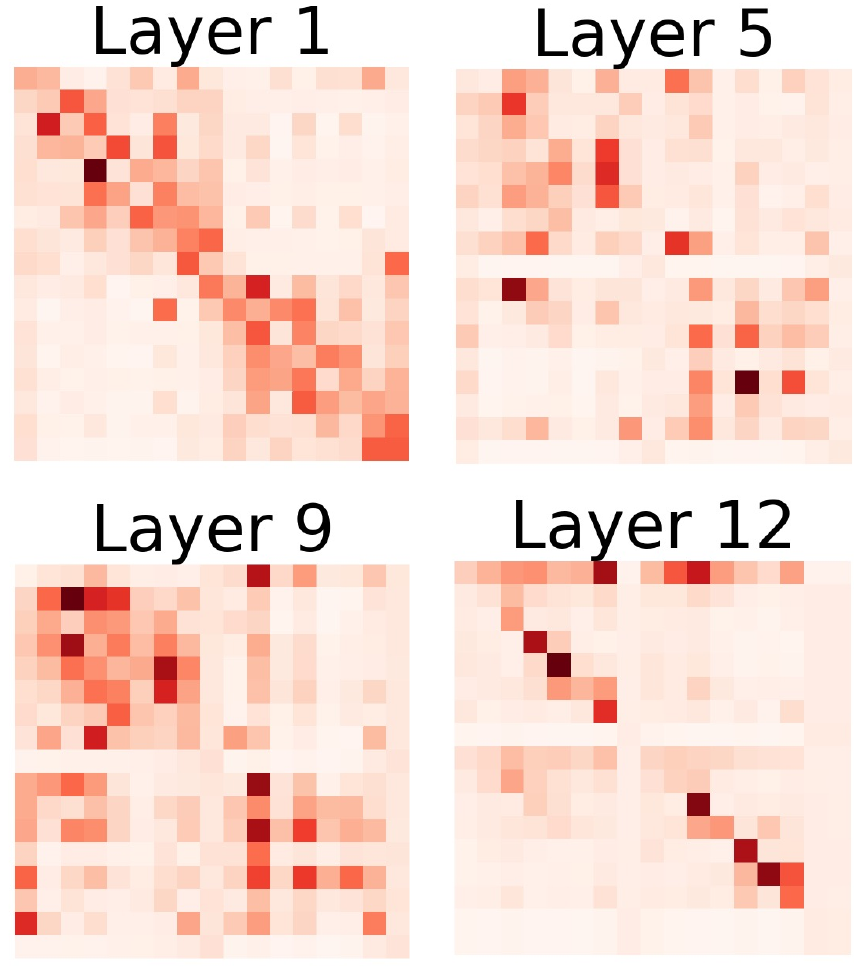}
        \subcaption{
        Existing analysis focusing only on the multi-head attention~\cite{kobayashi-etal-2020-attention}.
        }
        \label{fig:kobayashi_viz}
    \end{minipage}
    \;\;
    \begin{minipage}[t]{.47\hsize}
        \centering
        \includegraphics[height=3.9cm]{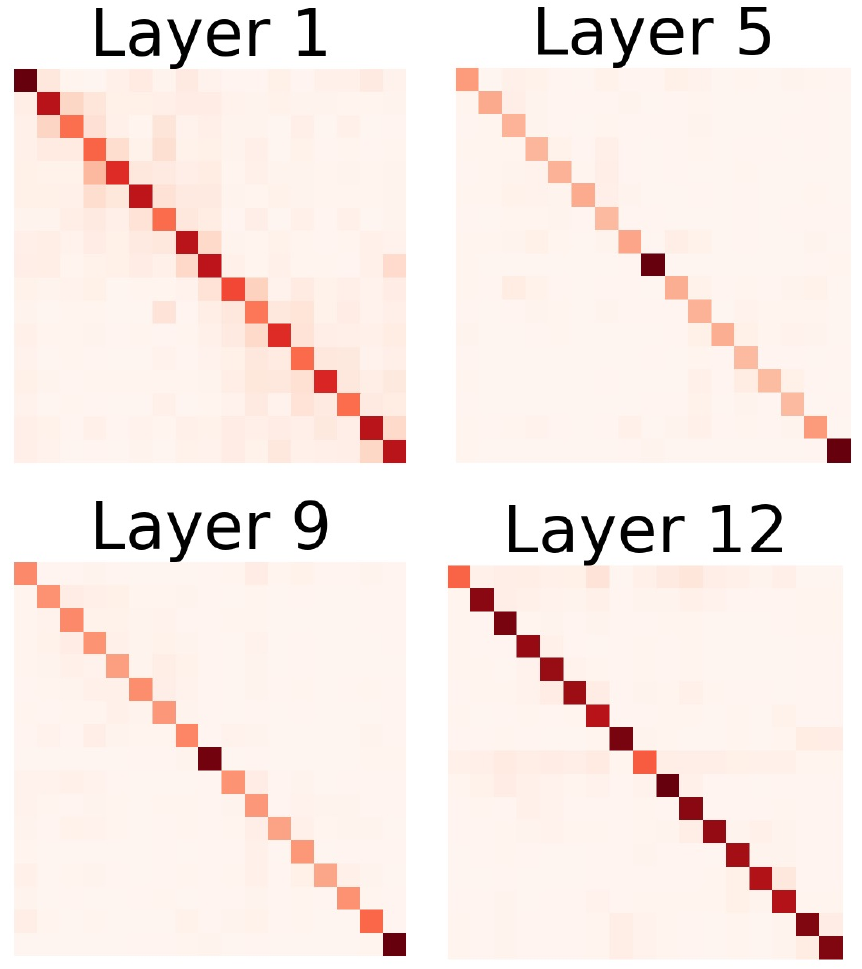}
        \subcaption{
        Proposed method incorporating the whole attention block (i.e., multi-head attention, residual connection, and layer normalization) into the analysis.
        }
        \label{fig:proposed_viz}
    \end{minipage}
    \caption{
    Visualizations of the token-by-token interactions in each layer when a sentence pair is fed into the pre-trained BERT-base.
    The diagonal elements correspond to the effect of preserving the original input information.
    The contrast between Figures~\ref{fig:kobayashi_viz} and~\ref{fig:proposed_viz} demonstrates that the contextual information contributed less to the computation of the output representations than previously expected.
    }
    \label{fig:visualize_comparison}
\end{figure}

In particular, the multi-head attention, a core component of Transformers, has been extensively analyzed~\cite{clark19,kovaleva19,coenen19,lin19,marecek19balustrades,Htut2019,raganato2018,tang_transformer_attention_analysis}. 
While these analyses suggest that the multi-head attention contributes to capturing linguistic information such as semantic and syntactic relations, 
some reports question the importance of attention.
For example, several studies in fields ranging from NLP~\cite{michel19,kovaleva19} to neuroscience~\cite{toneva-nips2019-interpreting} empirically found %
that discarding learned attention patterns from Transformers retains or even improves their performance in downstream tasks and the ability to simulate human brain activity.
These observations imply that Transformers do not heavily rely on the multi-head attention alone, and the other components contribute to their progressive performance.

In this study, we broaden the scope of the analysis from the multi-head attention to the whole attention block, i.e., the multi-head attention, residual connection, and layer normalization. %
Our analysis of  the Transformer-based masked language models~\cite{devlin2018bert,liu19}
revealed that the newly incorporated components have a larger impact than expected in previous studies~\cite{abnar-zuidema-2020-quantifying,kobayashi-etal-2020-attention} (Figure~\ref{fig:visualize_comparison}).

More concretely, we introduce an exact decomposition of the operations in the whole attention block exploiting the norm-based analysis~\cite{kobayashi-etal-2020-attention}. %
Our analysis quantifies the impact of the two contrasting effects of the attention block : (i) ``mixing'' the input representations via attention and (ii) ``preserving'' the original input mainly via residual connection (Section~\ref{sec:proposal}). %
Our analysis reveals that the preserving effect is more dominant in each attention block than previously estimated~\cite{abnar-zuidema-2020-quantifying,kobayashi-etal-2020-attention}.
The results also %
reveal the detailed mechanism of each component in the attention block.
The residual connections pass through much larger vectors than the vectors produced by the multi-head attention.
The layer normalization also reduces the effect of the operation via attention.

Our finding of the relatively small impact of the multi-head attention provides new intuitive interpretations for some existing reports, for example, discarding the learned attention patterns did not adversely affect their performance.
Our analysis also provides a new intuitive perspective on the behaviors of Transformer-based masked language models.
For example, BERT~\cite{devlin2018bert} highlights low-frequency (informative) words in encoding texts, which is consistent with the existing %
methods for effectively computing text representations~\cite{luhn-1958-summarization,arora-2017-sif}.

The contributions of this study are as follows:
\begin{itemize}
    \item  We expanded the scope of Transformers analysis from the multi-head attention to the attention block (i.e., multi-head attention, residual connection, and layer normalization).
    \item Our analysis revealed that the operation via residual connection and layer normalization contributes more to the internal representations than expected in previous studies~\cite{abnar-zuidema-2020-quantifying,kobayashi-etal-2020-attention}. %
    \item We detailed the functioning of BERT: %
    (i) BERT tends to mix a relatively large amount of contextual information into \texttt{[MASK]} in the middle and later layers;
    and
    (ii) the contribution of contextual information in the attention block is related to word frequency.
\end{itemize}

\section{Background: Transformer architecture\!\!}
\label{sec:background}

\begin{figure}[t]
    \centering
    \includegraphics[height=6cm]{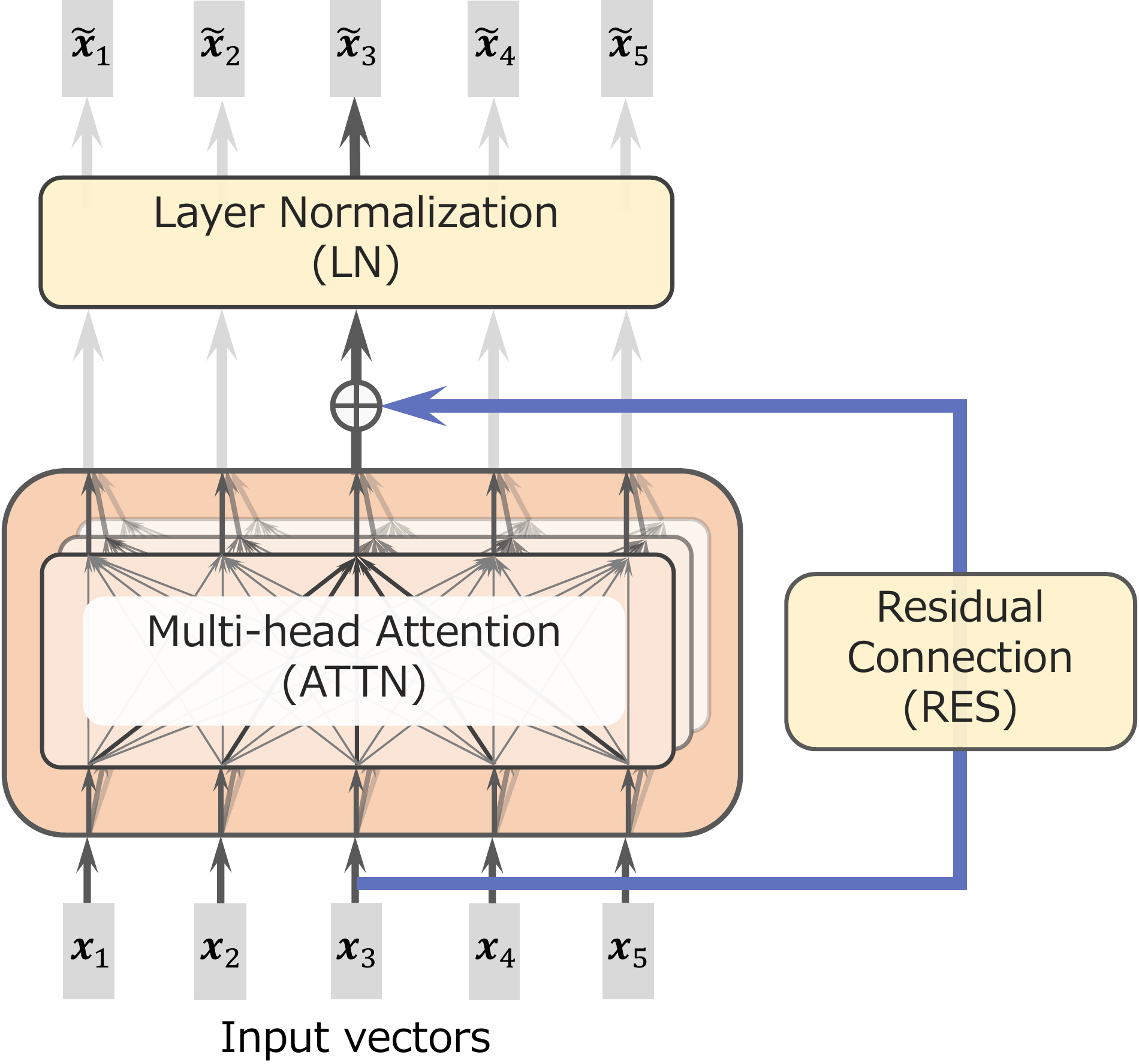}
    \caption{Visualization of the attention block, consisting of multi-head attention, residual connection, and layer normalization, in the Transformer layer.}
    \label{fig:analysis_range}
\end{figure}

The Transformer architecture consists of a stack of layers.
Each layer has an attention block, which is responsible for capturing the interactions between input tokens.
The attention block can be further divided into the three components: \textbf{multi-head attention (ATTN)}, \textbf{residual connection (RES)}, and \textbf{layer normalization (LN)} (Figure~\ref{fig:analysis_range}).
This block can be written as the following composite function:
\begin{equation}
\begin{split}
    \widetilde{\bm x}_i
    &=
    \mathrm{LN}\Bigl(\mathrm{ATTN}(\bm x_i, \bm X) + \bm x_i\Bigl)
    \text{,}
\end{split}
\end{equation}
where
$\bm x_i\in\mathbb R^d$ is the $i$-th input representation,
$\bm X := [\bm x_1,\ldots,\bm x_n] \in \mathbb{R}^{n\times d}$ is the sequence of input representations, %
and
$\widetilde{\bm x}_i \in \mathbb R^d$
is the output representation corresponding to $\bm x_i$.
Boldface letters such as $\bm x$ denote row vectors.
In the following, we review the computations in the ATTN, RES, and LN components.

\paragraph{Multi-head attention (ATTN):}
The ATTN takes the role of mixing contextual information into output representations.
Formally, given input representations $\bm X$, the $H$ head ATTN computes the output $\mathrm{ATTN}(\bm x_i, \bm X)\in\mathbb R^d$ for each input $\bm x_i$:
\begin{align}
    \mathrm{ATTN}(\bm x_i, \bm X)
    &= \sum_{h=1}^H \mathrm{ATTN}^{\encircled{h}}(\bm x_i, \bm X)
    \text{,}
    \label{eq:attn_multihead}
\end{align}
where $\mathrm{ATTN}^{\encircled{h}}(\bm x_i, \bm X) \in \mathbb R^d$
denotes the output vector from each attention head $h$. 
 $\mathrm{ATTN}^{\encircled{h}}(\bm x_i, \bm X)$ is computed by each attention head $h$ as follows: %
\begin{align}
    \begin{split}
        &\mathrm{ATTN}^{\encircled{h}}(\bm x_i, \bm X) \\
        &\;\;\;\;\;\;\;\;= \sum_{j=1}^n \alpha_{i,j}^{\encircled{h}} \left(\bm{x}_j\bm{W}_V^{\encircled{h}}+\bm{b}_V^{\encircled{h}} \right)\bm W_O^{\encircled{h}} 
        \text{,}
        \label{eq:attn_details}
    \end{split}
    \\
    &\alpha_{i,j}^{\encircled{h}} := \mathop{\operatorname{softmax}}_{\bm x_j \in \bm X}\Biggl(\frac{\bm q(\bm x_i) \bm k(\bm x_j)^\top}{\sqrt{d_h}}\Biggr)
    \text{,}
    \label{eq:attn_weight} \\
    &\bm q(\bm x) := \bm x \bm W_Q^{\encircled{h}} + \bm b_Q^{\encircled{h}} 
    \text{,} \\
    &\bm k(\bm x):= \bm x \bm W_K^{\encircled{h}} + \bm b_K^{\encircled{h}}
    \label{eq:attn_key}
    \text{,}
\end{align}
where $\bm{W}_Q^{\encircled{h}}, \bm{W}_K^{\encircled{h}}, \bm{W}_V^{\encircled{h}}\in \mathbb{R}^{d\times d_h}$, and $\bm{W}_O^{\encircled{h}} \in \mathbb{R}^{d_h\times d}$ are the weight matrices, and $\bm{b}_Q^{\encircled{h}},\bm{b}_K^{\encircled{h}}$, and $\bm{b}_V^{\encircled{h}}\in\mathbb{R}^{d_h}$ are the bias vectors.
$d_h$ denotes the dimension of each head ($64$ is usually used), and $d_h \times H = d$ holds.
Here, $Q$, $K$, and $V$ %
stand for Query, Key, and Value, respectively.
Note that in a typical attention analysis, the attention weight $\alpha^{\encircled{h}}_{i,j}$ has been assumed to represent the contribution of the input $\bm x_j$ to computing $\widetilde{\bm x_i}$.

\paragraph{Residual connection (RES):} 
In RES, the original input vector for the multi-head attention ($\bm x_i$) is added to its output vector:
\begin{align}
    \mathrm{ATTN}(\bm x_i, \bm X) \mapsto \mathrm{ATTN}(\bm x_i, \bm X) + \bm x_i
    \text{.}
    \label{eq:attn+res}
\end{align}

\paragraph{Layer normalization (LN):}
LN first normalizes the input vector and then applies a transformation with learnable parameters $\bm{\gamma} \in \mathbb R^d$ and $\bm{\beta}\in\mathbb{R}^d$:
\begin{align}
    \mathrm{LN}(\bm y) &= \frac{\bm y - m(\bm y)}{s(\bm y)}\odot\bm\gamma+\bm\beta\in\mathbb R^d
    \text{,}
    \label{eq:layernorm}
\end{align}
where $m(\bm y)\in \mathbb R$ and $s(\bm y) \in \mathbb R$ denote the element-wise mean and standard deviation\footnote{
    Specifically, $m(\bm y) := \frac{1}{d}\sum_k \bm y^{(k)}$ and $s(\bm y) := \sqrt{\frac{1}{d} \sum_k \bigl(\bm y^{(k)} - m(\bm y) + \epsilon\bigr)^2}$, where $\bm y^{(k)}$ denotes the $k$-th element of the vector $\bm y$ and $\epsilon\in\mathbb{R}$ is a small constant to stabilize the value.
}, respectively.
Here, subtraction and division are also performed on an element-wise basis.
The normalized vector, $(\bm y - m(\bm y)) / s(\bm y)$, is then transformed with $\bm \gamma$ and $\bm \beta$; here, $\odot$ denotes the element-wise multiplication.

Note that analyzing the feed-forward networks in each layer is beyond the scope of this study and will be carried out as future work.

\section{Proposal: Analyzing attention blocks}
\label{sec:proposal}
For analyzing Transformers, solely observing the attention weights has been a common method~\cite[etc.]{clark19,kovaleva19}.
We extend the scope of analysis to the whole attention block (\textsc{ATTN}, \textsc{RES}, and \textsc{LN}).

\subsection{Strategy: Norm-based analysis}
\label{sec:proposed_method}

\citet{kobayashi-etal-2020-attention} introduced the \emph{norm-based analysis} to extend the scope of analysis from the attention weights to the whole multi-head attention.
We follow this \emph{norm-based analysis} and extend its scope to the whole attention block.

The norm-based analysis first attempts to decompose the output vector $\widetilde{\bm x}_i$ into the sum of the transformed input vectors $\{\bm x_j\}$:
\begin{align}
    \textstyle
    \widetilde{\bm x}_i
    = \sum_j F_i(\bm x_j)
    \text{,}
\end{align}
where $F_i$ is an appropriate vector-valued function.
Then, the contribution of $\bm x_j$ to
$\widetilde{\bm x}_i$
can be expressed by the norm of $F_i(\bm x_j)$.
In the next subsection, we indicate that this norm-based method can be applied to analyzing the whole attention block.
In other words, the output of the attention block is also be decomposed into the sum of the transformed input vectors without any approximation.

\subsection{Decomposing output into a sum of inputs}
\label{subsec:proposal_overview}
The output $\widetilde{\bm x}_i$ is decomposed into a sum of terms associated with each input $\bm x_j$.
First, ATTN (Equation~\ref{eq:attn_multihead}) can be decomposed into a sum of vectors~\cite{kobayashi-etal-2020-attention}:
\begin{align}
    \mathrm{ATTN}(\bm x_i, \bm X)
    = \sum_{j=1}^n \sum_{h=1}^H \alpha^{\encircled{h}}_{i,j} f^{\encircled{h}}(\bm x_j)
    \text{,}
    \label{eq:attn_decomposed}
    \\
    f^{\encircled{h}}(\bm x) := \left(\bm{x} \bm{W}_V^{\encircled{h}}+\bm{b}_V^{\encircled{h}} \right)\bm W_O^{\encircled{h}}
    \label{eq:attn_f}
    \text{.}
\end{align}

\noindent
Second, in RES, no interaction between subscripts $i$ and $j$ occurs, and the form is already additively decomposed.
Third, by exploiting the linearity of $m(\cdot)$, we can derive the ``distributive law'' of LN and decompose it. 
Let $\bm y = \sum_j \bm y_j$ be the input to LN.
Then,
\begin{align}
    \mathrm{LN}(\bm y)
    &= \sum_j g_{\bm y}^{}(\bm y_j) + \bm \beta
    \label{eq:distributional_law_ln}
    \text{,}
    \\
    g_{\bm y}^{}(\bm y_j)
    &:= \frac{\bm y_j - m(\bm y_j)}{s(\bm y)}\odot \bm \gamma
    \label{eq:g_ln}
    \text{.}
\end{align}
See Appendix~\ref{ap:precise_equation} for the derivation.

With these decompositions of ATTN and LN, the output of the whole attention block can be written as the sum of vector-valued functions with each input vector in $\bm X$ as an argument:
\begin{align}
    \widetilde{\bm x}_i 
    &= \mathrm{LN}\left(\mathrm{ATTN}(\bm x_i, \bm X) + \bm x_i\right)
    \\
    &= \mathrm{LN}\Biggl( \sum_{h=1}^H \sum_{j=1}^n \alpha_{i,j}^{\encircled{h}} f^{\encircled{h}}(\bm{x}_j) + \bm{x}_i\Biggr)
    \\
    &= \sum_{j \neq i} g_{\bm y}\Bigl(\sum_{h=1}^H \alpha_{i,j}^{\encircled{h}} f^{\encircled{h}}(\bm{x}_j)\Bigr)
    \nonumber
    \\
    &\:\:\:\:
    + g_{\bm y}\Bigl(\sum_{h=1}^H \alpha_{i,i}^{\encircled{h}} f^{\encircled{h}}(\bm{x}_i)\Bigr)
    + g_{\bm y}(\bm{x}_i)
    \label{eq:decomposition}
    \\
    &\:\:\:\:
    + \bm \beta
    \text{,}
    \nonumber
    \\
    \bm y &:= \mathrm{ATTN}(\bm X, \bm x_i) + \bm x_i
    \text{.}
\end{align}

\subsection{Measuring the contribution of context}
Regarding the success of the contextualized representations in NLP, 
an interesting issue is the \textit{location and strength of the context mixing performed in the model}.
Based on this issue, we investigate the attention block by categorizing the terms in Equation~\ref{eq:decomposition} into the two effects:\footnote{
    The bias $\bm \beta$ affects neither the \textit{context-mixing} effect nor the \textit{preserving} effect. Thus, we ignored this term in our analysis.
}%
\begin{enumerate}
\setlength{\parskip}{0cm} 
\setlength{\itemsep}{0.1cm}
    \item \textit{Mixing} contextual information into the output representation by the ATTN:
    \vspace{-0.3cm}
    \begin{align}
    \nonumber
        \widetilde{\bm x}_{i\gets \emphmix{}{\scriptstyle \text{context}}}:=\sum_{{}{\scriptstyle j}\neq i} g_{\bm y}\Bigl(\sum_{h=1}^H \alpha_{i,j}^{\encircled{h}} f^{\encircled{h}}(\emphmix{}{\bm{x}_j})\Bigr)
        \text{.}
    \end{align}
    We measure the magnitude of this \emph{context-mixing effect} by the norm
    $\lVert \widetilde{\bm x}_{i\gets\text{context}}\rVert$.
        This strength refers to the amount of information from the surrounding contexts
    $\{\bm x_1,\dots,\bm x_n\}\setminus\{\bm x_i\}$
    in calculating
    $\widetilde{\bm x}_i$.
\vspace{0.2cm}
    \item \textit{Preserving} the original information via ATTN and RES:
    \vspace{-0.3cm}
    \begin{align}
    \nonumber
        \!\!\!
        \widetilde{\bm x}_{i\gets \emphpreserve{}{\scriptstyle i}}:= g_{\bm y}\Bigl(\sum_{h=1}^H \alpha_{i,i}^{\encircled{h}} f^{\encircled{h}}(\emphpreserve{}{\bm{x}_i})\!\Bigr)
        + g_{\bm r}(\emphpreserve{}{\bm x_i})
        \text{.}
    \end{align}
        We measure the magnitude of the \emph{preserving effect} by the norm $\lVert\widetilde{\bm x}_{i\gets i}\rVert$.
    This strength  refers to the amount of information from the original vector $\bm{x}_i$ used in calculating $\widetilde{\bm x}_i$.
    At the attention block, information from the input vector $\bm x_i$ can flow through two ways: %
    (i) attention to the original input (the first term)
    and (ii) residual connection (the second term).

\end{enumerate}

To summarize the relative strength of the context-mixing effect, the \textbf{context-mixing ratio} is defined as follows:
\begin{align}
    r_{i}
    = \frac
    {\lVert\widetilde{\bm x}_{i\gets\emphmix{}{\scriptstyle \text{context}}}\rVert}
    {\lVert\widetilde{\bm x}_{i\gets\emphmix{}{\scriptstyle \text{context}}}\rVert + \lVert\widetilde{\bm x}_{i\gets \emphpreserve{}{\scriptstyle i}}\rVert}
    \text{.}
    \label{eq:mixing_rate}
\end{align}
A higher mixing ratio indicates that the mixing effect is more dominant than the preserving effect in the computation of $\widetilde{\bm x}_i$.

Note that \citet{abnar-zuidema-2020-quantifying} assumed that 
the multi-head attention and residual connection always equally impact the output, i.e., $r \approx 0.5$ in the analysis of Transformers.
However, our experiments revealed that, in practical masked language models, the mixing ratio is considerably below $0.5$.

\section{Experiments: Analysis of mixing ratio}
\label{sec:experiments-pretrained_bert}
The context-mixing ratio of the attention blocks in pre-trained masked language models was analyzed using the proposed norm-based analysis.
The obtained results were different from those of the existing methods that analyze only some of the components in the attention block.

\subsection{General setup}
\label{sec:experiment_setup}

\paragraph{Model:}
We investigated the 32 variants of the masked language models (BERT with five different sizes, BERT-base trained with 25 different seeds, and RoBERTa with two different sizes). %
In Section~\ref{sec:experiments-pretrained_bert}, the results for BERT-base and RoBERTa-base are demonstrated.
The results for the other models are provided in Appendix~\ref{ap:mix_ratio_variant} and \ref{ap:singular_values}.
Note that most of our findings reported in this section generalize across these model variants.
Exceptions are discussed in the relevant section (Section~\ref{subsec:mechanism}).

\paragraph{Data:}
We experimented with the following four datasets:
(i) Wikipedia~\cite{clark19}%
, (ii) the Stanford Sentiment Treebank v2 dataset~\cite[SST-2,][]{socher-etal-2013-sst}, (iii) the Multi-Genre Natural Language Inference corpus~\cite[MNLI,][]{williams-etal-2018-mnli}, and (iv) the standard CoNLL-2003 Named Entity Recognition dataset~\cite[CoNLL'03-NER,][]{tjong-kim-sang-de-meulder-2003-conll2003}.
The statistics of the datasets are shown in Table~\ref{table:dataset_stats}.
Owing to the limitation of space, we only give the results for the Wikipedia data in Section~\ref{sec:experiments-pretrained_bert}.
The trends observed for the Wikipedia dataset were generalized across the other datasets (see Appendix~\ref{ap:mix_ratio_variant}).
Note that each sequence of the Wikipedia dataset consists of paired consecutive paragraphs.
Each sequence is fed into the models with masking applied to $15$\% of tokens $80$\% of the time.%
\footnote{
    For the other datasets, we used 1000 samples from their validation set or used all of their validation set if the number of sequences is less than 1000.
}%

\begin{table}[t]
\centering
\setlength{\tabcolsep}{1.3pt} 
\renewcommand{\arraystretch}{1.0}
{\small
\begin{tabular}{lccc}
\toprule
Data & \#Samples & Avg. length & Domains \\
\cmidrule(r){1-1} \cmidrule(lr){2-2} \cmidrule(lr){3-3} \cmidrule(l){4-4}
Wikipedia & 992 & 122 & Web encyclopedia \\
SST-2 & 872 & 25 & Movie reviews \\
MNLI & 1000 & 39 & 10 distinct genres \\
CoNLL'03-NER & 1000 & 21 & News \\
\bottomrule
\end{tabular}
}
\caption{
Details of the datasets. Avg. length is the number of tokens segmented by BERT per sample. 
}
\label{table:dataset_stats}
\end{table}

\paragraph{Analysis methods:}
We compared the context-mixing ratio computed with the following five analyzing methods:
\begin{itemize}
\setlength{\parskip}{0cm} 
\setlength{\itemsep}{0.1cm}
    \item \textsc{Attn-w}: Analyzing ATTN via attention weights, which has been applied in many existing studies~\cite[etc.]{clark19,kovaleva19,marecek19balustrades}.
    The ratio, where attention weight assigned to the 
    original input vector $\alpha_{i,i}$ %
    corresponds to the preserving effect, and the others correspond to the mixing effect, is calculated as follows:
    \vspace{-0.2cm}
    \begin{align}
        \small
        \nonumber
        \frac{1}{H} \sum_{h=1}^H \frac{\sum_{j\neq i}\alpha_{i,j}^{\encircled{h}}}{\sum_{j\neq i}\alpha_{i,j}^{\encircled{h}} + \alpha_{i,i}^{\encircled{h}}} = \frac{\sum_{h}\sum_{j\neq i}\alpha_{i,j}^{\encircled{h}}}{H}
        \text{.}
    \end{align}
    \item \textsc{Attn-n}: Analyzing ATTN via the vector norm~\cite{kobayashi-etal-2020-attention}.
    The mixing ratio is calculated as
        \vspace{-0.2cm}
    \begin{align}
        \small
        \nonumber
        \frac{\norm{\sum_{h} \sum_{j\neq i} \alpha_{i,j}^{\encircled{h}} f^{\encircled{h}}(\bm x_j)}}{\norm{\sum_{h}\sum_{j\neq i} \alpha_{i,j}^{\encircled{h}} f^{\encircled{h}}(\bm x_j)} + \norm{\sum_{h=}\alpha_{i,i}^{\encircled{h}} f^{\encircled{h}}(\bm x_i)}}
        \text{.}
    \end{align}
    \item \textsc{AttnRes-w}: Analyzing ATTN and RES via attention weights, as \citet{abnar-zuidema-2020-quantifying} did.
    They assumed that the residual-aware attention matrix is constructed as $0.5\bm A + 0.5\bm I$.
    Here, $\bm A$ is the actual attention matrix and $\bm I$ is the identity matrix considered as the effect of residual connection.
    The mixing ratio is calculated as
        \vspace{-0.2cm}
    \begin{align}
        \small
        \nonumber
        \frac{1}{H} \sum_{h=1}^H\frac{\sum_{j\neq i}0.5\alpha_{i,j}^{\encircled{h}}}{\sum_{j}0.5\alpha_{i,j}^{\encircled{h}}+0.5}
        \text{.}
    \end{align}
    \item \textsc{AttnRes-n} (\textbf{proposed}): Analyzing ATTN and RES via the vector norm -- a version of our proposed method that does not consider LN.
    The mixing ratio is calculated as
        \vspace{-0.2cm}
    \begin{align}
        \small
        \nonumber
        \!\!\!\! \frac{\norm{\sum_{h}\sum_{j\neq i}\alpha_{i,j}^{\encircled{h}} f^{\encircled{h}}(\bm x_j)}}{\norm{\sum_{h}\sum_{j\neq i}\alpha_{i,j}^{\encircled{h}} f^{\encircled{h}}(\bm x_j)} + \norm{\sum_{h}\alpha_{i,i}^{\encircled{h}} f^{\encircled{h}}(\bm x_i) + \bm x_i}}
        \text{.}
    \end{align}
    \item \textsc{AttnResLn-n} (\textbf{proposed}): Analyzing ATTN, RES, and LN via the vector norm -- the method proposed in Section~\ref{sec:proposal}.
    This corresponds to the $r_i$ in Equation~\ref{eq:mixing_rate}.
\end{itemize}

\subsection{Results}
\label{subsec:bert_exp:mixing_rate}

\begin{table}[t]
\centering
\setlength{\tabcolsep}{4pt} 
\renewcommand{\arraystretch}{0.85}
    {\small
\begin{tabular}{lrrr}
\toprule
\multicolumn{1}{c}{Methods} & Mean & Max & Min \\
\cmidrule(r){1-1} \cmidrule(lr){2-2} \cmidrule(lr){3-3} \cmidrule(l){4-4}
\multicolumn{1}{c}{--- BERT-base ---} \\
\textsc{Attn-w} & 97.1 & 100.0 & 45.0 \\
\textsc{Attn-n}  &  85.2  &  100.0 &  4.9 \\ 
\textsc{AttnRes-w}  &  48.6  &  50.0 &  22.5 \\ 
\textsc{AttnRes-n} &  22.3 & 65.7 & 2.0 \\
\textsc{AttnResLn-n} &  {\bf 18.8} & 61.3 & 1.3 \\
\multicolumn{1}{c}{--- RoBERTa-base ---} \\
\textsc{Attn-w} & 95.8 & 100.0 & 3.8 \\
\textsc{Attn-n}  &  84.4  &  100.0 &  13.8 \\ 
\textsc{AttnRes-w}  &  47.9  &  50.0 &  1.9 \\ 
\textsc{AttnRes-n} &  19.6 & 69.9 & 1.8 \\
\textsc{AttnResLn-n} &  {\bf 16.2} & 73.4 & 1.5 \\

\bottomrule
\end{tabular}
}
\caption{Mean, maximum, and minimum values of the mixing ratio computed with each method.
}
\label{table:bert_base_overall_mixing_ratio}
\end{table}

We computed the mixing ratio of each token in each layer (each attention block) of the models with the five analysis methods (Section~\ref{sec:experiment_setup}).
The average, maximum, and minimum mixing ratios are shown in Table~\ref{table:bert_base_overall_mixing_ratio}.
Each row corresponds to a different analysis method.

\paragraph{Lower mixing ratio than in existing methods:}
Table~\ref{table:bert_base_overall_mixing_ratio} shows that the mixing ratios obtained from the proposed \textsc{AttnRes-n} and \textsc{AttnResLn-n} largely differ from those obtained from the existing methods.
Whereas the attention analyses (\textsc{Attn-w} and \textsc{Attn-n}) yield mixing ratios of $84$--$97\%$ and \textsc{AttnRes-w} yields $48\%$--$49$\%, our proposed method (\textsc{AttnResLn-n}) yields about $16$ and $19\%$ on average.
The visualizations of the token-by-token interactions in the common attention map style become almost diagonal patterns (Figure~\ref{fig:visualize_comparison}).
These demonstrate that each layer's context mixing is lower than previously expected, and RES and LN largely cancel the mixing by ATTN.
Observing the only ATTN and making an inference about the Transformer layer may lead to misleading.
Note that \citet{srivastava2015training} reported a similar trend that stacked feed-forward networks tend to prioritize the ``preserving'' effect in skip connections.

\paragraph{Consistent trends across model sizes:}
Our method consistently shows the lowest mixing ratio among the compared methods for BERT and RoBERTa models of various sizes (BERT-large, medium, small, tiny, and RoBERTa-large) (Appendix~\ref{ap:mix_ratio_variant}).
Interestingly, the context-mixing ratio is higher in the models with fewer layers ($37\%$ in BERT-tiny, but $15\%$ in BERT-large).

\begin{figure*}[t]
    \centering
    \begin{minipage}[t]{.19\hsize}
        \centering
        \includegraphics[height=6cm]{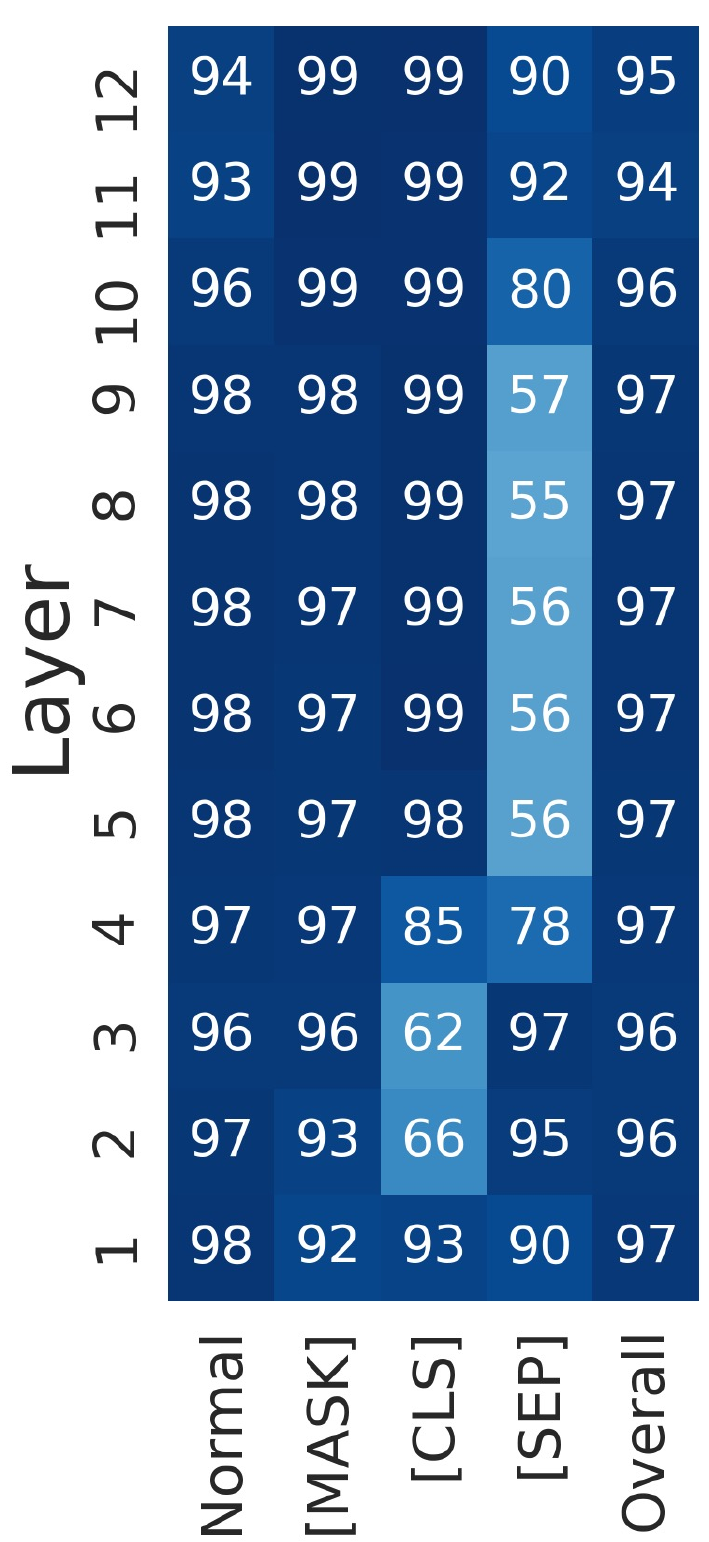}
        \subcaption{
        \textsc{Attn-w}.
        }
        \label{fig:bert_weight_rate}
    \end{minipage}
    \;
    \begin{minipage}[t]{.17\hsize}
        \centering
        \includegraphics[height=6cm]{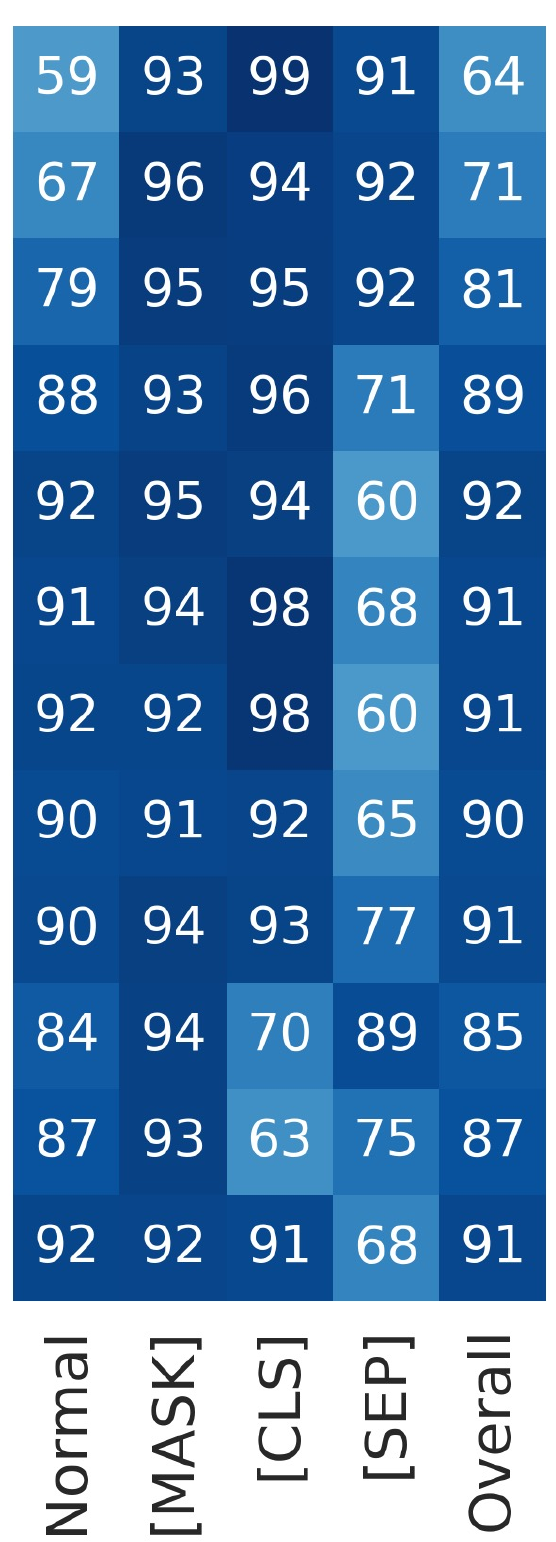}
        \subcaption{
        \textsc{Attn-n} \cite{kobayashi-etal-2020-attention}.
        }
        \label{fig:bert_kobayashi_rate}
    \end{minipage}
    \;
    \begin{minipage}[t]{.17\hsize}
        \centering
        \includegraphics[height=6cm]{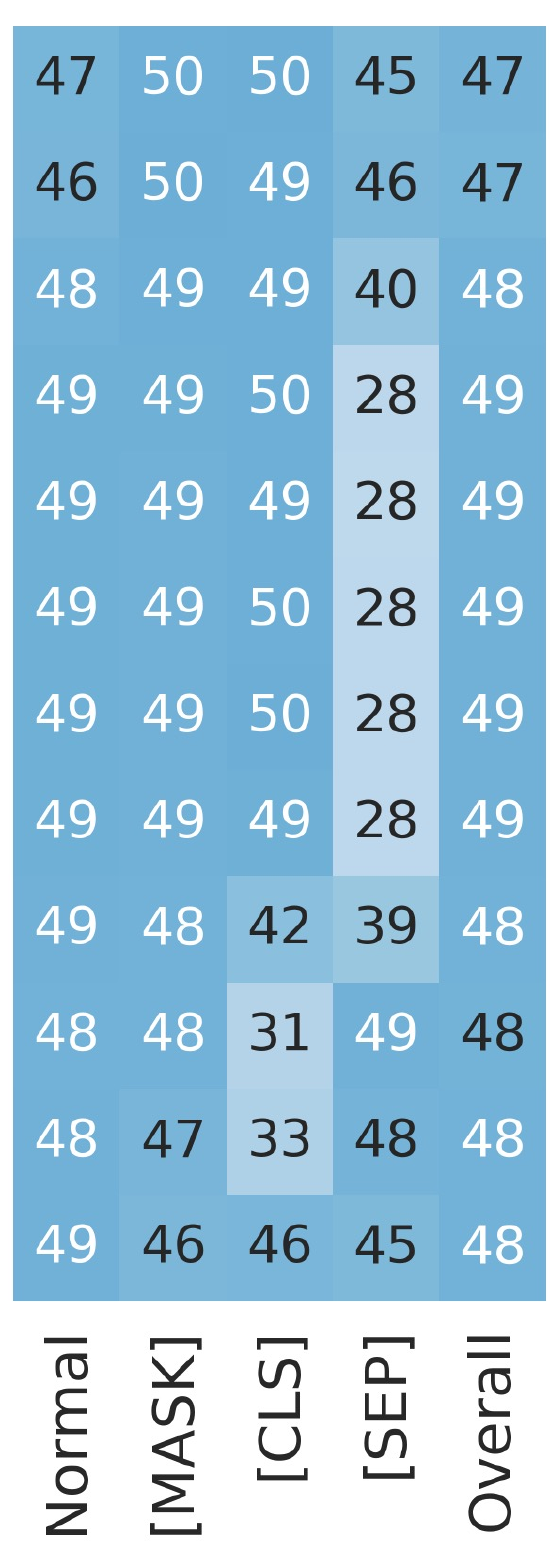}
        \subcaption{
        \textsc{AttnRes-w} \cite{abnar-zuidema-2020-quantifying}.
        }
        \label{fig:bert_abnar_rate}
    \end{minipage}
    \;
    \begin{minipage}[t]{.17\hsize}
    \centering
    \includegraphics[height=6cm]{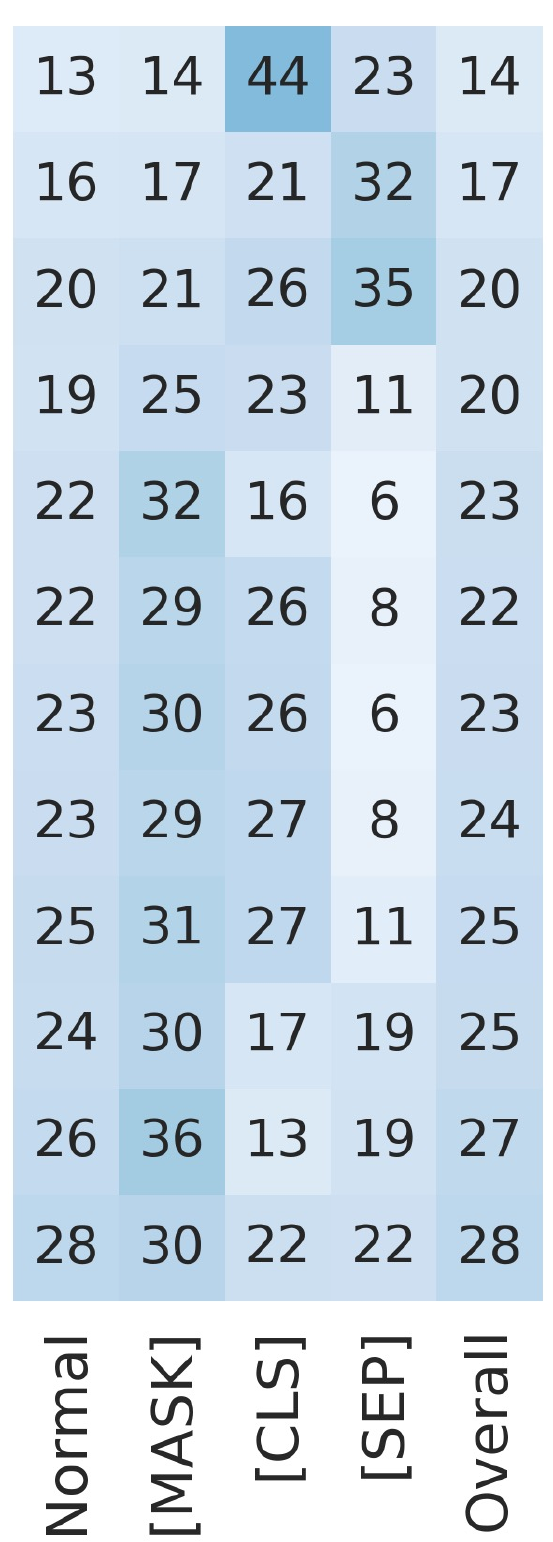}
    \subcaption{
    \textsc{AttnRes-n}.
    }
    \label{fig:before_ln_mixing_rate}
    \end{minipage}
    \;
    \begin{minipage}[t]{.22\hsize}
    \centering
    \includegraphics[height=6cm]{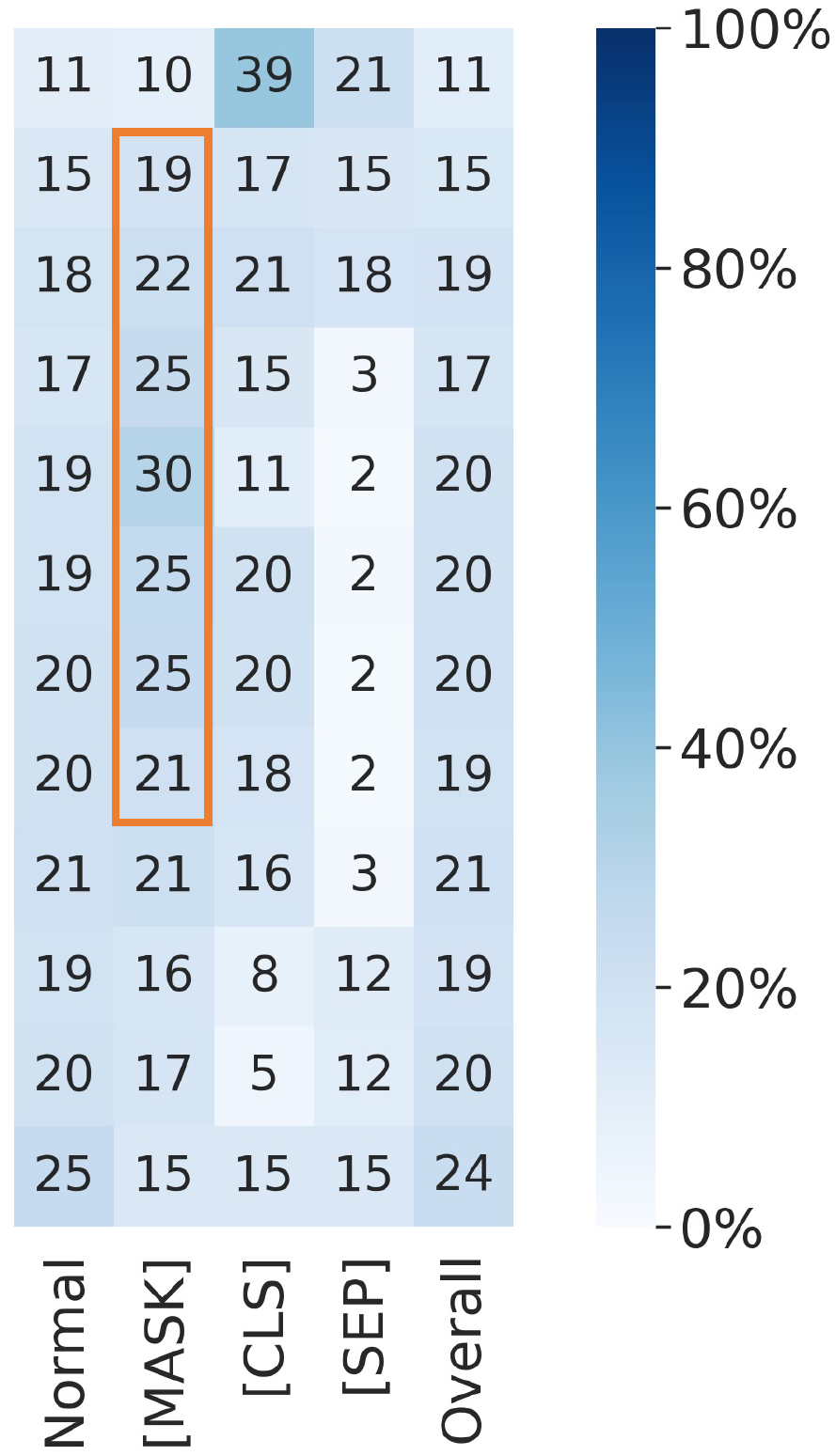}
    \subcaption{
    \textsc{AttnResLn-n}.
    }
    \label{fig:bert_mix_rate}
    \end{minipage}
    \caption{
    Mixing ratio in each layer of BERT calculated from each method.
    }
    \label{fig:mixing_rate}
\end{figure*}

\subsection{Connections to previous studies}
Our finding of a lower mixing ratio than previously thought provides explanations for previous results and is consistent with the pre-training strategy.

\paragraph{Token identifiability:}
The low context-mixing ratio is consistent with \citet{brunner19}'s reports on what they called ``token identifiability.''
They showed that input tokens can be well predicted only from the corresponding internal representations within BERT, especially in shallower layers,
suggesting that context mixing is performed little by little.
Our analysis results of the whole attention block were consistent with this finding. 

\paragraph{Masked language modeling objective:}
Regarding the masked token prediction task\footnote{%
    For masked language modeling in BERT and RoBERTa pre-training, $15\%$ of the tokens are randomly chosen from the input sequence, of which $80\%$ are replaced with \texttt{[MASK]}, $10\%$ are replaced with random words, and $10\%$ are kept unchanged.
} during the pre-training, BERT and RoBERTa learn to conduct the following operations for a given input sequence: (i) infilling the \texttt{[MASK]} with plausible words,
(ii) replacing the normal (non-special) tokens that might not fit their context (i.e., randomly replaced tokens) with plausible one, and (iii) reconstructing the original input tokens that might fit their context.%

In our experiments and in common practical use, most tokens in the input sequence are not masked and fit their context. 
Thus, BERT is assumed to reconstruct the inputs for these tokens (i.e., behave as an auto-encoder).
From this point of view, the superiority of the preserving effect is the intuitive behaviors of the masked language models.

\paragraph{Low impact of discarding learned attention patterns:}
Several studies have reported the low impact of discarding the learned attention patterns in Transformers.
\citet{michel19} and \citet{kovaleva19} reported that the attention patterns of many attention heads in Transformers can be removed or overwritten into the uniform patterns with almost no change in their performance, and this even brought about improvements in some cases.
\citet{voita19-analyzing-self-attn} also reported the same phenomenon using a pruning method with additional training.
In addition, \citet{toneva-nips2019-interpreting} reported that using uniform attention in early layers of BERT instead of the learned attention patterns leads to a better ability to simulate human brain activity.

Our analysis shows that most of the attention signal is reduced by the immediately following modules, RES and LN.
This fact may explain the above observations that discarding the learned attention patterns of many attention heads does not cause a severe difference.

\subsection{Mechanism}
\label{subsec:mechanism}
How is the mixing effect conducted in multi-head attention largely suppressed in the whole attention block?
We discuss the mechanism role of ATTN and LN in suppressing the mixing ratio. 

\paragraph{ATTN reduces context-mixing ratio:}
RES is a mechanism that equally adds together the output of ATTN and the input in a one-to-one fashion (Equation~\ref{eq:attn+res}).
Considering this, the mixing ratio in the scope of ATTN and RES is expected to be about $50\%$, while the mixing ratio was actually substantially below $50\%$ ($19$--$22\%$ in \textsc{AttnRes-n}) (Section~\ref{subsec:bert_exp:mixing_rate}).
This suggests that the output of ATTN is much smaller than the input; 
in other words, ATTN seems to have the effect of largely \emph{shrinking} inputs to compute the output.
How is this achieved?

Recall that the output of ATTN is a weighted sum of the affine-transformed vector $f^{\encircled{h}}(\bm x)$ using with attention weight $\alpha^{\encircled{h}}_{i,j}$ (Equation~\ref{eq:attn_decomposed}).
We describe and empirically show that (i) the affine transformation in ATTN has the effect of shrinking the inputs, and (ii) the attention weights and affine-transformed vectors cancel each other out on specific vectors.
We describe a brief idea here and provide the detailed derivation of each equation in the Appendix~\ref{ap:singular_values}.

First, under a coarse assumption, multiple affine transformations performed in the multi-head attention can be integrated into a single one:
\begin{align}
    \textstyle
    f(\bm x) := \sum_{h=1}^H f^{\encircled{h}}(\bm x)
    \text{.}
\end{align}
Assume that the input vector $\mathbf x$ is a sample from the standard normal distribution: 
$\mathbf x \sim \mathcal N(\bm 0, I_d)$.
Then we can estimate its magnitude by $\mathbb E \norm{\mathbf x} \approx \sqrt d$ and the magnitude after affine transformation by $\mathbb E \norm{f(\mathbf x)} \approx \sqrt{\sum_{k=1}^d \sigma_k^2}$,
where $\sigma_1,\dots,\sigma_{d}$ denote singular values of $f$.
Thus, the expansion rate of $f$
is approximately estimated by
\begin{align}
     \textstyle \norm{f(\mathbf x)} / \norm{\mathbf x} \approx \sqrt{\sum_{k=1}^d \sigma_k^2} \,/\, \sqrt{d}
     \text{.}
\end{align}

If the ratio is lower than one, $f$ has a tendency of shrinking the input.
For commonly used large models, results stably demonstrated the shrinking tendency (layer mean of the expansion rate was $0.88 < 1.0$ for BERT-base and $0.80 < 1.0$ for BERT-large).
Note that, for smaller models, results demonstrated the expanding tendency (layer mean $1.24$ for BERT-mini and $1.86$ for BERT-tiny).
This is consistent with the result that the latter models tended to have a higher mixing ratio than the former models (Section~\ref{subsec:bert_exp:mixing_rate}).
Detailed results are shown in Appendix~\ref{ap:subsec:singular_results_models}.

Furthermore, attention weight $\alpha^{\encircled{h}}_{i,j}$ boosts the shrinking effect.
\citet{kobayashi-etal-2020-attention} reported the negative correlation between $\norm{f^{\encircled{h}}(\bm x_j)}$ and $\alpha^{\encircled{h}}_{i,j}$ on frequent tokens.
That is, ATTN wastes a lot of attention weights $\alpha^{\encircled{h}}_{i,j}$ by assigning them to small vectors $\norm{f^{\encircled{h}}(\bm x_j)}$.

To summarize, ATTN's shrinking effect is probably achieved by (i) the shrinking in $f$ alone and (ii) further shrinking through the cancellation of $\alpha$ and $\norm{f(\bm x)}$.
By these mechanisms, ATTN can contribute to decreasing the mixing ratio.

\paragraph{LN reduces the context-mixing ratio:}
LN contains not only the vector normalization but also the affine transformation with learnable parameters (Equation~\ref{eq:layernorm}).
Although the validity or usage of LN has been investigated in terms of stability and speed of training~\cite{parisotto2020stabilizing,liu-etal-2020-rethinking}, the effects of affine transformation have rarely been explored. %
By comparing the mixing ratios obtained from \textsc{AttbRes-n} and \textsc{AttnResLn-n}
(Table~\ref{table:bert_base_overall_mixing_ratio}), we discovered that LN reduced the context-mixing ratio.
This suggests that the scaling (by $\bm \gamma$) of the affine transformation shrinks the vector from ATTN and emphasizes %
RES over ATTN.

\section{Detailed analysis}
\label{sec:detailed_analysis}
We further analyzed the mixing ratio of the masked language models in detail from the perspectives of both the layer and word attributes.
In this section, we inherit the experimental setup (Section~\ref{sec:experiment_setup}) from the previous section and demonstrate results for BERT-base with the Wikipedia dataset.
The results for the other experimental settings are shown in Appendix~\ref{ap:mix_ratio_variant} and \ref{ap:relation_with_freq_variant}.
Note that only the finding reported in Section~\ref{subsec:bert_exp:freq} did not generalize across model variants, and we exceptionally discuss this point in the body.

\subsection{Differences by layers and token types}
\label{subsec:analysis:pipeline}

Figure~\ref{fig:mixing_rate} shows the mixing ratio in each layer
of the BERT model (results for other models are shown in Appendix~\ref{ap:mix_ratio_variant}).
Each subfigure corresponds to a different analysis method, each row represents a layer, and each column represents a token type. 
The averaged results of the following token categories and their overall average (``overall'') are reported: (i) non-special tokens (``normal''), (ii) \texttt{[MASK]}, (iii) \texttt{[CLS]}, and (iv) \texttt{[SEP]}.

\paragraph{Results and discussion: }
Our proposed method showed that the mixing ratio is higher in the earlier layers than in the later ones (see the ``overall'' trend in Figure~\ref{fig:bert_mix_rate}).%
\footnote{
    The Spearman's $\rho$ between the ``overall'' mixing ratio and the layer depth are $-0.67$ and $-0.98$ in ``overall'' of BERT-base and RoBERTa-base, respectively.
} 
This trend mirrors the tendency that a deep neural network with ``gates'' similar to residual connections passes through the input more in the later layers~\cite{srivastava2015training}.

Furthermore, our method showed a distinctive trend for the \texttt{[MASK]} tokens.
In the middle and deep layers, the mixing ratio for \texttt{[MASK]} becomes higher ($19$--$30\%$) than the overall trends ($15$--$20\%$).
Note that this trend becomes clearer when considering the \textsc{RES} and \textsc{LN}.
This trend implies that in the middle and deep layers, BERT refers to contextual information for predicting the masked words. 
The trends of the other masked language models are shown in Appendix~\ref{ap:mix_ratio_variant}.

\begin{table}[t]
\centering
\setlength{\tabcolsep}{2pt} 
\renewcommand{\arraystretch}{1.0}
{\small
\begin{tabular}{lrr}
\toprule
\multicolumn{1}{c}{\multirow{2}{*}{Methods}} & \multicolumn{2}{c}{Spearman's $\rho$}     \\
\multicolumn{1}{c}{}                         & all tokens & w/o special tokens \\ 
\cmidrule(r){1-1} \cmidrule(lr){2-2} \cmidrule(l){3-3}
\textsc{Attn-w} & $0.16$      & $0.14$   \\
\textsc{Attn-n}  &  $-0.39$  &  $-0.41$  \\ 
\textsc{AttnRes-w}  &  $0.16$  &  $0.14$  \\ 
\textsc{AttnRes-n}         & $-0.84$       & $-0.86$   \\
\textsc{AttnResLn-n}      & $-0.54$       & $-0.58$    \\ 
\bottomrule
\end{tabular}
}
\caption{
Spearman's $\rho$ between the frequency rank and the mixing ratio calculated by each method.
In the ``w/o special tokens'' setting, it was calculated without \texttt{[CLS]} and \texttt{[SEP]}.
}
\label{table:relation_with_freq}
\end{table}

\begin{figure}[t]
    \centering
    \includegraphics[height=5cm]{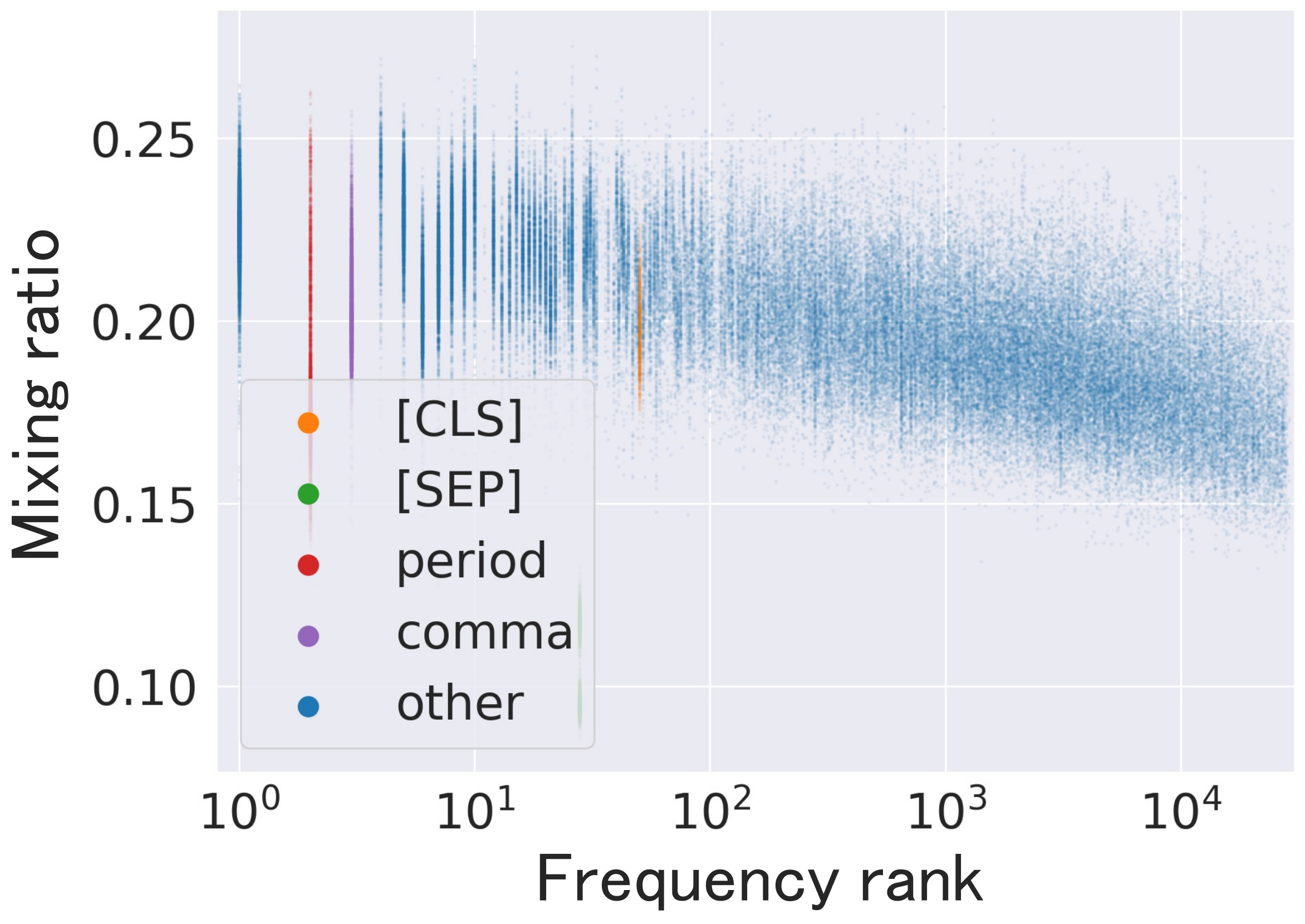}
    \caption{
    Relationship between the frequency rank of tokens and the mixing ratio calculated with the \textsc{AttnResLn-n}.
    }
    \label{fig:relation_with_freq}
\end{figure}

\subsection{Word frequency and mixing ratio}
\label{subsec:bert_exp:freq}
In this section, we will discuss the property of BERT related to the word frequency.%
\footnote{
    Following \citet{kobayashi-etal-2020-attention}, we counted the frequency for each word type by reproducing the training data of BERT.
}

\paragraph{Results: }
Table~\ref{table:relation_with_freq} lists the Spearman's rank correlation $\rho$ between the \textbf{frequency rank}
(e.g., $\mathrm{rank}(\text{``the''}) = 1$, $\mathrm{rank}(\text{``and''}) = 6$, etc.) and the \textbf{mixing ratio} across tokens in the text data.

The results obtained from \textsc{AttnRes-n} and \textsc{AttnResLn-n} indicate a surprisingly stronger negative correlation than the results obtained by the existing methods (Figure~\ref{fig:relation_with_freq}).
This indicates that BERT discounts the information of high-frequency words compared with low-frequency ones.%
\footnote{
    \citet{kobayashi-etal-2020-attention} reported that ATTN in BERT tends to discounts frequent words when mixing contexts.
    We found even stronger trends after broadening the scope of analysis.
}

\paragraph{Discussion: }
Discounting high-frequency words is %
a common practice for making the semantic representation of a sentence or a text from word representations; examples are Luhn's heuristic in classical text summarization~\cite{luhn-1958-summarization} and the smooth inverse frequency (SIF) weighting in sentence vector generation~\cite{arora-2017-sif}.
Our frequency-based results reveal that attention blocks in BERT achieve this desirable property.

Our observation may also explain the phenomenon that adding up BERT's internal or output representations does not produce a good sentence vector~\cite{reimers-gurevych-2019-sentencebert}. 
In contrast, in static word embeddings (e.g., word2vec~\cite{mikolov2013-word2vec}), the norm encodes the word importance derived from its frequency~\cite{schakel-2015-measuring};
we can generate a good sentence vector by simply adding these static word vectors~\cite{yokoi-etal-2020-word}. 
Our finding suggests that BERT encodes the token's importance through %
the context-mixing ratio rather than the norm.%
\footnote{
    In BERT, it may be difficult for the norm to encode the token importance, because the norm is fixed at each layer normalization.
} 
In this sense, it is plausible that additive composition using BERT's internal or output representations does not perform well.

\paragraph{Generalizability: }
Contrary to the other experimental results, only the relationship between word frequency and mixing ratio (Figure~\ref{fig:relation_with_freq}) was not consistent across different model sizes. 
For the larger variant (BERT-large), a stronger negative correlation between them was indicated than for BERT-base, while for the smaller variants (BERT-medium, BERT-small, BERT-mini, and BERT-tiny), even a positive correlation or no correlation was indicated (see Appendix~\ref{ap:relation_with_freq_variant}).
Generally, larger BERT models (BERT-base and BERT-large) achieve better performance on downstream tasks. %
The different results across model sizes suggest that this %
desirable property can be learned when the representational power is sufficient.

\section{Related work}
\label{sec:related_work}

\subsection{Probing Transformers}
As current neural-based models have an end-to-end, black-box nature, existing studies have adopted several strategies to interpret their inner workings~\cite{electronics8080832,rogers-etal-2020-primer,visualizing-survey}.
In analyzing Transformers, previous studies have mainly employed the following approaches: (i) observing the vanilla attention weights~\cite{clark19,kovaleva19,coenen19,lin19,marecek19balustrades,Htut2019,raganato2018,tang_transformer_attention_analysis} or the extended version~\cite{brunner19,abnar-zuidema-2020-quantifying},
(ii) computing the gradient~\cite{clark19,brunner19},
and (iii) analyzing the vector norm~\cite{kobayashi-etal-2020-attention}.
We adopted the norm-based analysis because this method can be naturally extended to the analysis of the whole attention block and it has some advantages~\cite{kobayashi-etal-2020-attention} that will also be discussed in the following paragraph.

As for broadening the scope of the analysis,
\citet{abnar-zuidema-2020-quantifying} modified the attention matrix to incorporate the residual connections into the analysis. 
However, they assumed that the multi-head attention and residual connection equally contributed to the computation of the output representations without any justification (Section~\ref{sec:experiment_setup}).
\citet{brunner19} employed a gradient-based approach for analyzing the interaction of input representations; however, the gradient ignores the impact of the input vector (i.e., only observing $\partial \widetilde{\bm x_i}/\partial \bm x_j$ neglects the impact of $\bm x_j$ itself) as described in Section 6.2 of \citet{kobayashi-etal-2020-attention}.
Note that our norm-based analysis can include the magnitude of the impact of inputs in the analysis.

\subsection{RES and LN in Transformers}
Although residual connections (RES)~\cite{he2016deep} and layer normalization (LN)~\cite{Ba2016_Layernormalization} have rarely been considered in probing studies, they are known to play important roles in both model performance and training convergence~\cite{parisotto2020stabilizing,liu-etal-2020-rethinking}.
\citet{dong21-attention-doubly-exp} revealed that the residual connections are important in attention-based architectures.
They demonstrated that the output of self-attention networks without residual connections converges to a rank-1 matrix quickly with increasing its layer depth.
In addition, as a similar component to RES, \citet{srivastava2015training} proposed ``gates'' that adjust the amount of routing of the input information.
Their experiments using stacked feed-forward networks for image classification also show consistent trends with ours -- the effect of preserving the original input is %
dominant especially in the later layers.
Inspired by this observation, \citet{liu-etal-2020-rethinking} modified the Transformer architecture to enhance the original input in the residual connections and demonstrated that this extension leads to better performance and convergence.
Note that several variants of the Transformer-based architecture with different arrangements of RES and LN have also been proposed~\cite{klein2018opennmt,xiong2020on,parisotto2020stabilizing}, and analyzing these models is one of our future works.

\section{Conclusions}
\label{sec:conclusions}
In this paper, we have extended a norm-based analysis to broaden the scope of analyzing Transformers from the multi-head attention alone to the whole attention block, i.e., multi-head attention, residual connection, and layer normalization.
Our analysis of the masked language models revealed that the context-mixing ratio in each block is much lower than expected in previous studies, demonstrating that RES and LN largely cancel the mixing by ATTN.
This observation can provide new explanations for %
some unexpected results were reported on Transformers in fields ranging from NLP to neuroscience (e.g., discarding the learned attention patterns did not adversely affect the performance).
Our detailed analysis further suggested that BERT discounts highly frequent, low-informative tokens.

Although our method is applicable to analyzing other variants of Transformers, our experiments were limited to the Transformer-based masked language models.
In addition, the Transformer is not composed of only the attention block; feed-forward and embedding layers also exist.
We plan to extend this work in both directions.

\clearpage
\section*{Acknowledgements}
We would like to thank the members of the Tohoku NLP Lab for their insightful comments, particularly Benjamin Heinzerling for his valuable suggestions on content and wording. 
This work was supported by JST CREST Grant Number JPMJCR20D2, Japan; JST ACT-X Grant Number JPMJAX200S, Japan; and JSPS KAKENHI Grant Number JP20J22697.

\section*{Ethical considerations}
One recent issue in the whole NLP community is that neural-network-based models have non-intended \emph{biases} (e.g., gender bias) induced during the training process.
This paper gives a method for interpreting the inner workings of real-world machine learning models, which may help us understand such biased behaviors of the models in the future.

\bibliography{custom} %
\bibliographystyle{acl_natbib}

\clearpage

\appendix

\section{Derivation of ``distributive law'' of LN}
\label{ap:precise_equation}
In Section~\ref{subsec:proposal_overview}, we introduced the ``distributive law'' for LN (layer normalization) in Equations~\ref{eq:distributional_law_ln} and \ref{eq:g_ln}.
Here, we show its derivation.
Let $\bm z = \sum_j \bm z_j$ be the input to LN.
Then, Equations~\ref{eq:distributional_law_ln} and \ref{eq:g_ln} are derived as follows:
\begin{align}
    \mathrm{LN}(\bm z) &= \frac{\bm z  - \frac{1}{d} \sum_k \bm z^{(k)}}{s(\bm z)}\odot \bm \gamma + \bm \beta \\
    &= \frac{\sum_j \bm z_j  - \frac{1}{d} \sum_k \bigl( \sum_j \bm z_j \bigr)^{(k)} }{s( \bm z )}\odot \bm \gamma + \bm \beta \\
    &= \sum_j \frac{\bm z_j  - \frac{1}{d} \sum_k \bm z_j^{(k)}}{s( \bm z )}\odot \bm \gamma + \bm \beta \\
    &= \sum_j \frac{\bm z_j  - m(\bm z_j)}{s( \bm z )}\odot \bm \gamma + \bm \beta \\
    &= \sum_j g_{\bm z}(\bm z_j) + \bm \beta \text{.}
\end{align}

\section{Mixing ratio in other settings}
\label{ap:mix_ratio_variant}
\begin{figure*}[t]
    \centering
    \begin{minipage}[t]{.19\hsize}
        \centering
        \includegraphics[height=10cm]{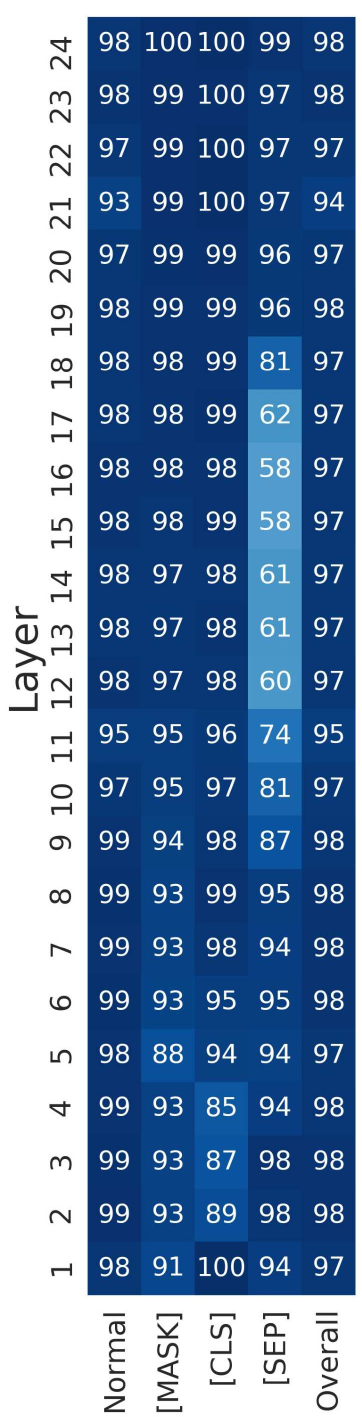}
        \subcaption{
        \textsc{Attn-w}.
        }
    \end{minipage}
    \;
    \begin{minipage}[t]{.17\hsize}
        \centering
        \includegraphics[height=10cm]{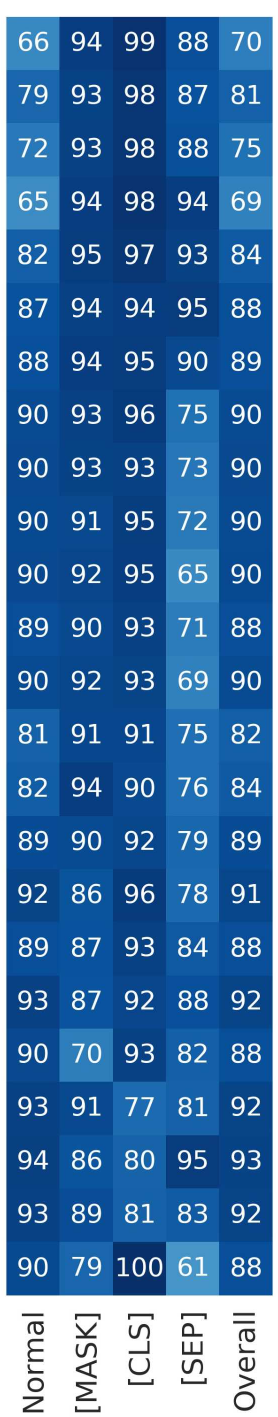}
        \subcaption{
        \textsc{Attn-n} \cite{kobayashi-etal-2020-attention}.
        }
    \end{minipage}
    \;
    \begin{minipage}[t]{.17\hsize}
        \centering
        \includegraphics[height=10cm]{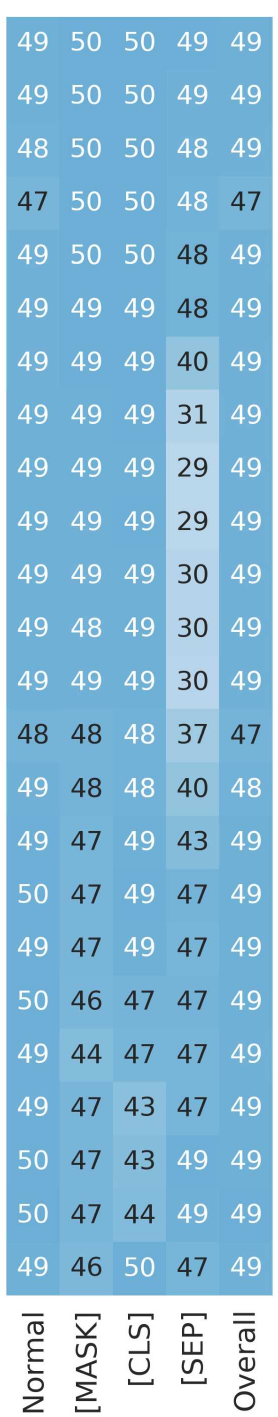}
        \subcaption{
        \textsc{AttnRes-w} \cite{abnar-zuidema-2020-quantifying}.
        }
    \end{minipage}
    \;
    \begin{minipage}[t]{.17\hsize}
    \centering
    \includegraphics[height=10cm]{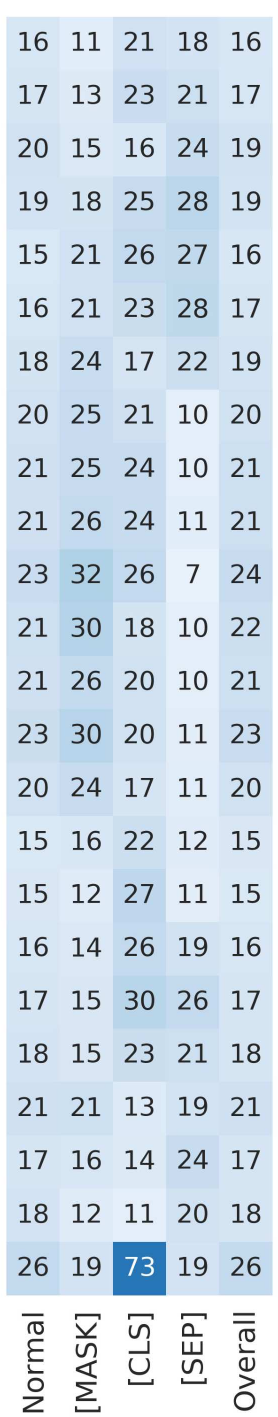}
    \subcaption{
    \textsc{AttnRes-n}.
    }
    \end{minipage}
    \;
    \begin{minipage}[t]{.22\hsize}
    \centering
    \includegraphics[height=10cm]{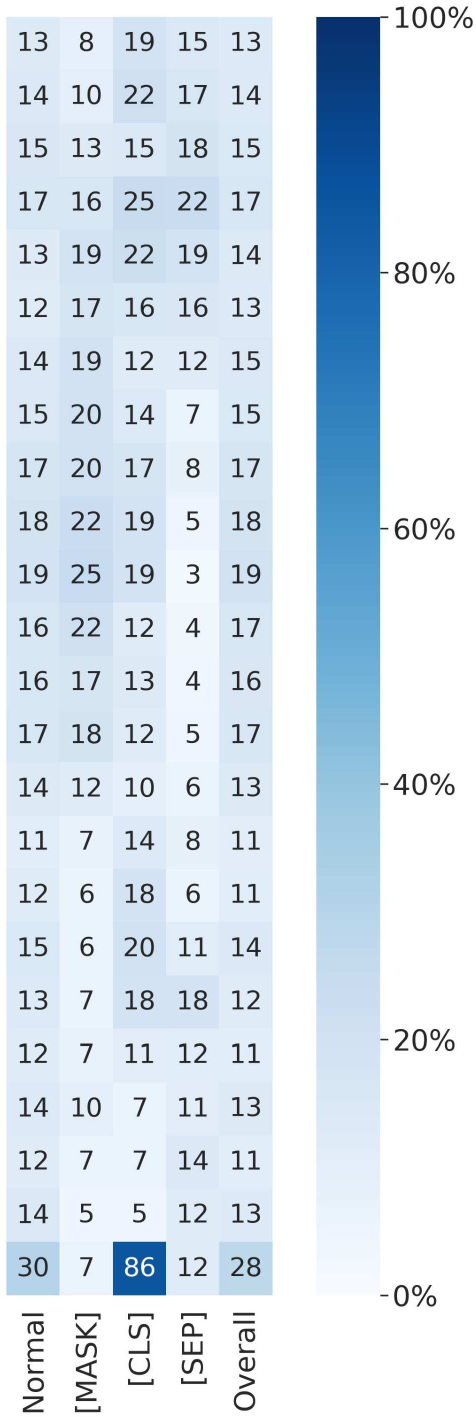}
    \subcaption{
    \textsc{AttnResLn-n}.
    }
    \end{minipage}
    \caption{
    Mixing ratio at each layer of BERT-large calculated from each method.
    }
    \label{fig:mixing_rate_large}
\end{figure*}

\begin{figure*}[t]
    \centering
    \begin{minipage}[t]{.19\hsize}
        \centering
        \includegraphics[height=4.5cm]{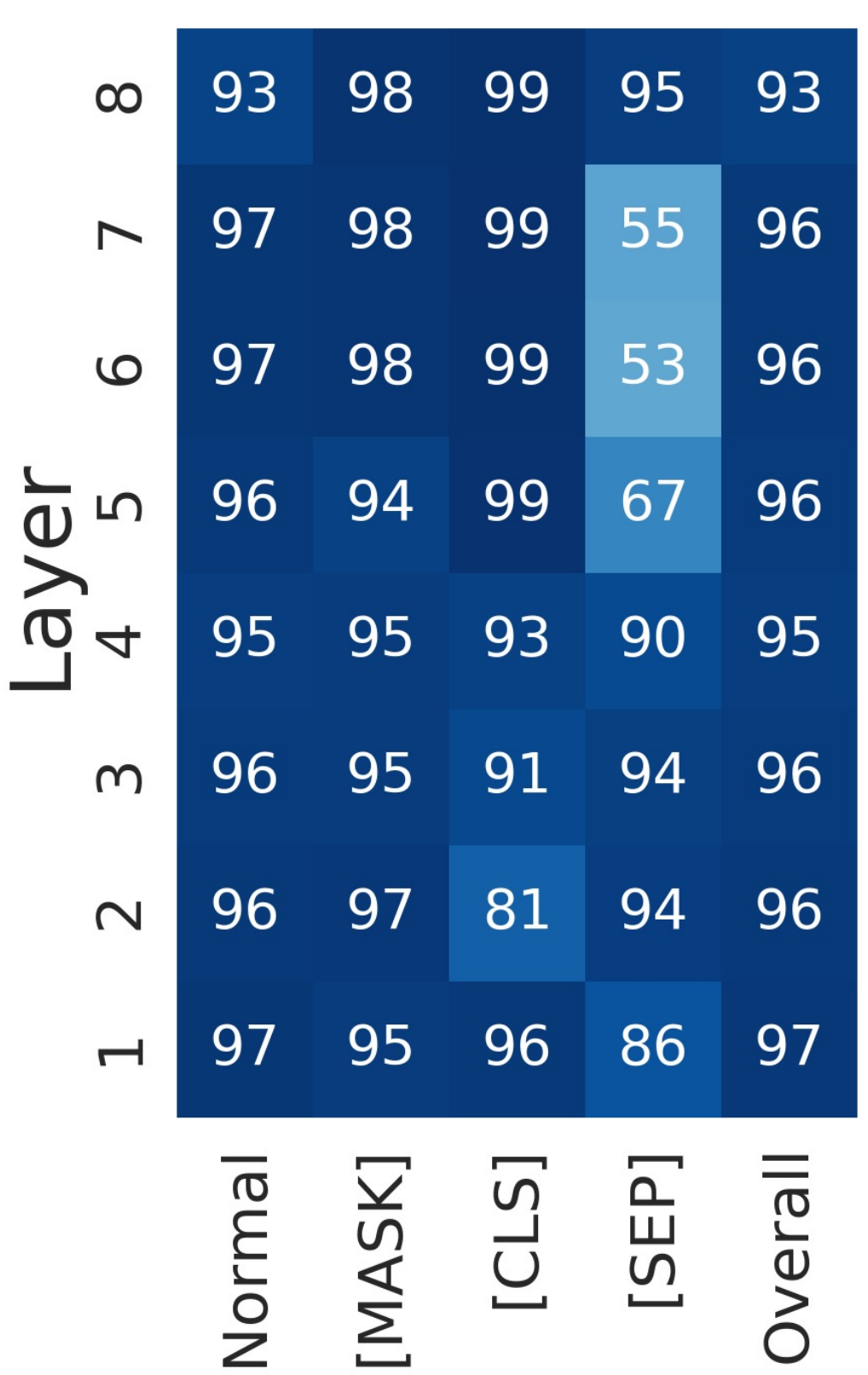}
        \subcaption{
        \textsc{Attn-w}.
        }
    \end{minipage}
    \;
    \begin{minipage}[t]{.17\hsize}
        \centering
        \includegraphics[height=4.5cm]{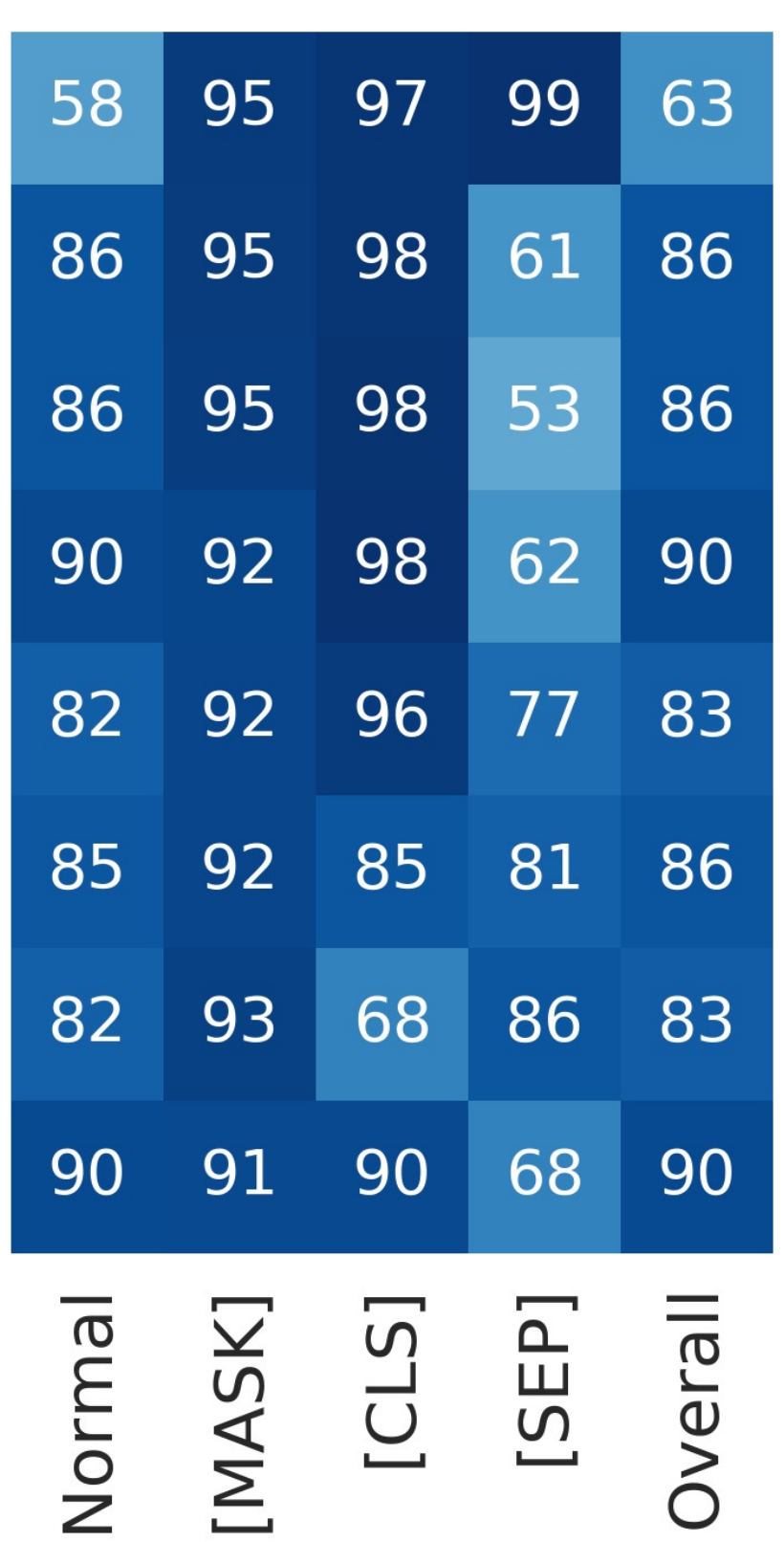}
        \subcaption{
        \textsc{Attn-n} \cite{kobayashi-etal-2020-attention}.
        }
    \end{minipage}
    \;
    \begin{minipage}[t]{.17\hsize}
        \centering
        \includegraphics[height=4.5cm]{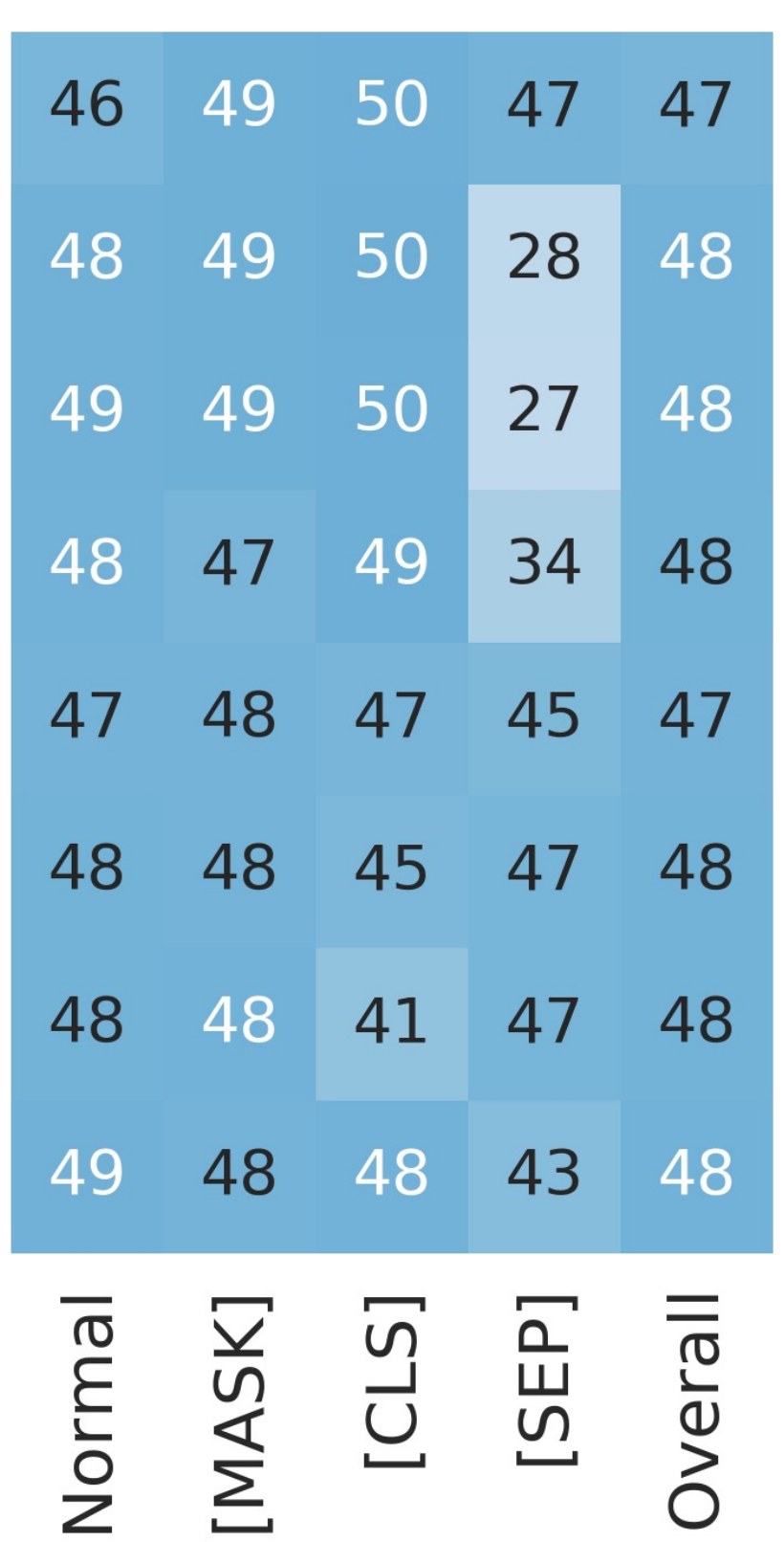}
        \subcaption{
        \textsc{AttnRes-w} \cite{abnar-zuidema-2020-quantifying}.
        }
    \end{minipage}
    \;
    \begin{minipage}[t]{.17\hsize}
    \centering
    \includegraphics[height=4.5cm]{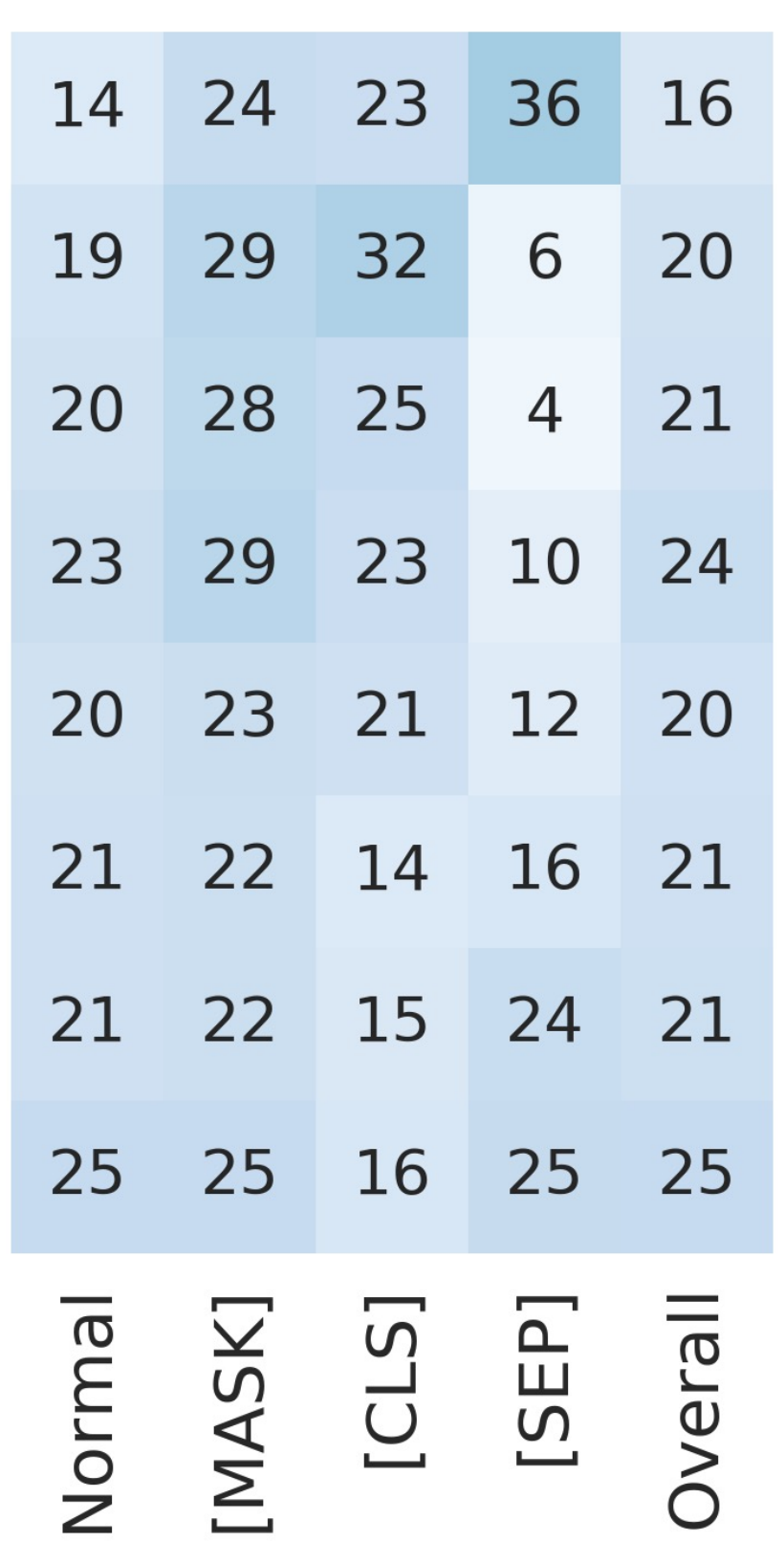}
    \subcaption{
    \textsc{AttnRes-n}.
    }
    \end{minipage}
    \;
    \begin{minipage}[t]{.22\hsize}
    \centering
    \includegraphics[height=4.5cm]{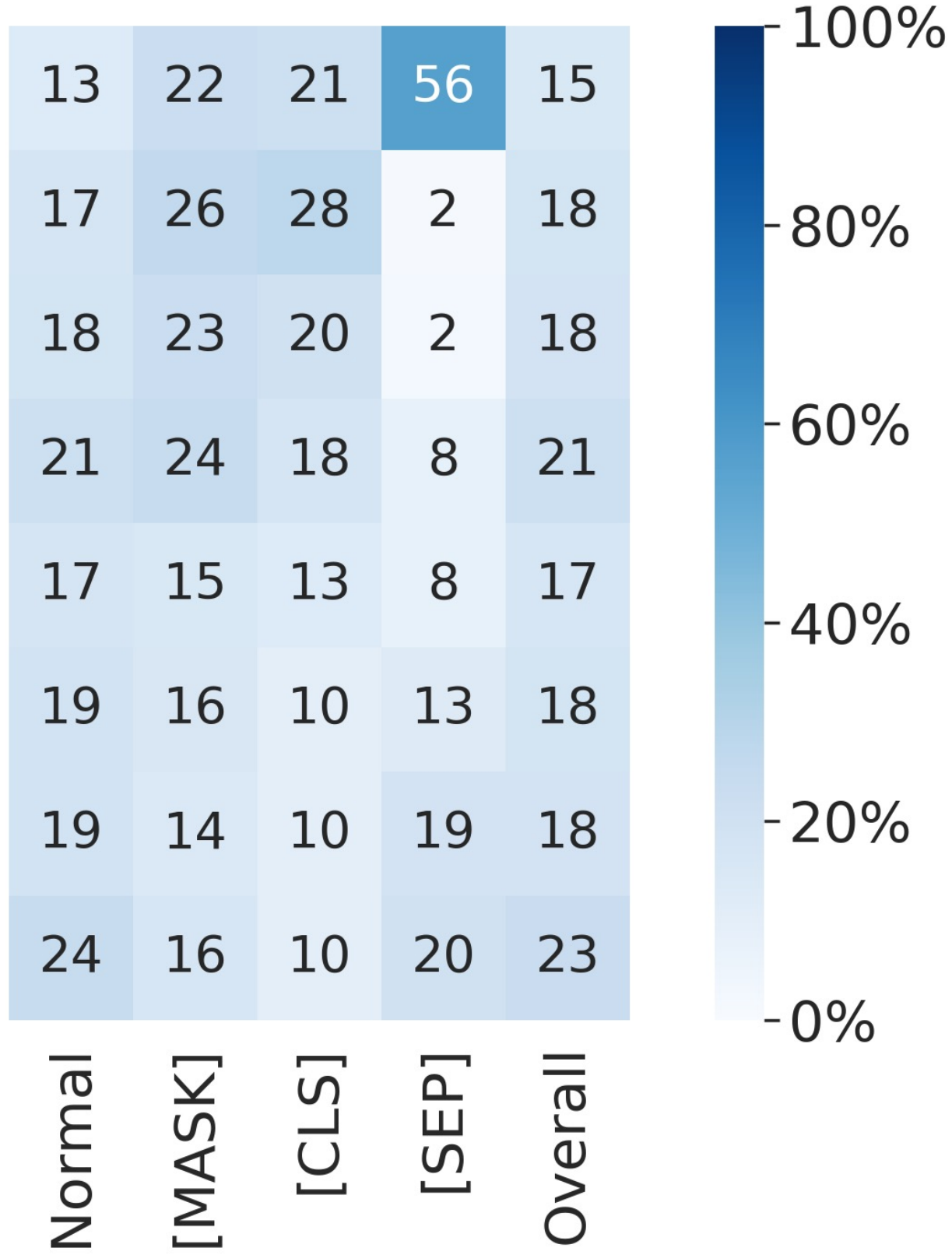}
    \subcaption{
    \textsc{AttnResLn-n}.
    }
    \end{minipage}
    \caption{
    Mixing ratio at each layer of BERT-medium calculated from each method.
    }
    \label{fig:mixing_rate_medium}
\end{figure*}

\begin{figure*}[t]
    \centering
    \begin{minipage}[t]{.19\hsize}
        \centering
        \includegraphics[height=3cm]{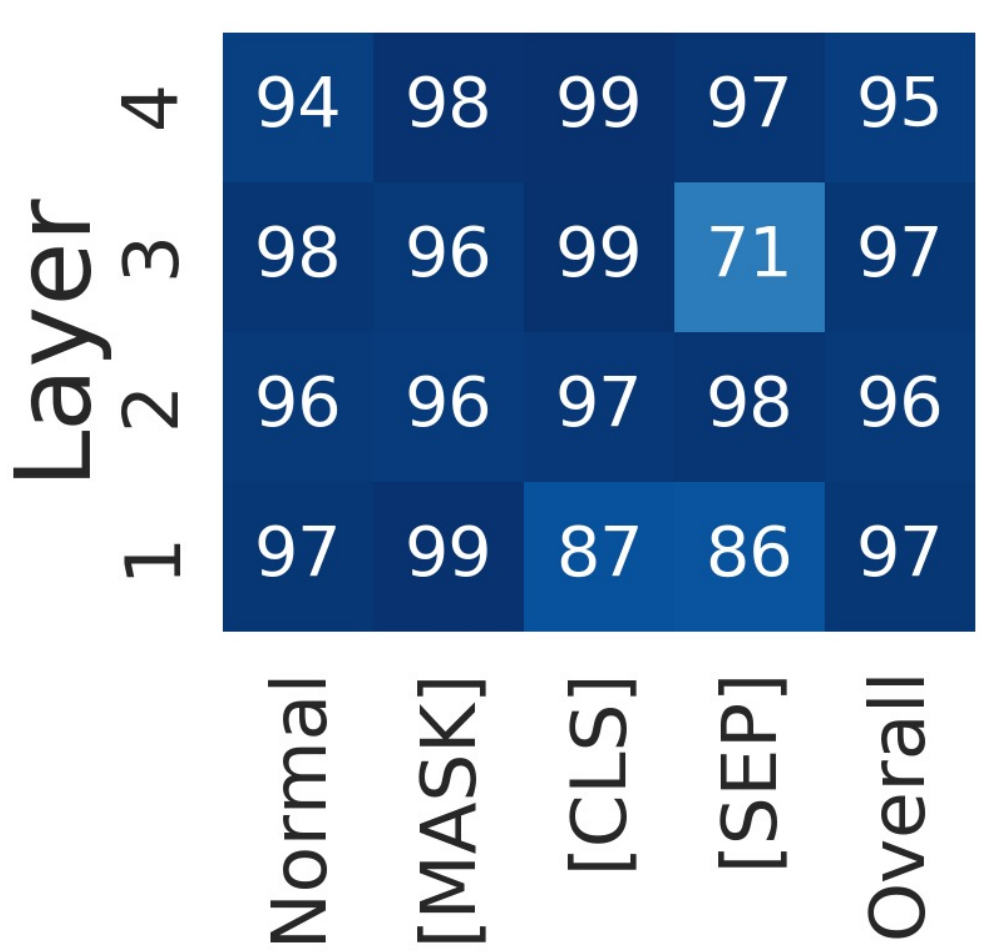}
        \subcaption{
        \textsc{Attn-w}.
        }
    \end{minipage}
    \;
    \begin{minipage}[t]{.17\hsize}
        \centering
        \includegraphics[height=3cm]{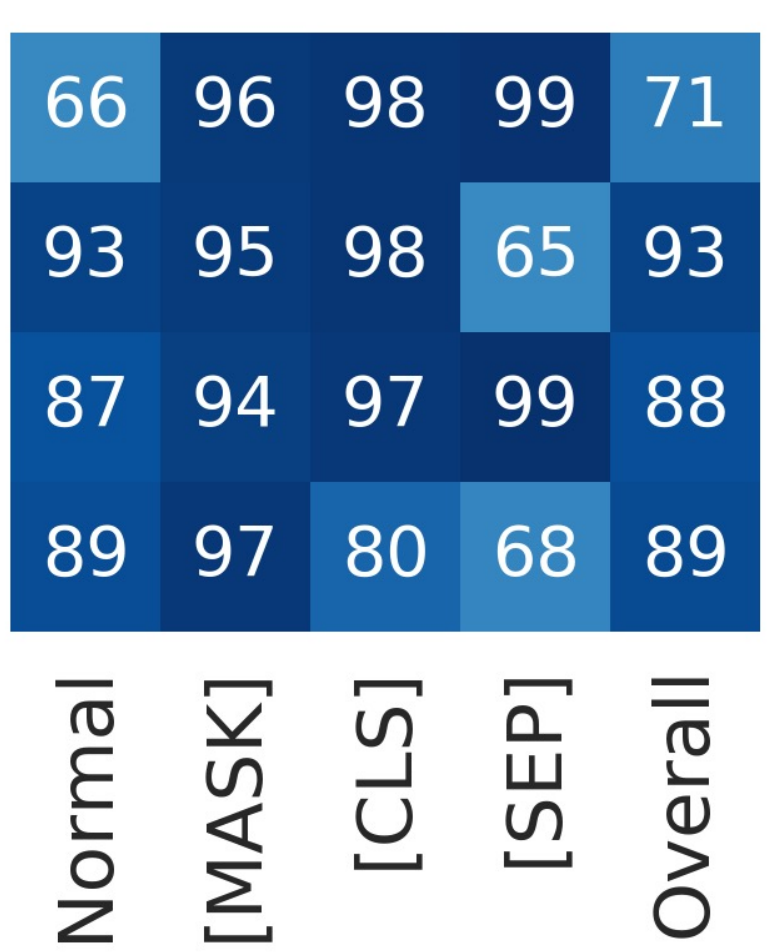}
        \subcaption{
        \textsc{Attn-n} \cite{kobayashi-etal-2020-attention}.
        }
    \end{minipage}
    \;
    \begin{minipage}[t]{.17\hsize}
        \centering
        \includegraphics[height=3cm]{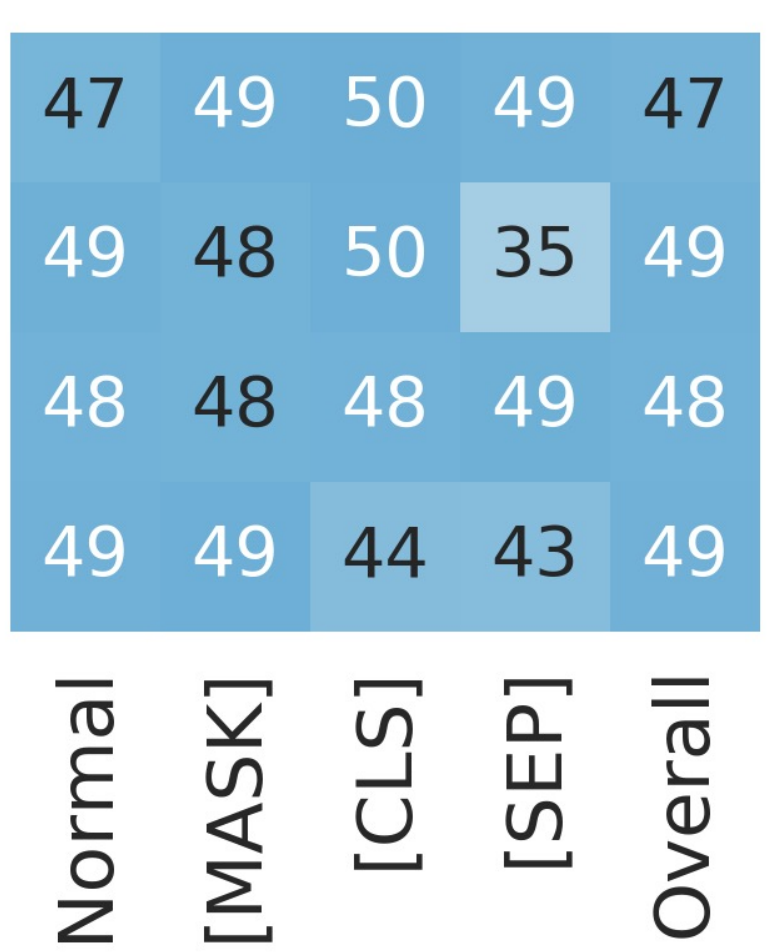}
        \subcaption{
        \textsc{AttnRes-w} \cite{abnar-zuidema-2020-quantifying}.
        }
    \end{minipage}
    \;
    \begin{minipage}[t]{.17\hsize}
    \centering
    \includegraphics[height=3cm]{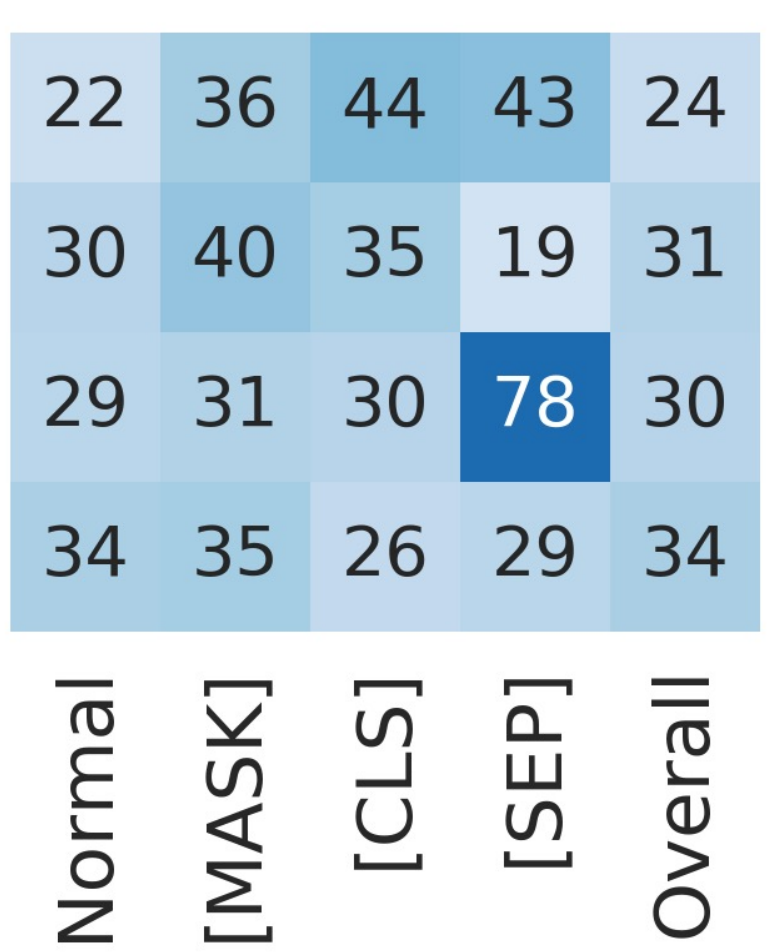}
    \subcaption{
    \textsc{AttnRes-n}.
    }
    \end{minipage}
    \;
    \begin{minipage}[t]{.22\hsize}
    \centering
    \includegraphics[height=3cm]{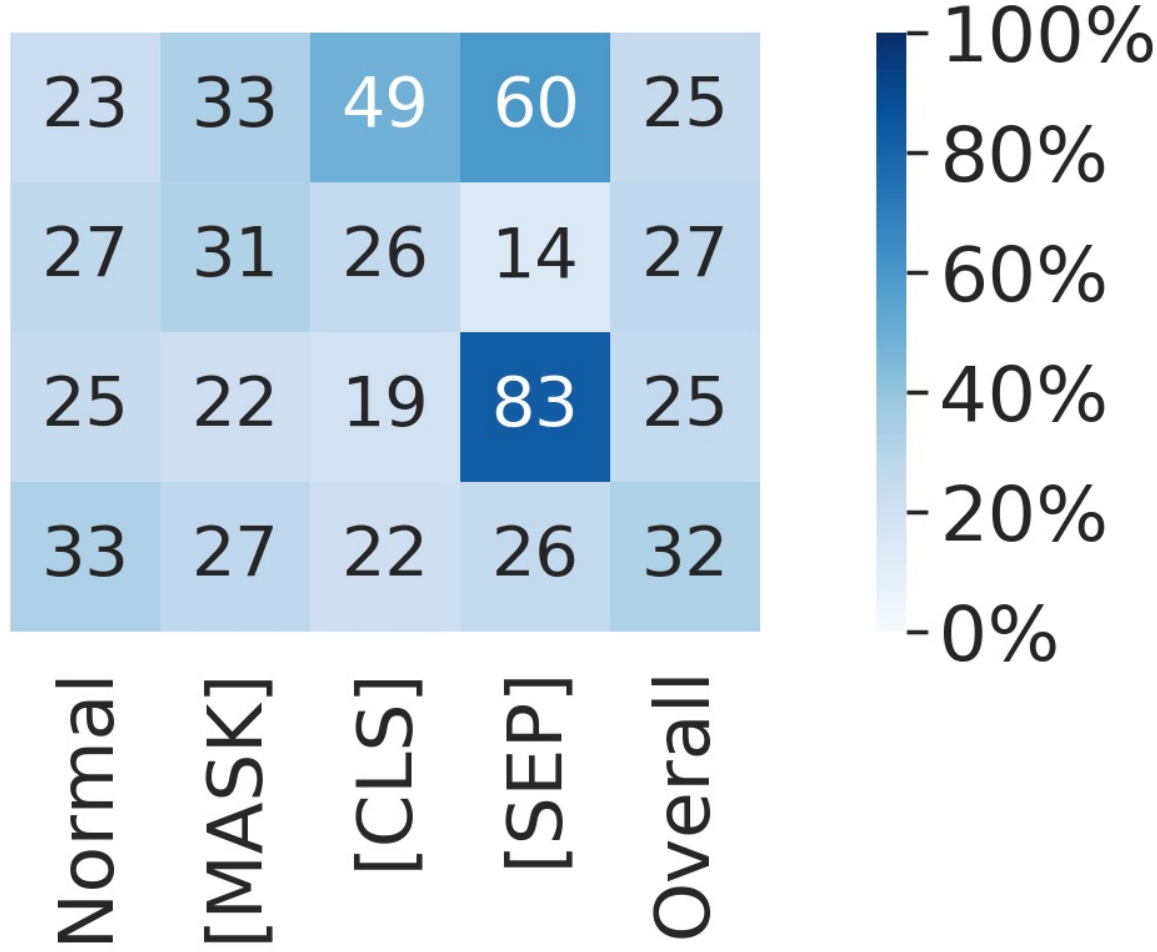}
    \subcaption{
    \textsc{AttnResLn-n}.
    }
    \end{minipage}
    \caption{
    Mixing ratio at each layer of BERT-small calculated from each method.
    }
    \label{fig:mixing_rate_small}
\end{figure*}

\begin{figure*}[t]
    \centering
    \begin{minipage}[t]{.19\hsize}
        \centering
        \includegraphics[height=3cm]{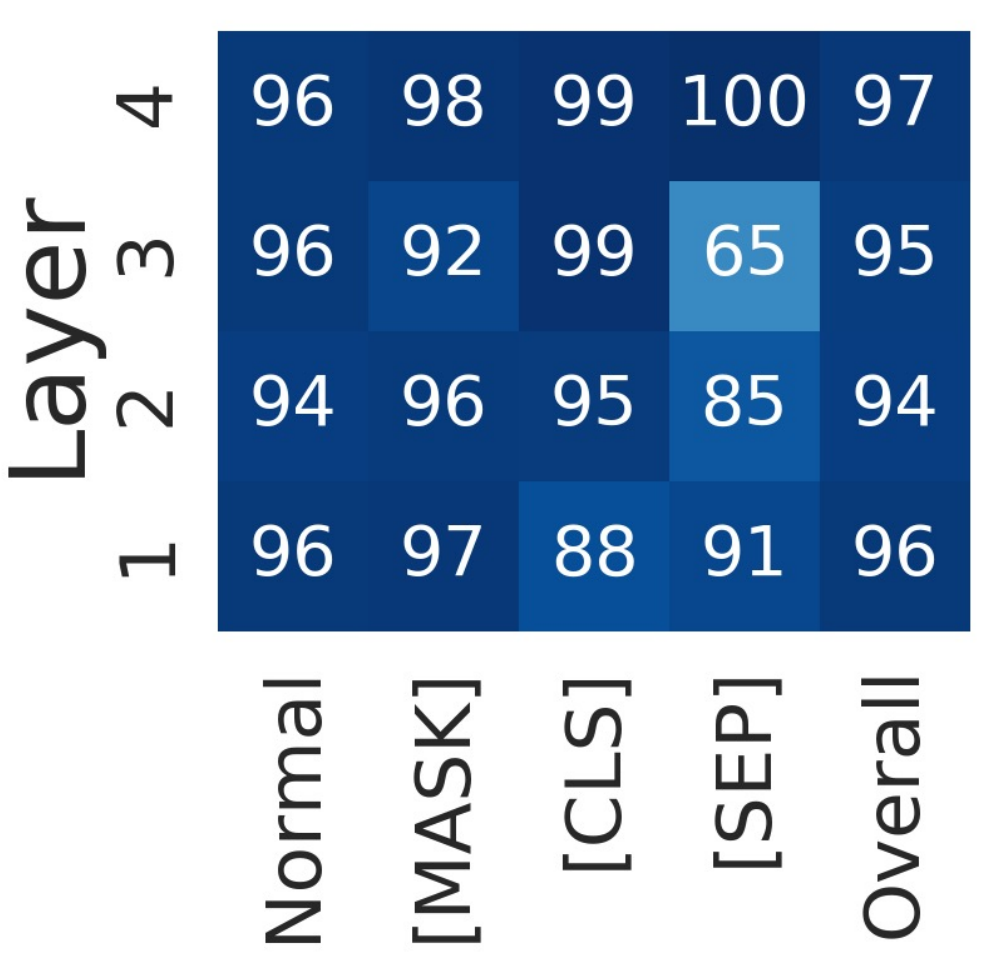}
        \subcaption{
        \textsc{Attn-w}.
        }
    \end{minipage}
    \;
    \begin{minipage}[t]{.17\hsize}
        \centering
        \includegraphics[height=3cm]{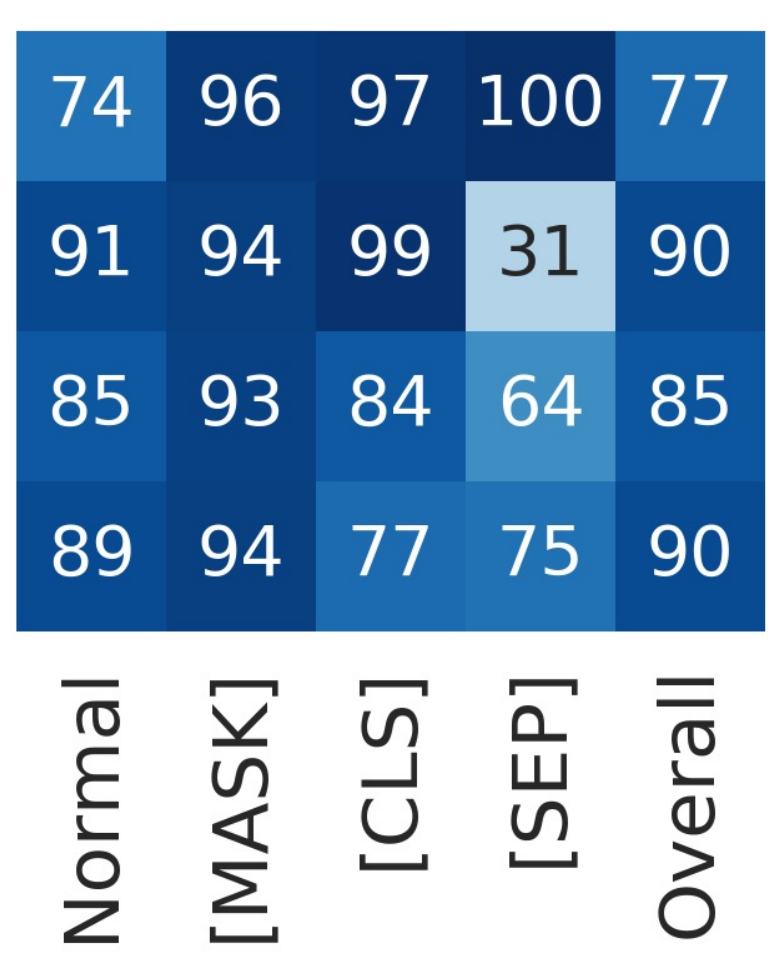}
        \subcaption{
        \textsc{Attn-n} \cite{kobayashi-etal-2020-attention}.
        }
    \end{minipage}
    \;
    \begin{minipage}[t]{.17\hsize}
        \centering
        \includegraphics[height=3cm]{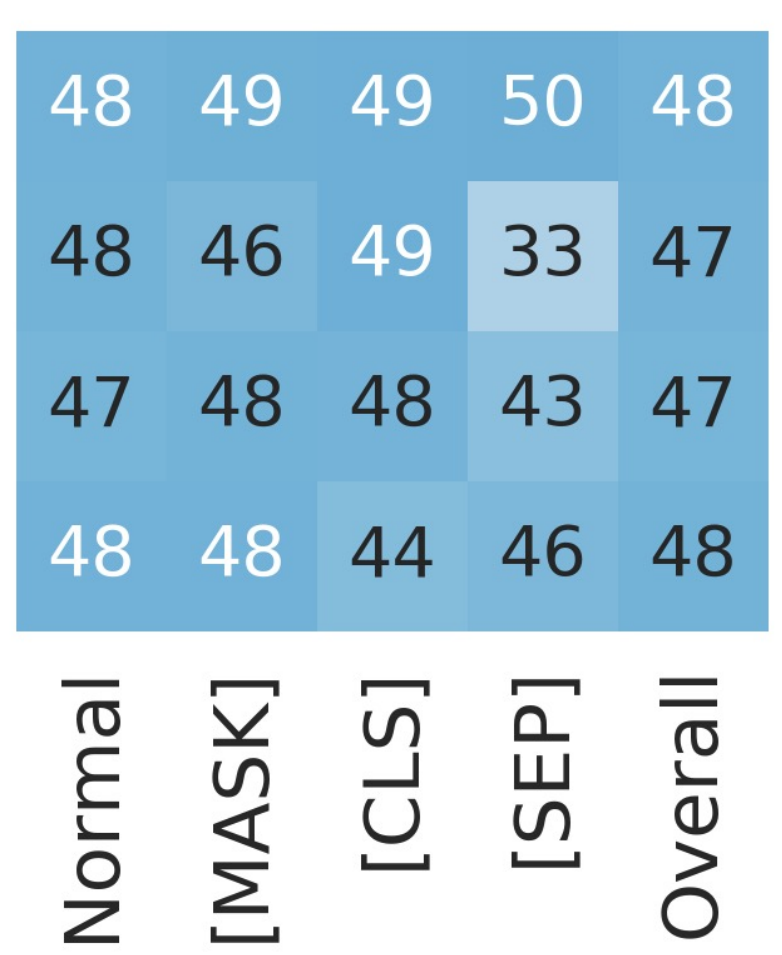}
        \subcaption{
        \textsc{AttnRes-w} \cite{abnar-zuidema-2020-quantifying}.
        }
    \end{minipage}
    \;
    \begin{minipage}[t]{.17\hsize}
    \centering
    \includegraphics[height=3cm]{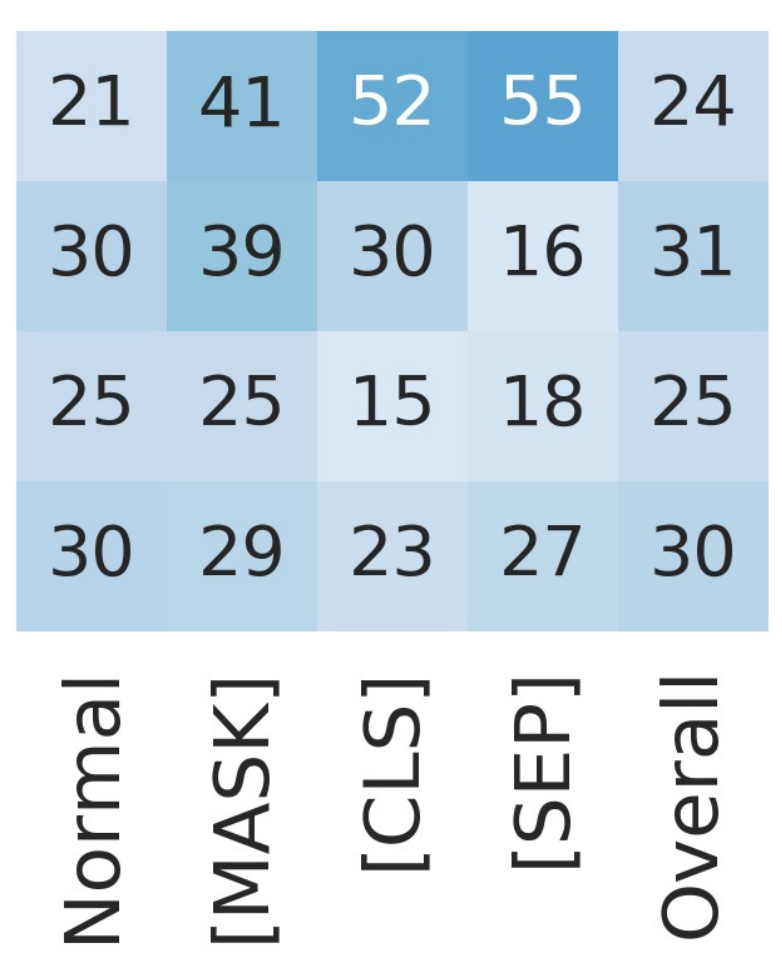}
    \subcaption{
    \textsc{AttnRes-n}.
    }
    \end{minipage}
    \;
    \begin{minipage}[t]{.22\hsize}
    \centering
    \includegraphics[height=3cm]{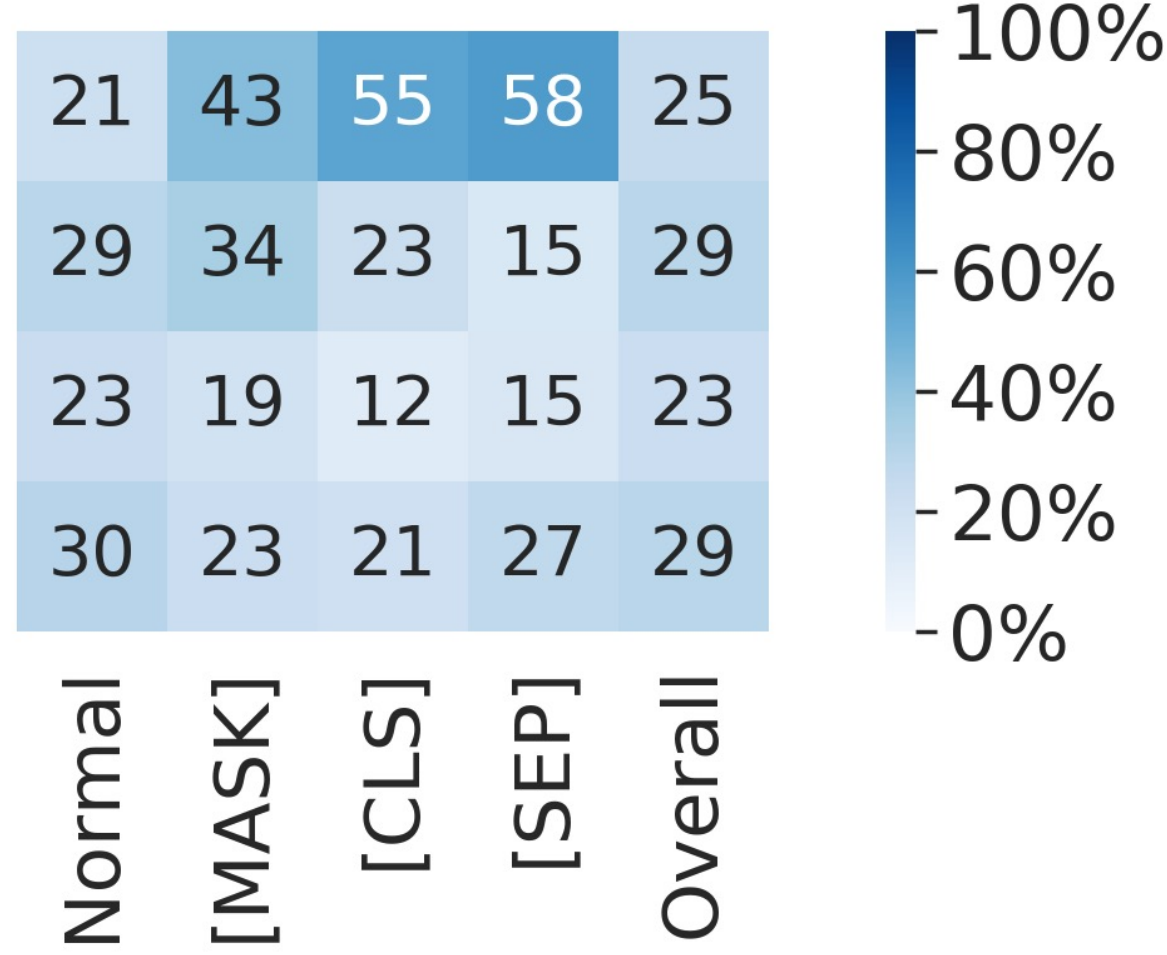}
    \subcaption{
    \textsc{AttnResLn-n}.
    }
    \end{minipage}
    \caption{
    Mixing ratio at each layer of BERT-mini calculated from each method.
    }
    \label{fig:mixing_rate_mini}
\end{figure*}

\begin{figure*}[t]
    \centering
    \begin{minipage}[t]{.19\hsize}
        \centering
        \includegraphics[height=2cm]{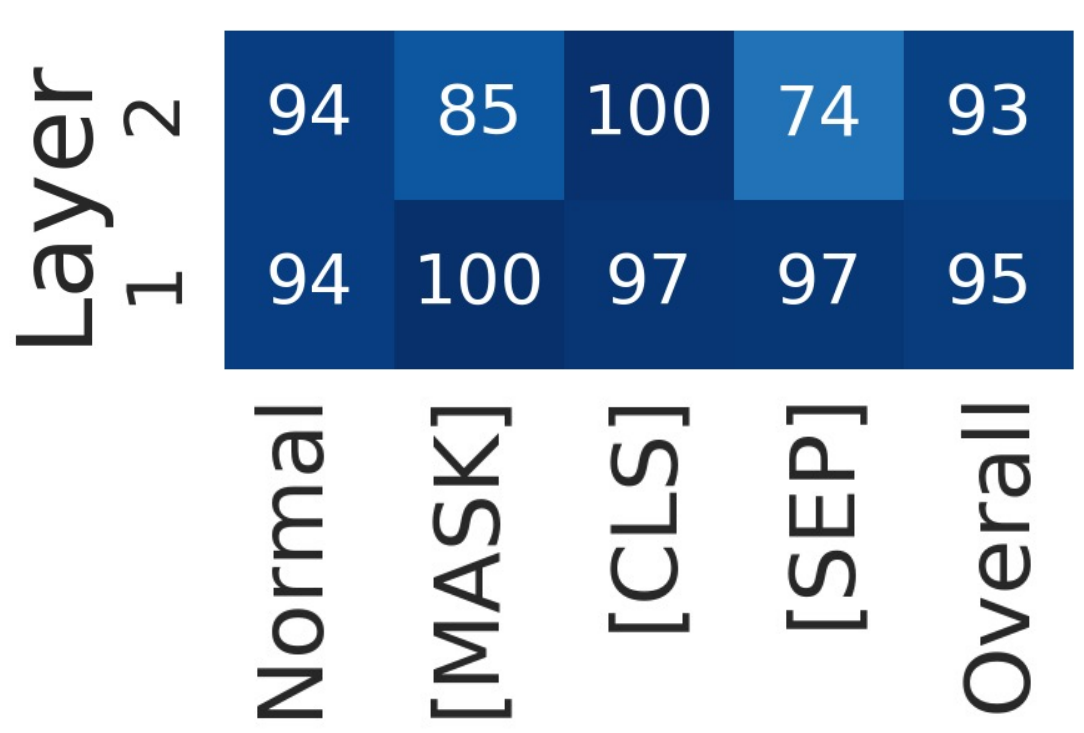}
        \subcaption{
        \textsc{Attn-w}.
        }
    \end{minipage}
    \;
    \begin{minipage}[t]{.17\hsize}
        \centering
        \includegraphics[height=2cm]{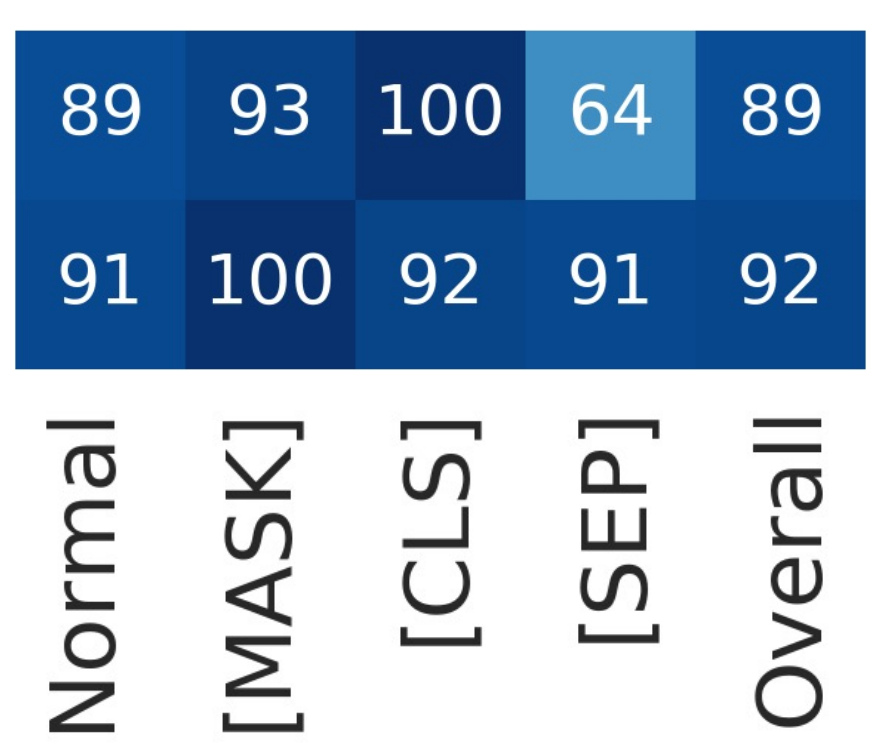}
        \subcaption{
        \textsc{Attn-n} \cite{kobayashi-etal-2020-attention}.
        }
    \end{minipage}
    \;
    \begin{minipage}[t]{.17\hsize}
        \centering
        \includegraphics[height=2cm]{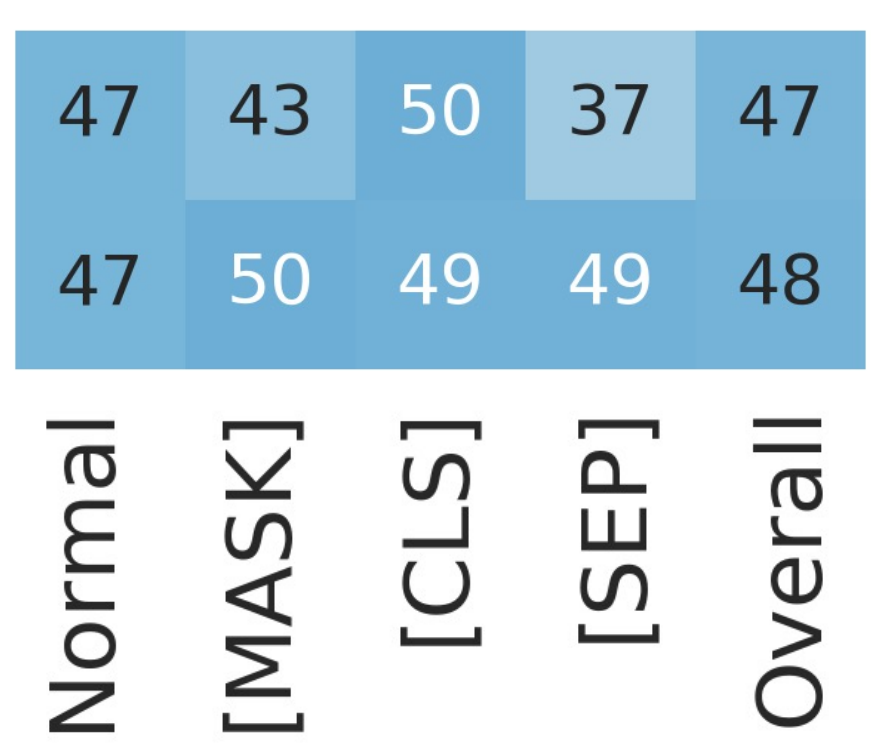}
        \subcaption{
        \textsc{AttnRes-w} \cite{abnar-zuidema-2020-quantifying}.
        }
    \end{minipage}
    \;
    \begin{minipage}[t]{.17\hsize}
    \centering
    \includegraphics[height=2cm]{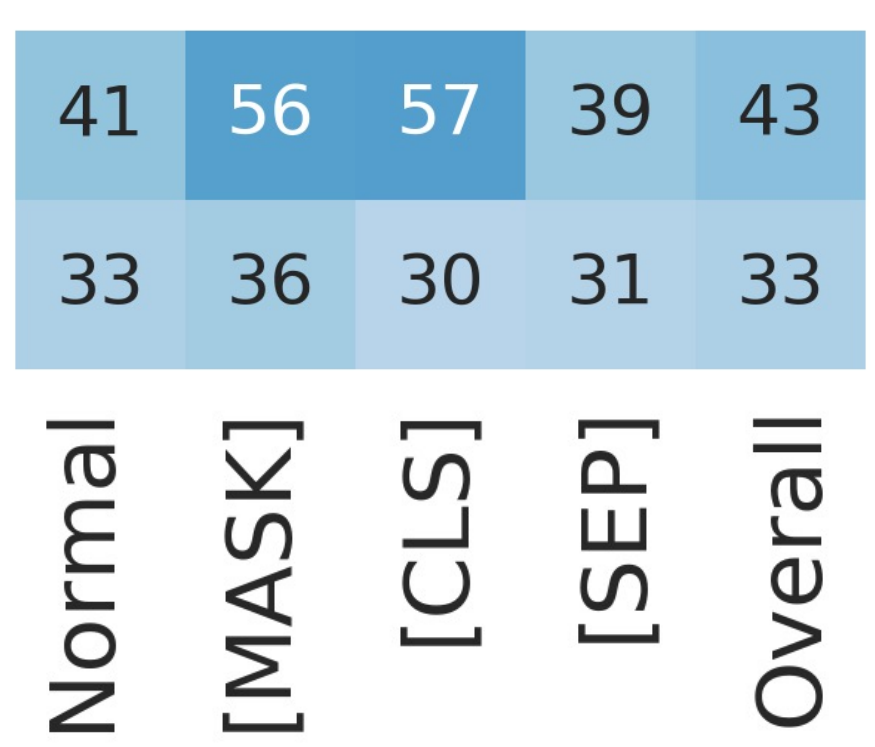}
    \subcaption{
    \textsc{AttnRes-n}.
    }
    \end{minipage}
    \;
    \begin{minipage}[t]{.22\hsize}
    \centering
    \includegraphics[height=2cm]{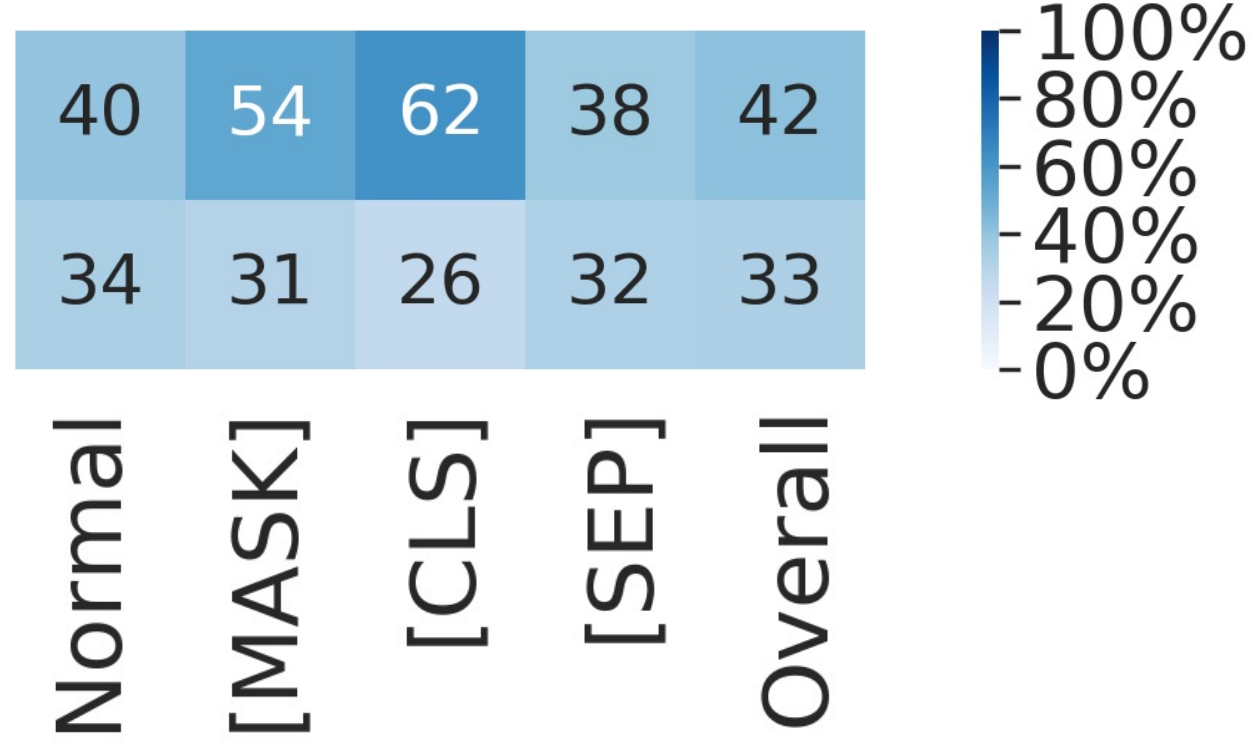}
    \subcaption{
    \textsc{AttnResLn-n}.
    }
    \end{minipage}
    \caption{
    Mixing ratio at each layer of BERT-tiny calculated from each method.
    }
    \label{fig:mixing_rate_tiny}
\end{figure*}

\begin{figure*}[t]
    \centering
    \begin{minipage}[t]{.19\hsize}
        \centering
        \includegraphics[height=6cm]{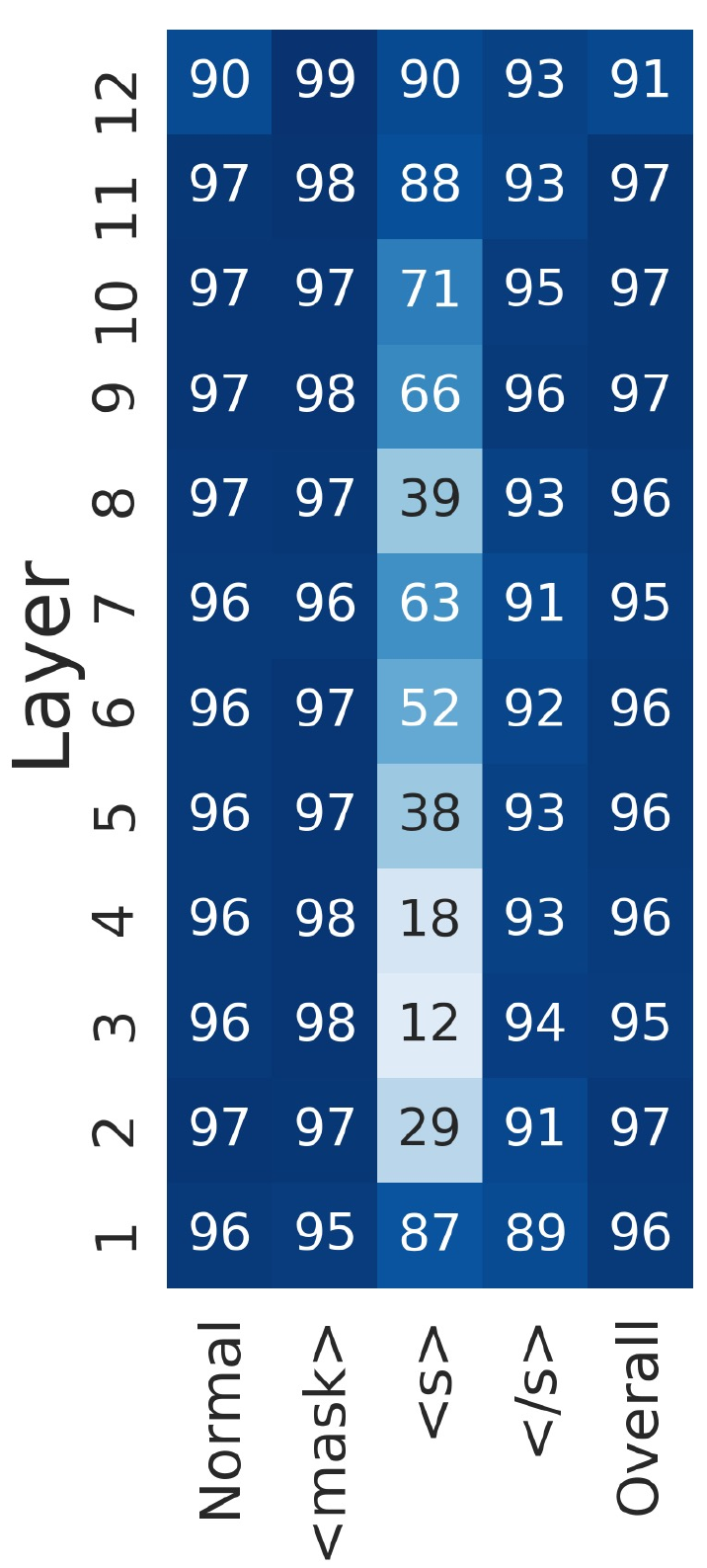}
        \subcaption{
        \textsc{Attn-w}.
        }
        \label{fig:roberta_weight_rate}
    \end{minipage}
    \;
    \begin{minipage}[t]{.17\hsize}
        \centering
        \includegraphics[height=6cm]{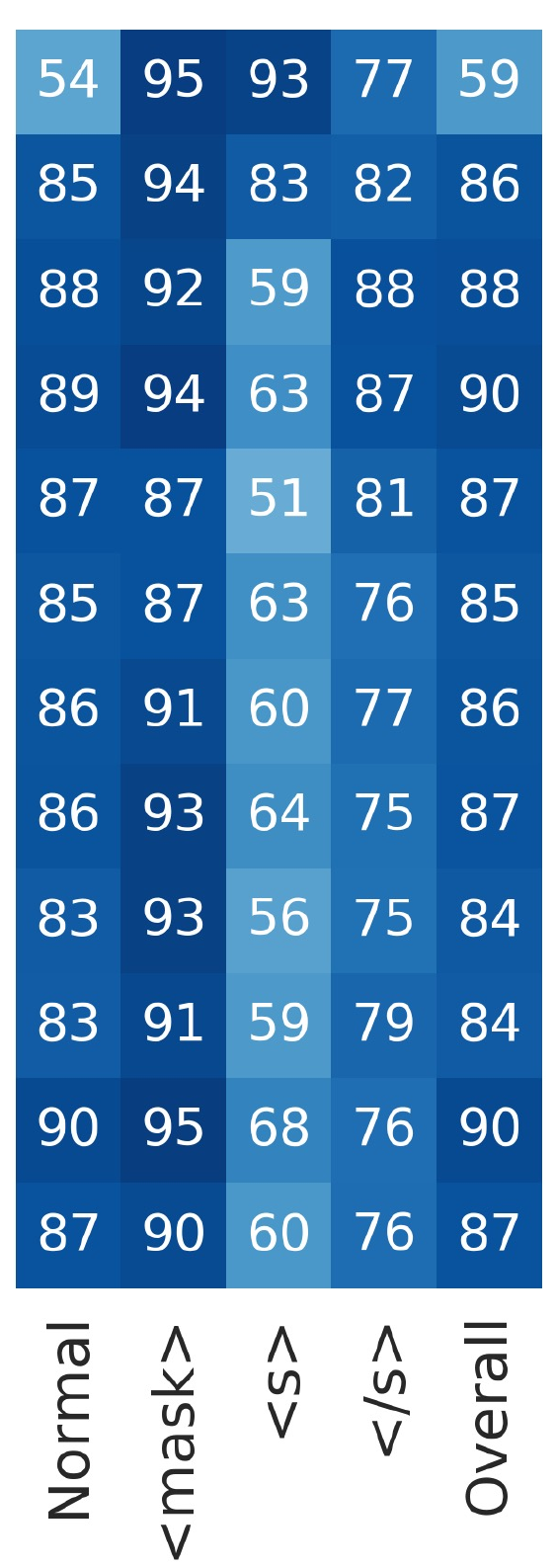}
        \subcaption{
        \textsc{Attn-n} \cite{kobayashi-etal-2020-attention}.
        }
        \label{fig:roberta_kobayashi_rate}
    \end{minipage}
    \;
    \begin{minipage}[t]{.17\hsize}
        \centering
        \includegraphics[height=6cm]{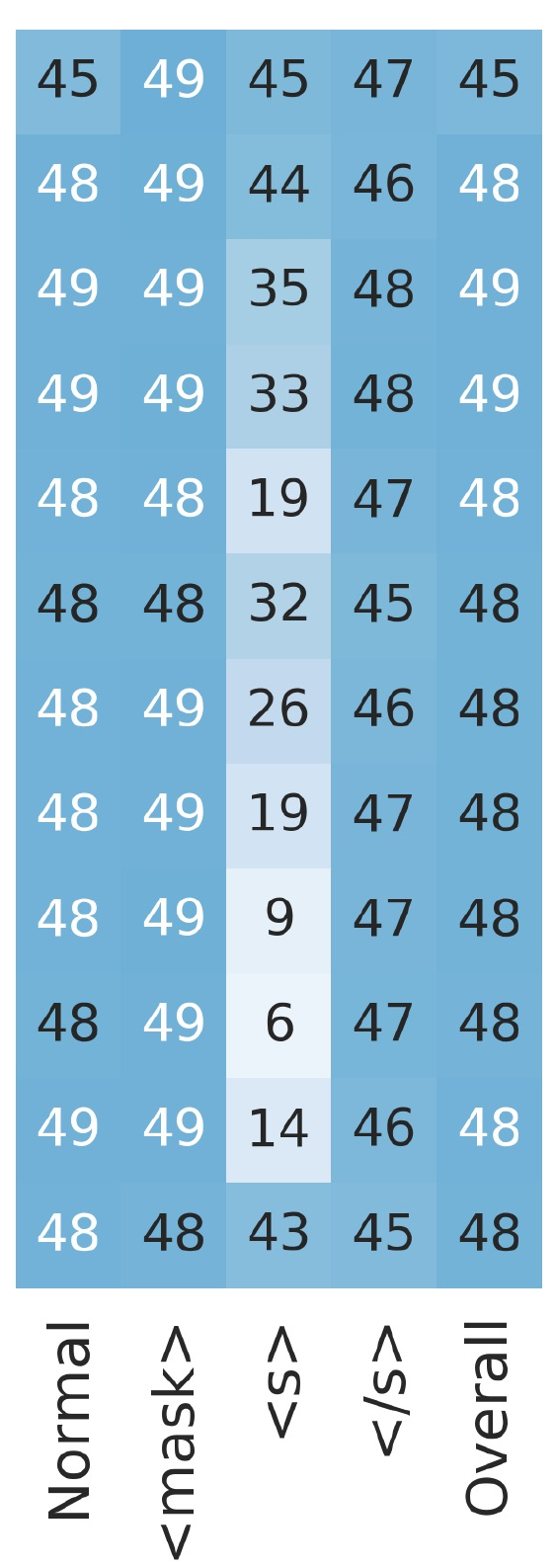}
        \subcaption{
        \textsc{AttnRes-w} \cite{abnar-zuidema-2020-quantifying}.
        }
        \label{fig:roberta_abnar_rate}
    \end{minipage}
    \;
    \begin{minipage}[t]{.17\hsize}
    \centering
    \includegraphics[height=6cm]{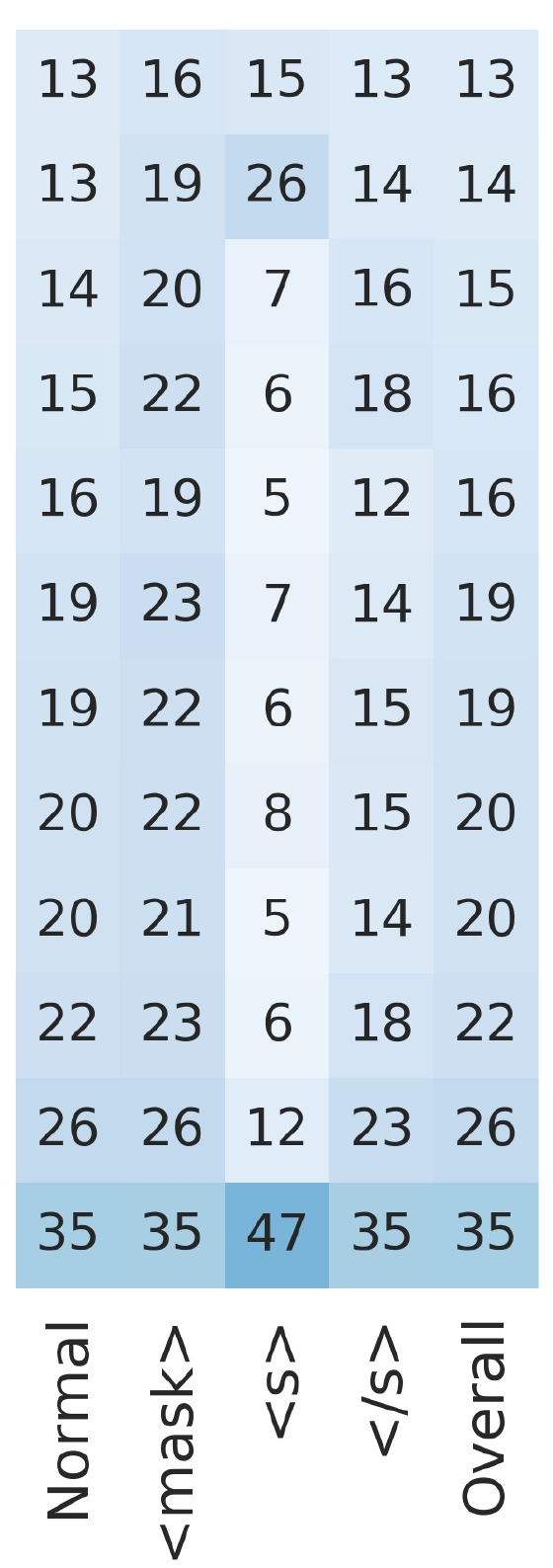}
    \subcaption{
    \textsc{AttnRes-n}.
    }
    \label{fig:roberta_before_ln_mixing_rate}
    \end{minipage}
    \;
    \begin{minipage}[t]{.22\hsize}
    \centering
    \includegraphics[height=6cm]{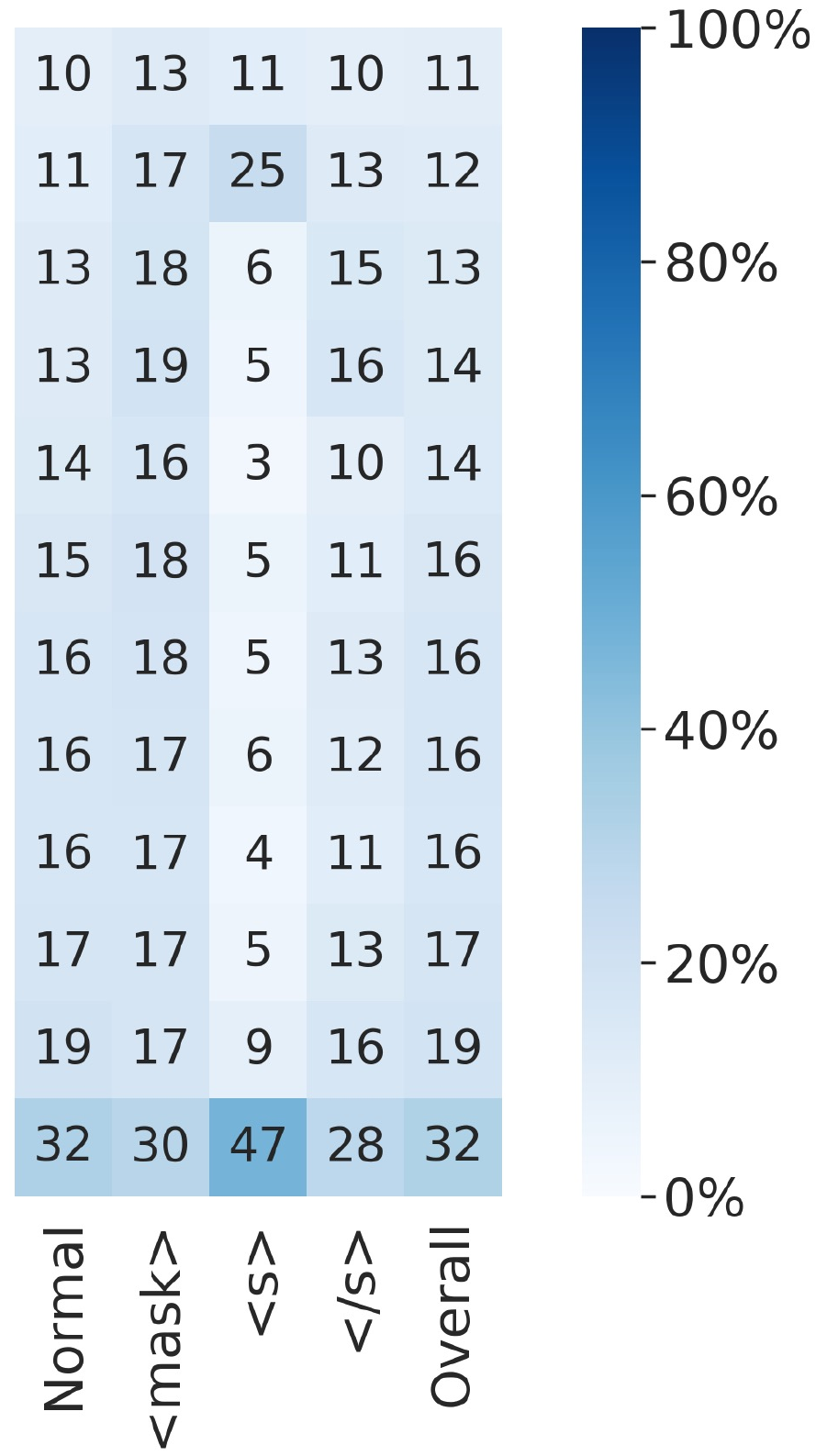}
    \subcaption{
    \textsc{AttnResLn-n}.
    }
    \label{fig:roberta_mix_rate}
    \end{minipage}
    \caption{
    Mixing ratio at each layer of RoBERTa-base calculated from each method.
    }
    \label{fig:mixing_rate_roberta}
\end{figure*}

\begin{figure*}[t]
    \centering
    \begin{minipage}[t]{.19\hsize}
        \centering
        \includegraphics[height=9cm]{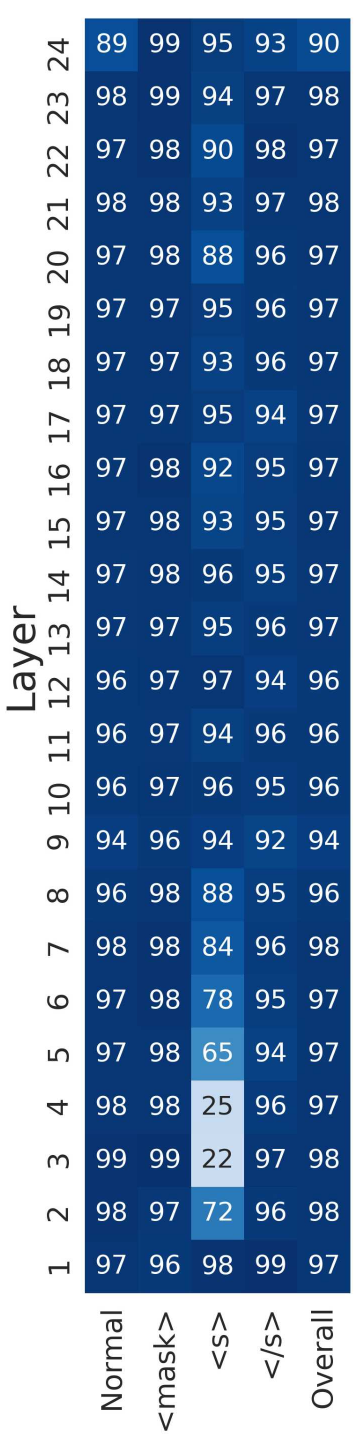}
        \subcaption{
        \textsc{Attn-w}.
        }
    \end{minipage}
    \;
    \begin{minipage}[t]{.17\hsize}
        \centering
        \includegraphics[height=9cm]{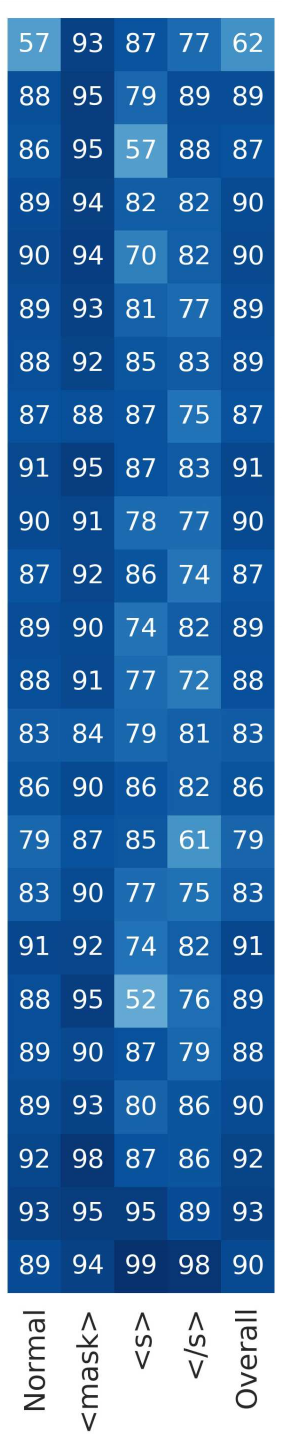}
        \subcaption{
        \textsc{Attn-n} \cite{kobayashi-etal-2020-attention}.
        }
    \end{minipage}
    \;
    \begin{minipage}[t]{.17\hsize}
        \centering
        \includegraphics[height=9cm]{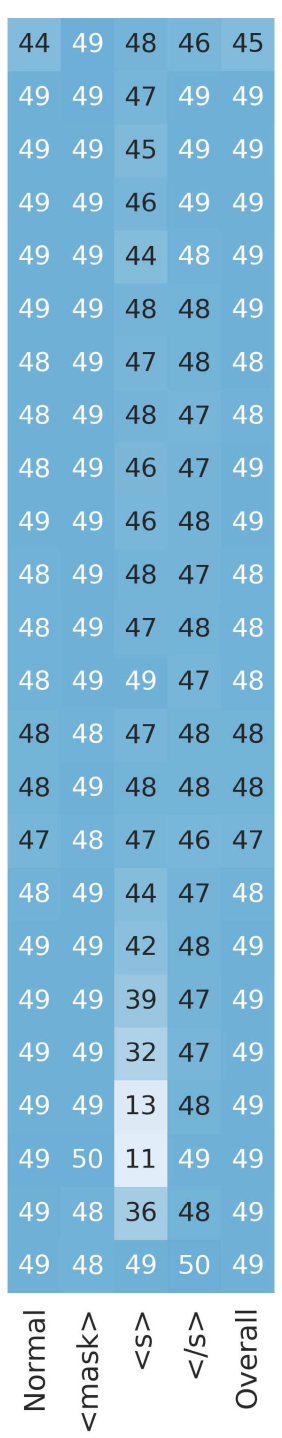}
        \subcaption{
        \textsc{AttnRes-w} \cite{abnar-zuidema-2020-quantifying}.
        }
    \end{minipage}
    \;
    \begin{minipage}[t]{.17\hsize}
    \centering
    \includegraphics[height=9cm]{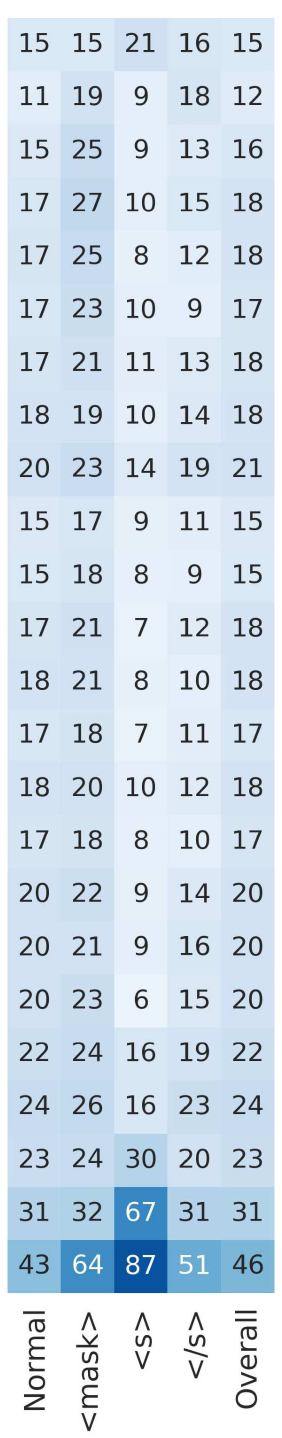}
    \subcaption{
    \textsc{AttnRes-n}.
    }
    \end{minipage}
    \;
    \begin{minipage}[t]{.22\hsize}
    \centering
    \includegraphics[height=9cm]{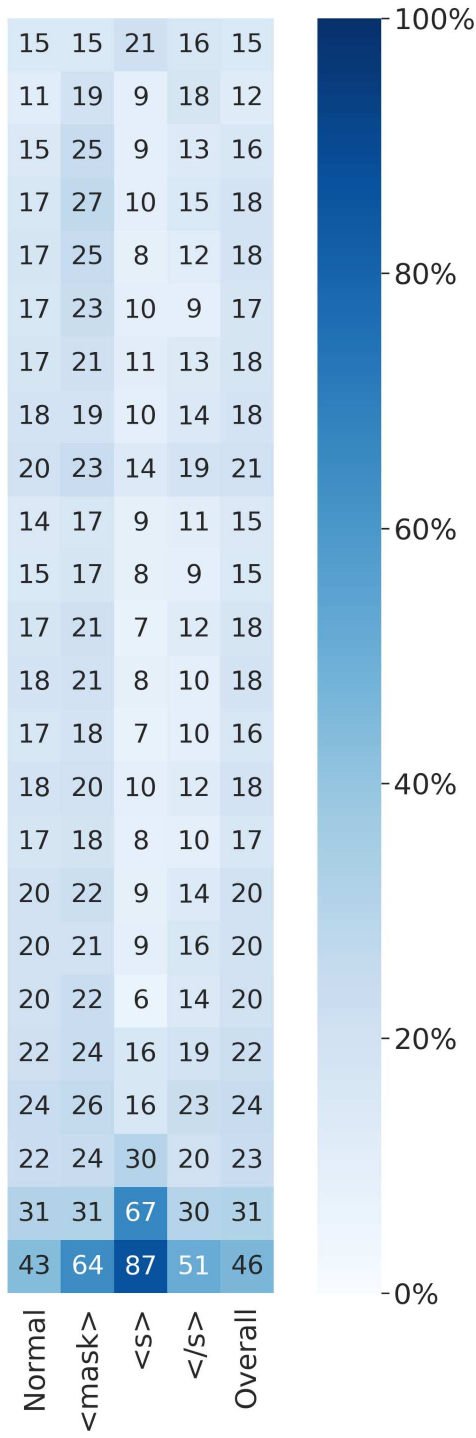}
    \subcaption{
    \textsc{AttnResLn-n}.
    }
    \end{minipage}
    \caption{
    Mixing ratio at each layer of RoBERTa-large calculated from each method.
    }
    \label{fig:roberta_mixing_rate_large}
\end{figure*}

In Sections~\ref{sec:experiments-pretrained_bert} and \ref{sec:detailed_analysis}, we showed the experimental results of mixing ratio for BERT-base with Wikipedia dataset.
We also conducted the experiments with the pre-trained BERT-large~\cite{devlin2018bert}, BERT-medium, BERT-small, BERT-mini, BERT-tiny~\cite{turc2019wellread}, and RoBERTa-large~\cite{liu19}.
Table~\ref{table:bert_size} shows the architecture hyperparameters of each model.
Table~\ref{table:overall_mixing_ratio_models} shows the statistics of the mixing ratio for each model.
Figures~\ref{fig:mixing_rate_large} to \ref{fig:roberta_mixing_rate_large} show the mixing ratio at each layer (each attention block) of each model.

\begin{figure*}[t]
    \centering
    \begin{minipage}[t]{.19\hsize}
        \centering
        \includegraphics[height=6cm]{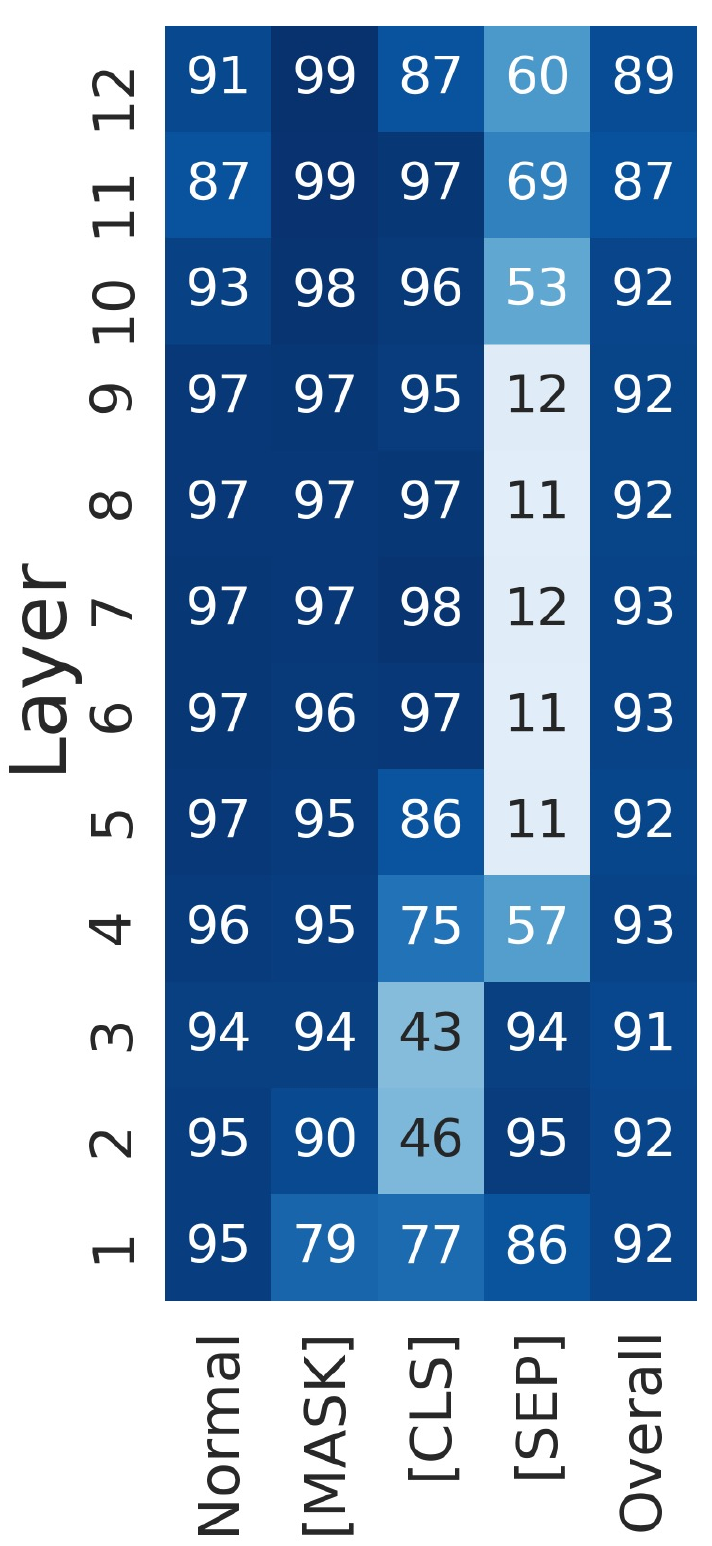}
        \subcaption{
        \textsc{Attn-w}.
        }
        \label{fig:bert_weight_rate_sst}
    \end{minipage}
    \;
    \begin{minipage}[t]{.17\hsize}
        \centering
        \includegraphics[height=6cm]{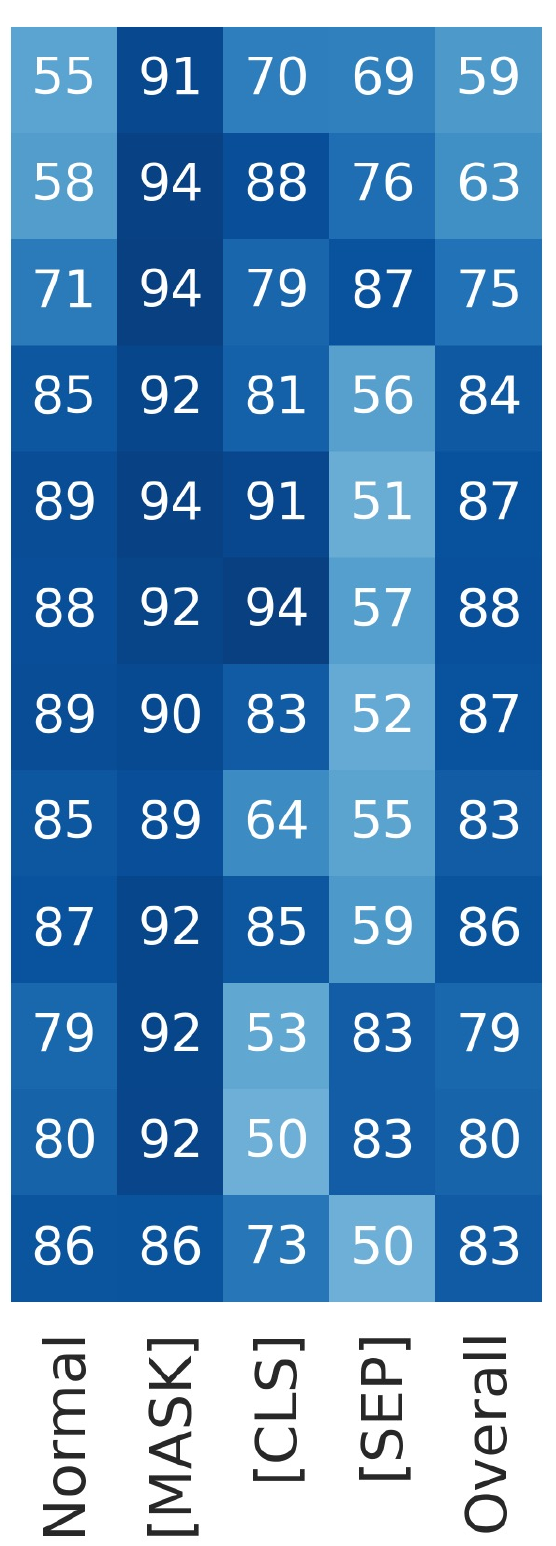}
        \subcaption{
        \textsc{Attn-n} \cite{kobayashi-etal-2020-attention}.
        }
        \label{fig:bert_kobayashi_rate_sst}
    \end{minipage}
    \;
    \begin{minipage}[t]{.17\hsize}
        \centering
        \includegraphics[height=6cm]{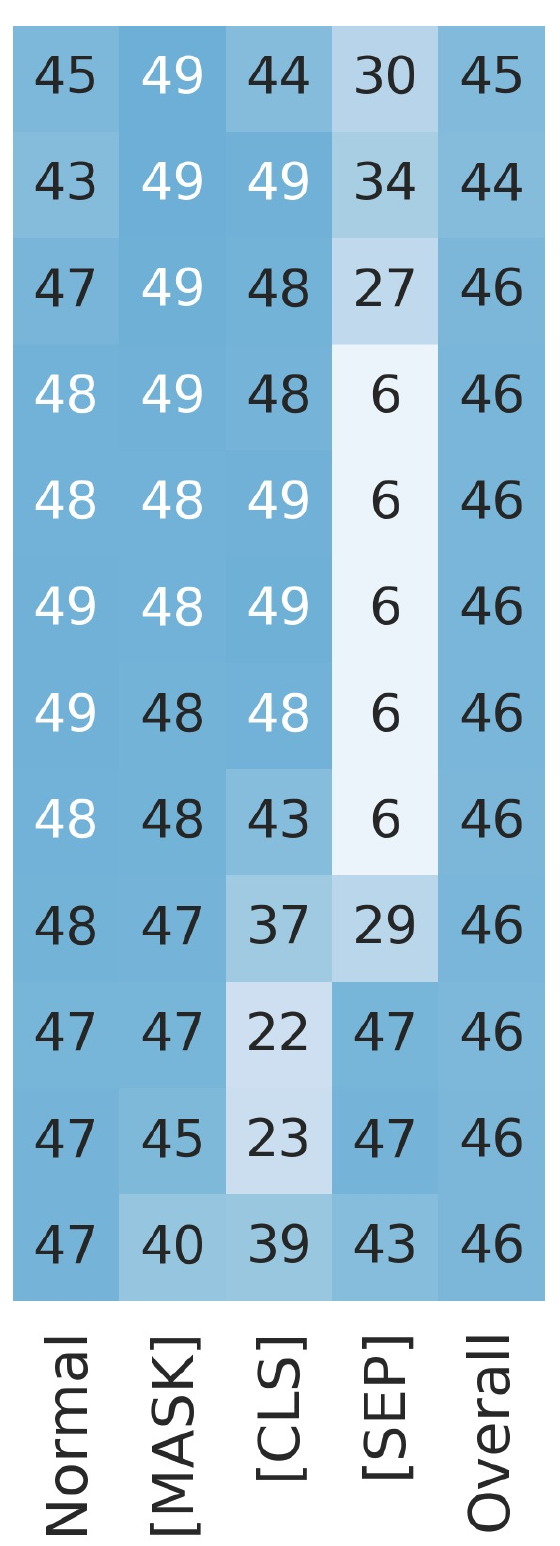}
        \subcaption{
        \textsc{AttnRes-w} \cite{abnar-zuidema-2020-quantifying}.
        }
        \label{fig:bert_abnar_rate_sst}
    \end{minipage}
    \;
    \begin{minipage}[t]{.17\hsize}
    \centering
    \includegraphics[height=6cm]{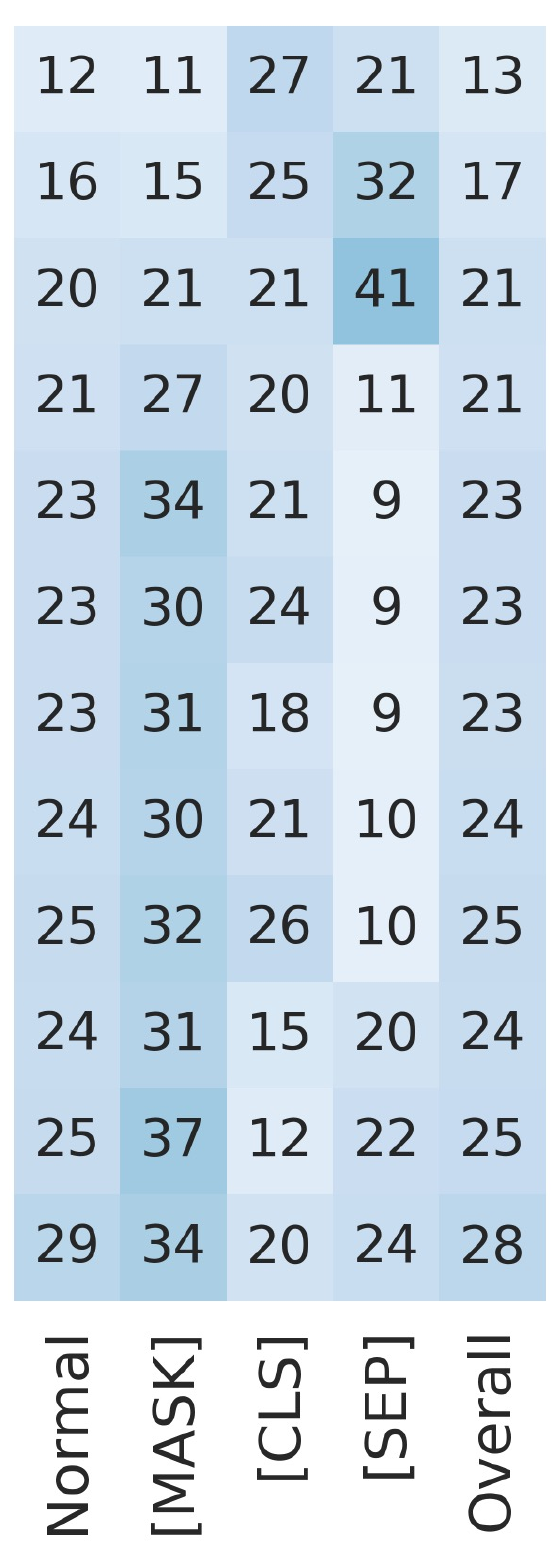}
    \subcaption{
    \textsc{AttnRes-n}.
    }
    \label{fig:before_ln_mixing_rate_sst}
    \end{minipage}
    \;
    \begin{minipage}[t]{.22\hsize}
    \centering
    \includegraphics[height=6cm]{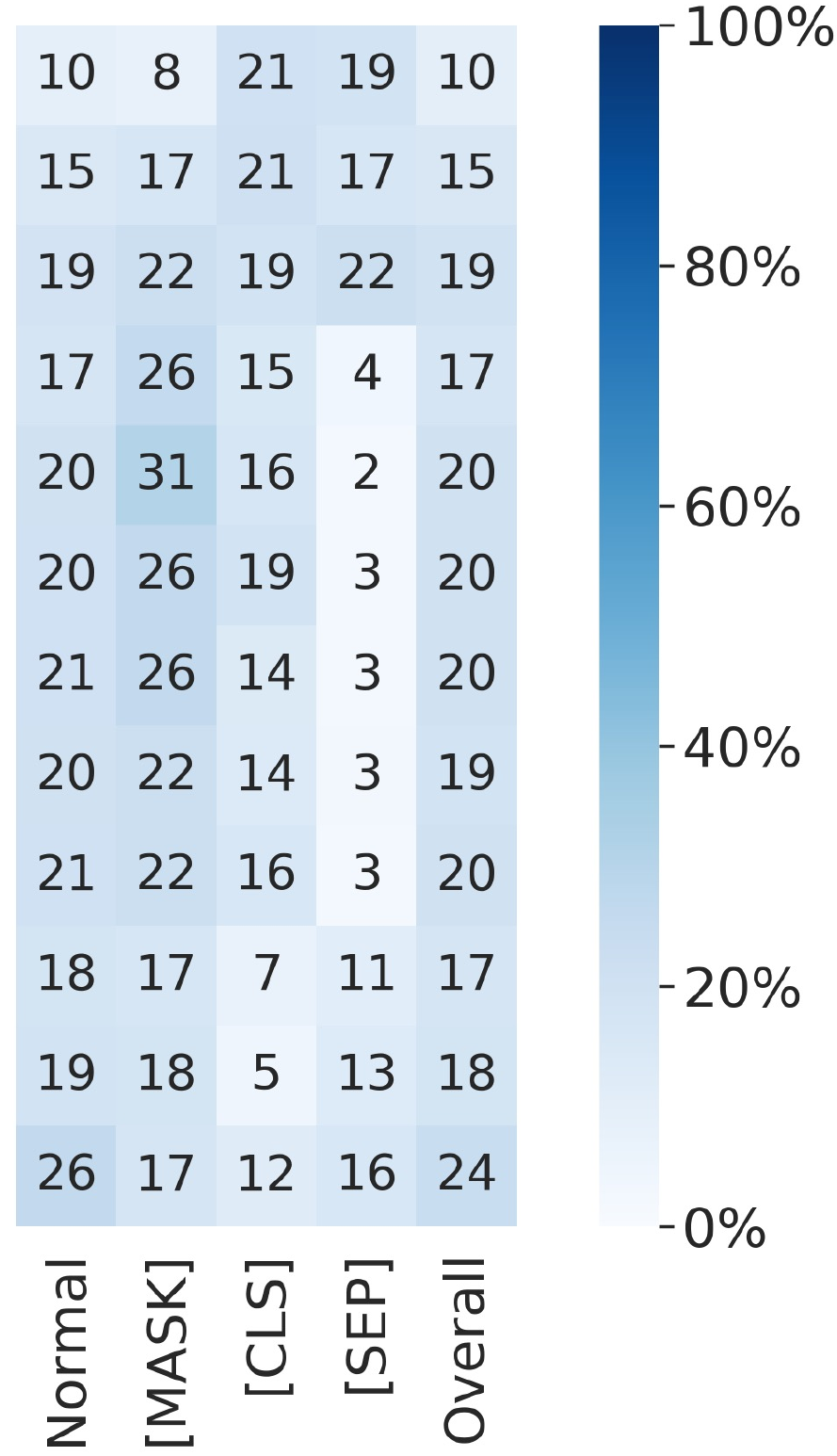}
    \subcaption{
    \textsc{AttnResLn-n}.
    }
    \label{fig:bert_mix_rate_sst}
    \end{minipage}
    \caption{
    Mixing ratio at each layer of BERT-base calculated from each method on the SST-2.
    }
    \label{fig:mixing_rate_sst}
\end{figure*}

\begin{figure*}[t]
    \centering
    \begin{minipage}[t]{.19\hsize}
        \centering
        \includegraphics[height=6cm]{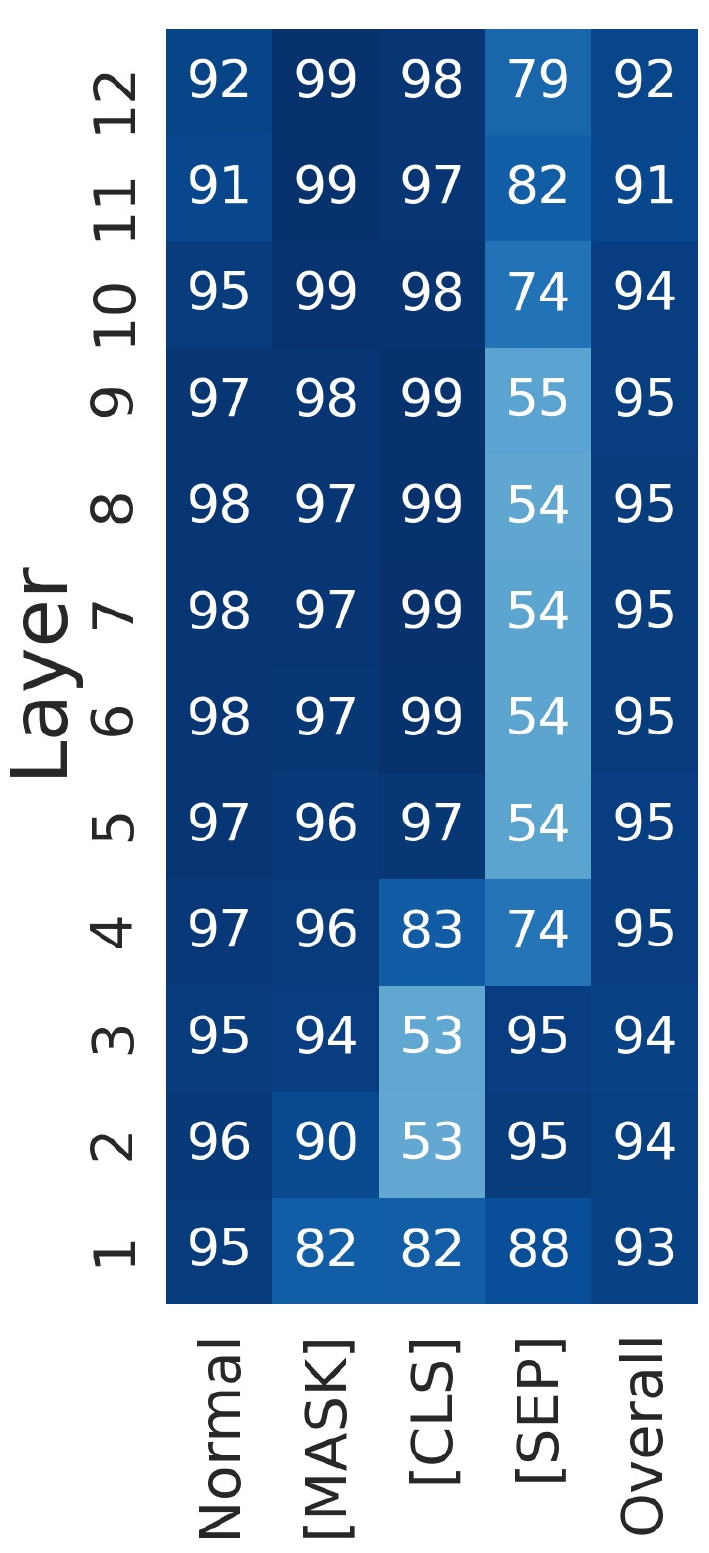}
        \subcaption{
        \textsc{Attn-w}.
        }
        \label{fig:bert_weight_rate_mnli}
    \end{minipage}
    \;
    \begin{minipage}[t]{.17\hsize}
        \centering
        \includegraphics[height=6cm]{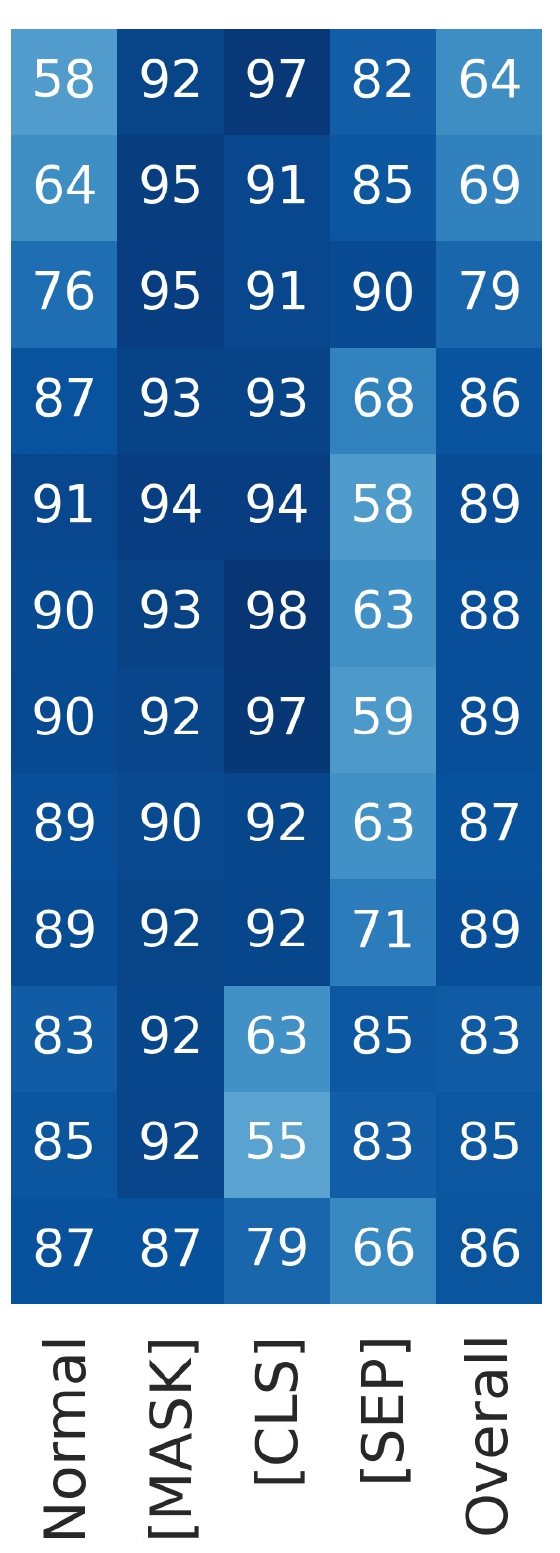}
        \subcaption{
        \textsc{Attn-n} \cite{kobayashi-etal-2020-attention}.
        }
        \label{fig:bert_kobayashi_rate_mnli}
    \end{minipage}
    \;
    \begin{minipage}[t]{.17\hsize}
        \centering
        \includegraphics[height=6cm]{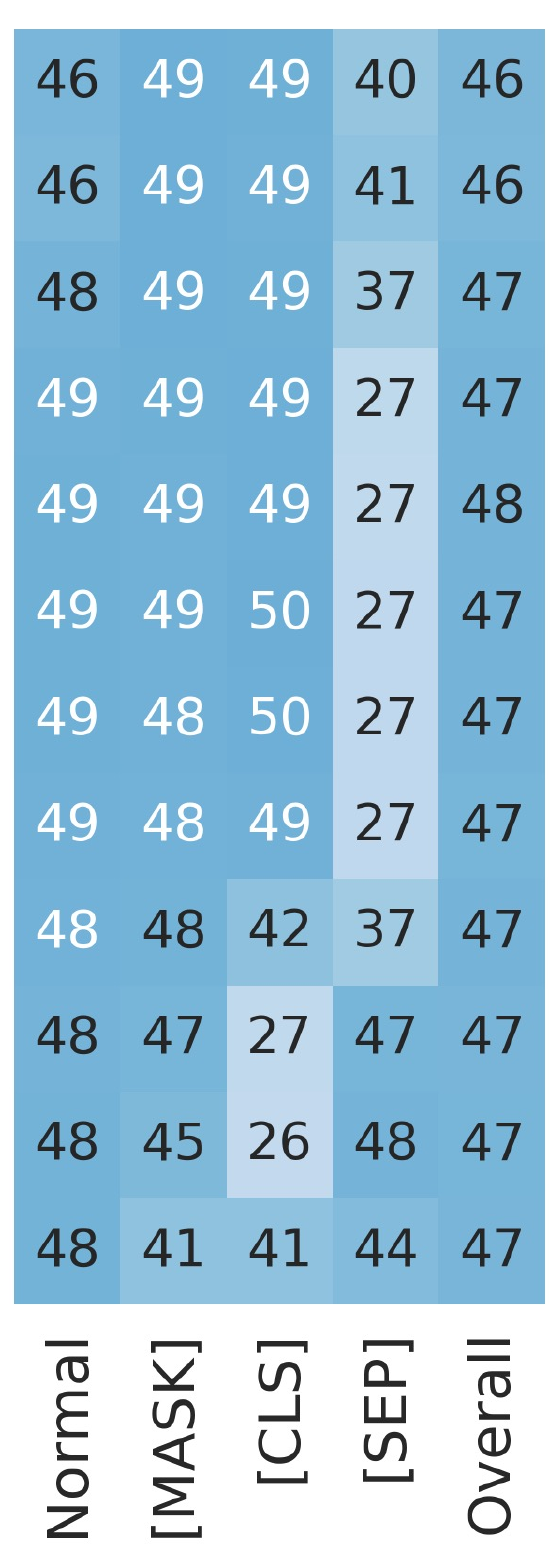}
        \subcaption{
        \textsc{AttnRes-w} \cite{abnar-zuidema-2020-quantifying}.
        }
        \label{fig:bert_abnar_rate_mnli}
    \end{minipage}
    \;
    \begin{minipage}[t]{.17\hsize}
    \centering
    \includegraphics[height=6cm]{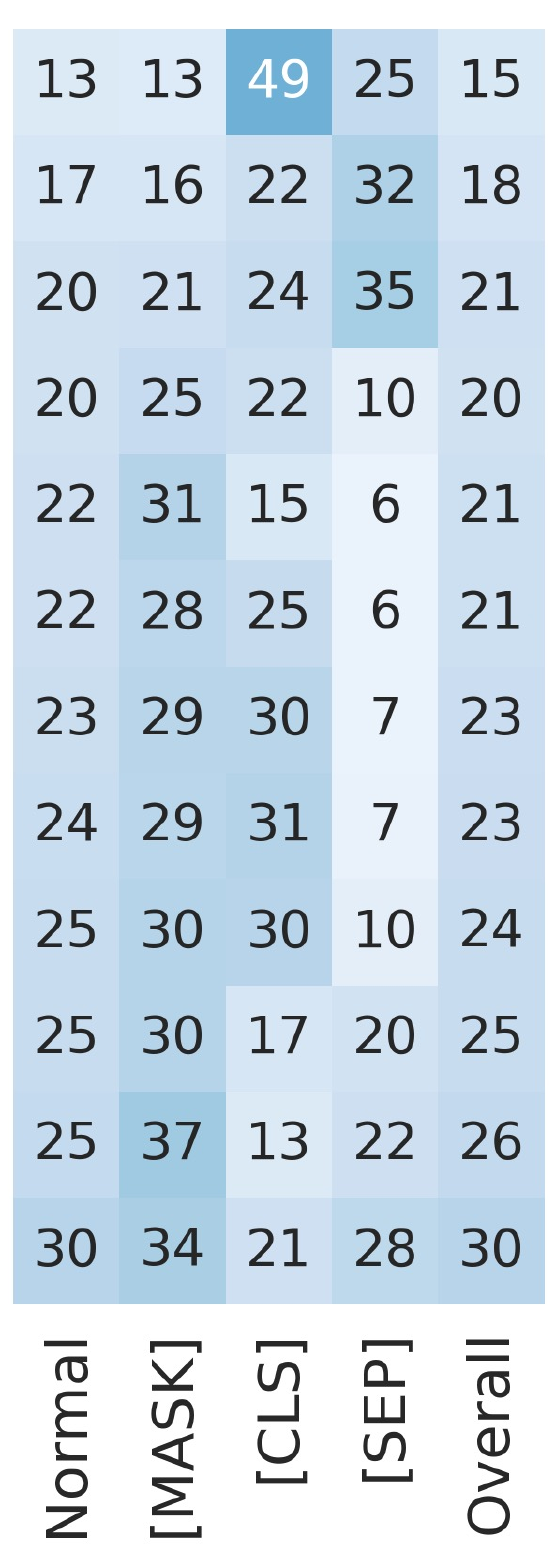}
    \subcaption{
    \textsc{AttnRes-n}.
    }
    \label{fig:before_ln_mixing_rate_mnli}
    \end{minipage}
    \;
    \begin{minipage}[t]{.22\hsize}
    \centering
    \includegraphics[height=6cm]{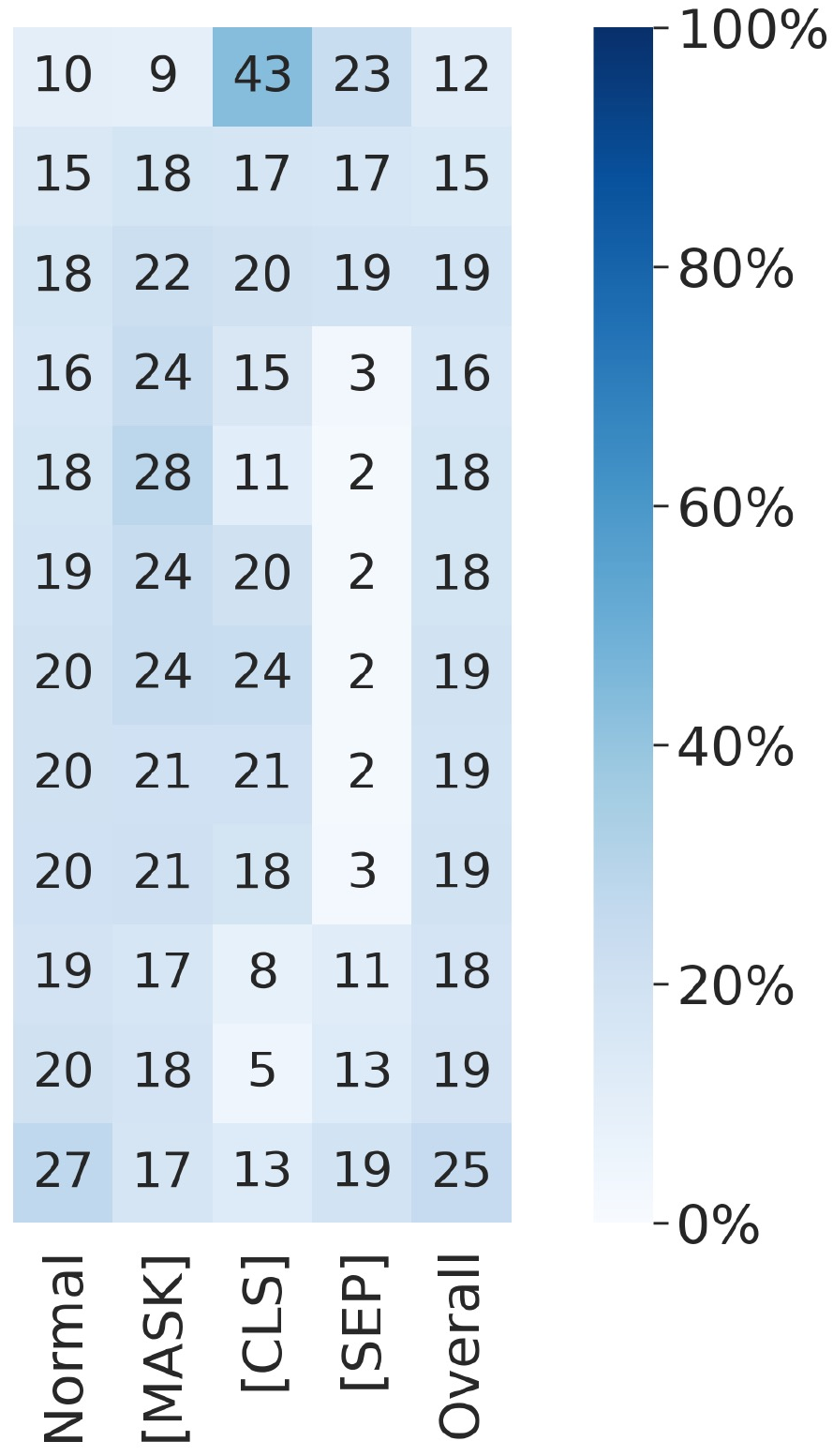}
    \subcaption{
    \textsc{AttnResLn-n}.
    }
    \label{fig:bert_mix_rate_mnli}
    \end{minipage}
    \caption{
    Mixing ratio at each layer of BERT-base calculated from each method on the MNLI.
    }
    \label{fig:mixing_rate_mnli}
\end{figure*}

\begin{figure*}[t]
    \centering
    \begin{minipage}[t]{.19\hsize}
        \centering
        \includegraphics[height=6cm]{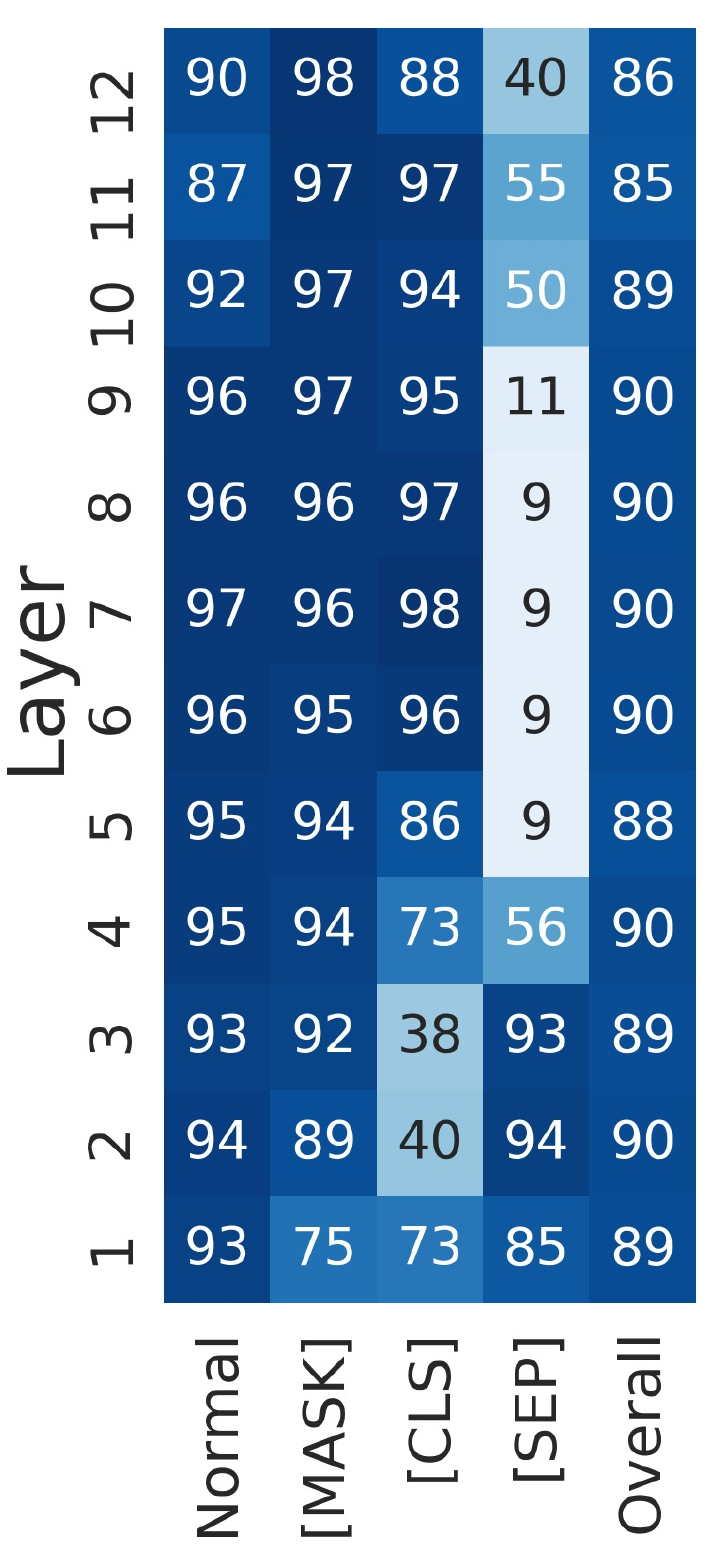}
        \subcaption{
        \textsc{Attn-w}.
        }
        \label{fig:bert_weight_rate_ner}
    \end{minipage}
    \;
    \begin{minipage}[t]{.17\hsize}
        \centering
        \includegraphics[height=6cm]{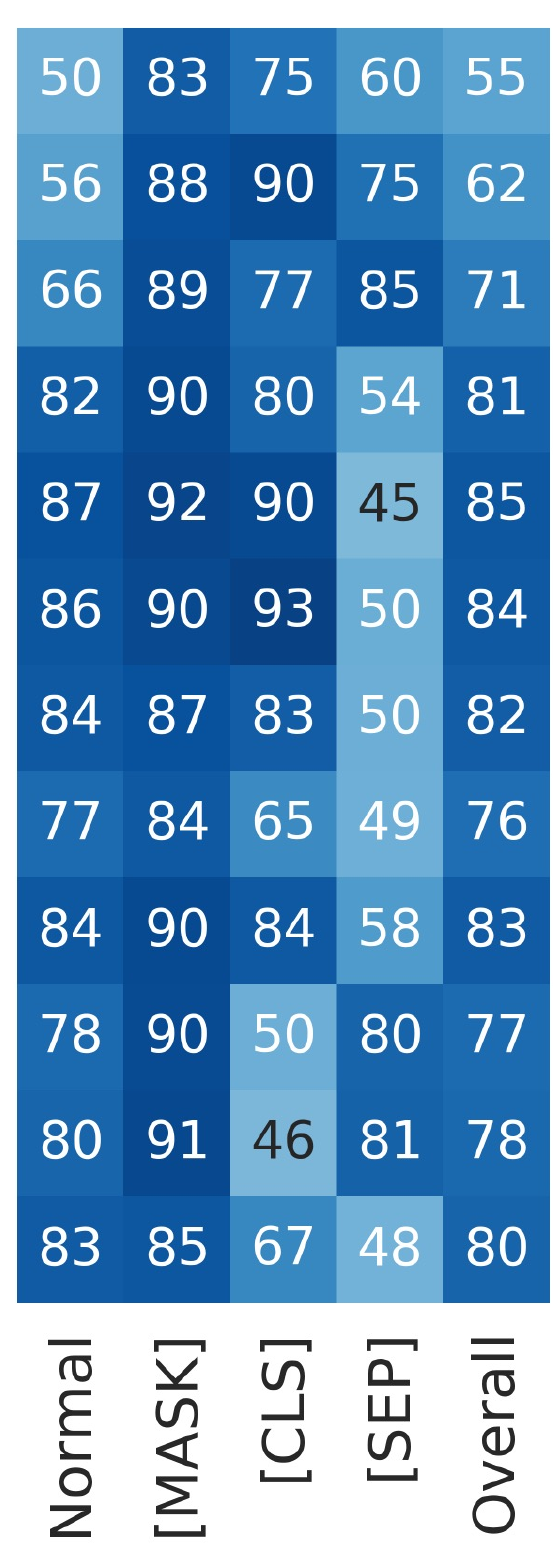}
        \subcaption{
        \textsc{Attn-n} \cite{kobayashi-etal-2020-attention}.
        }
        \label{fig:bert_kobayashi_rate_ner}
    \end{minipage}
    \;
    \begin{minipage}[t]{.17\hsize}
        \centering
        \includegraphics[height=6cm]{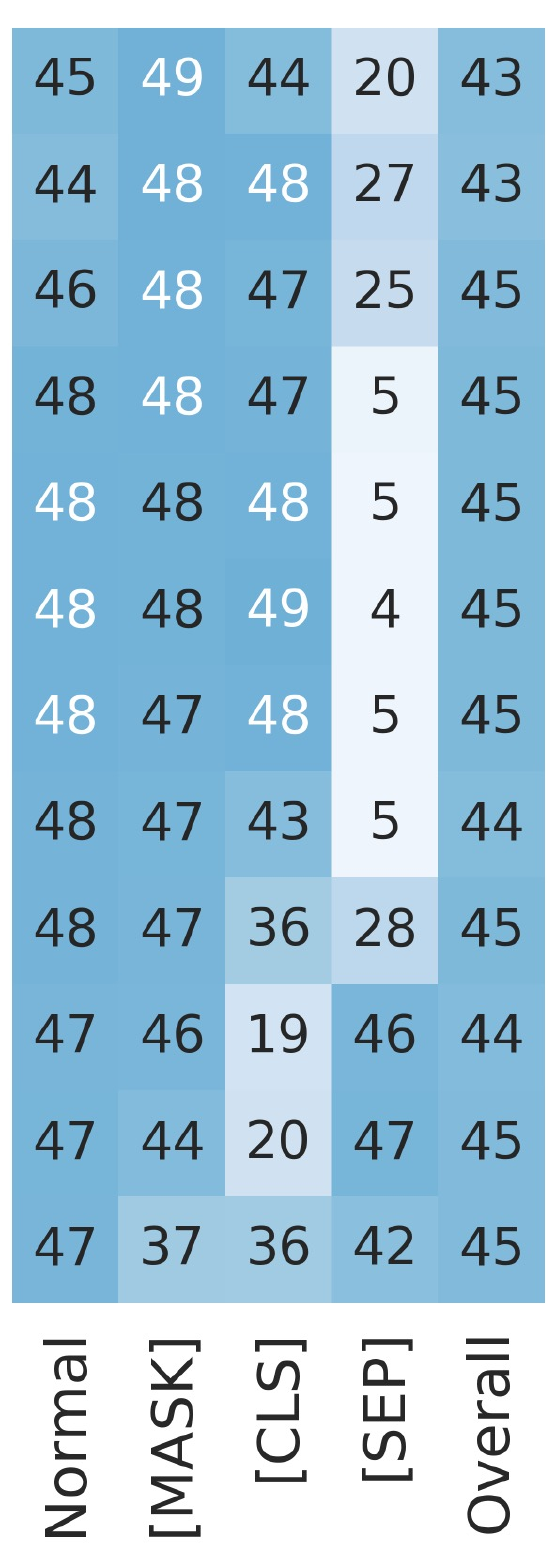}
        \subcaption{
        \textsc{AttnRes-w} \cite{abnar-zuidema-2020-quantifying}.
        }
        \label{fig:bert_abnar_rate_ner}
    \end{minipage}
    \;
    \begin{minipage}[t]{.17\hsize}
    \centering
    \includegraphics[height=6cm]{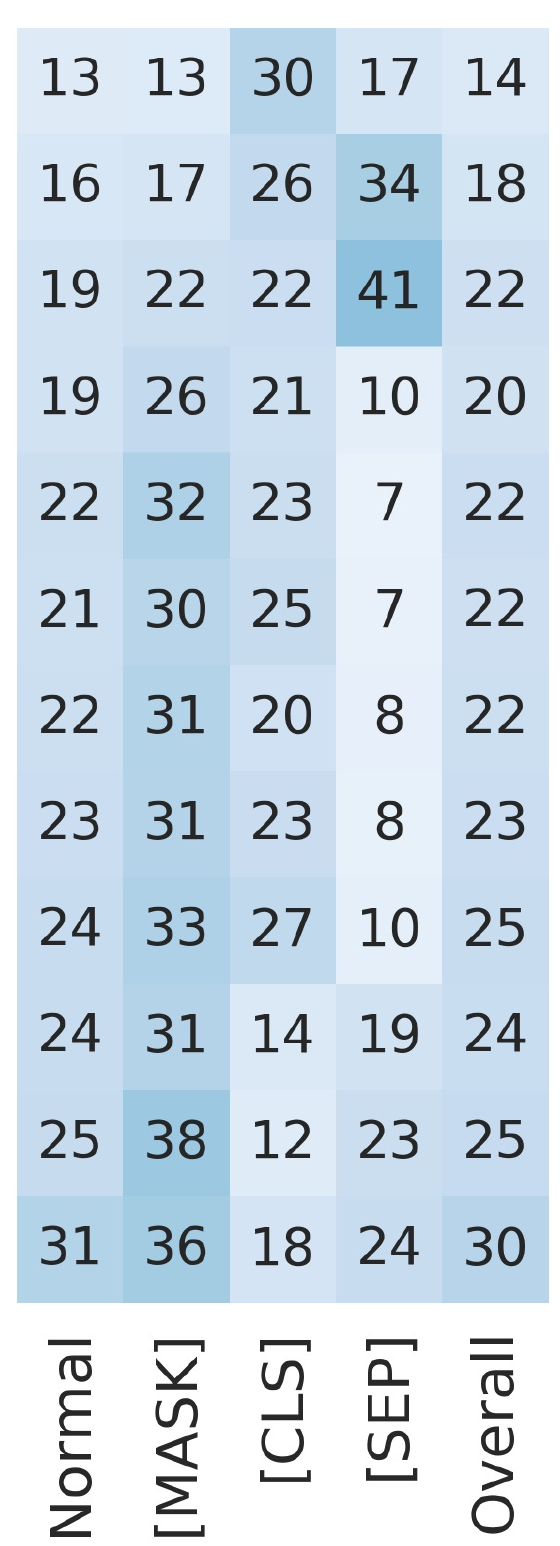}
    \subcaption{
    \textsc{AttnRes-n}.
    }
    \label{fig:before_ln_mixing_rate_ner}
    \end{minipage}
    \;
    \begin{minipage}[t]{.22\hsize}
    \centering
    \includegraphics[height=6cm]{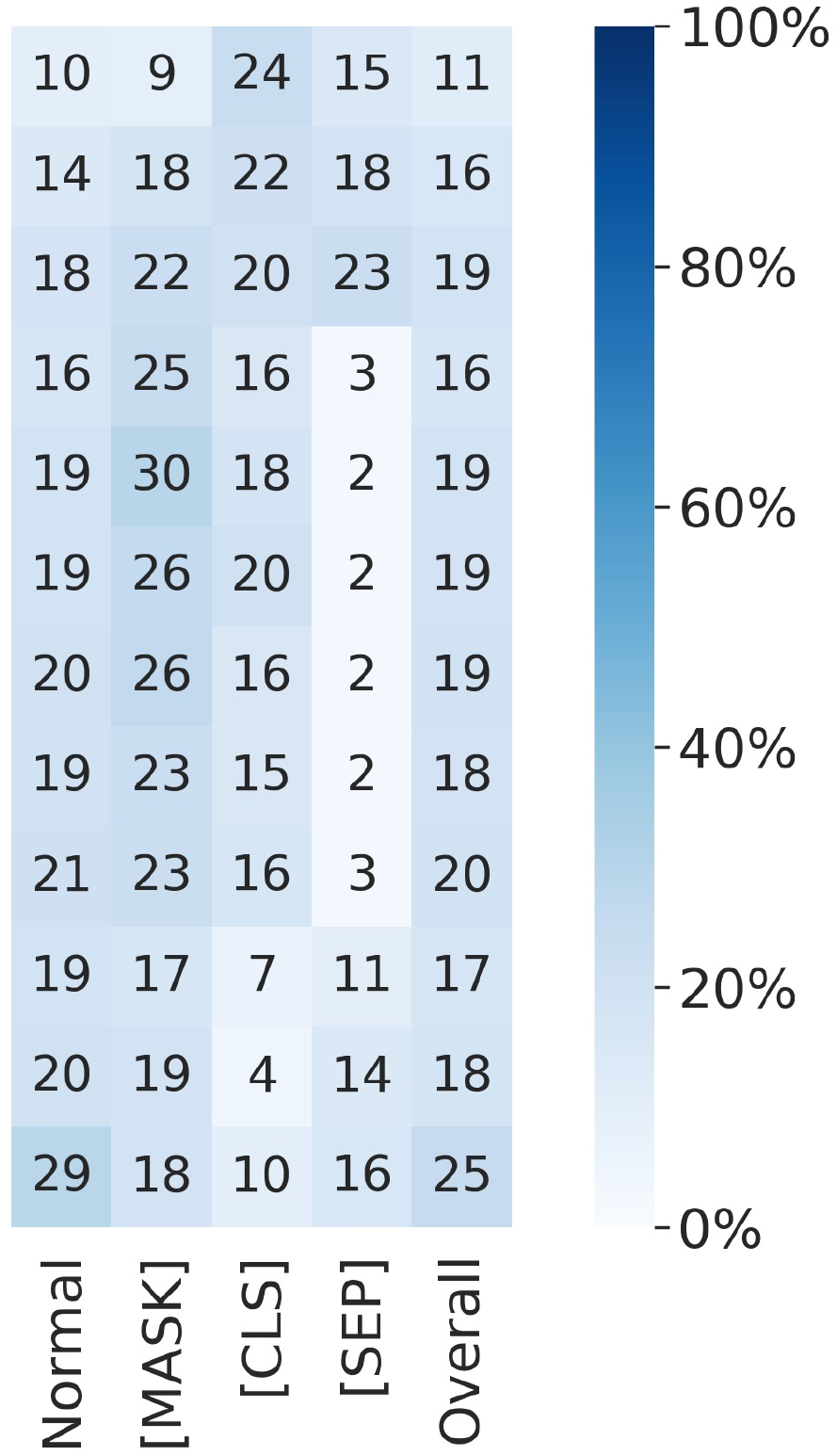}
    \subcaption{
    \textsc{AttnResLn-n}.
    }
    \label{fig:bert_mix_rate_ner}
    \end{minipage}
    \caption{
    Mixing ratio at each layer of BERT-base calculated from each method on the CoNLL'03 NER.
    }
    \label{fig:mixing_rate_ner}
\end{figure*}

We also conducted it with the other three datasets.
Table~\ref{table:overall_mixing_ratio_data} shows the statistics of the mixing ratio for BERT-base on each dataset.
Figures~\ref{fig:mixing_rate_sst} to \ref{fig:mixing_rate_ner} show the mixing ratio at each layer of BERT-base on each dataset.

\begin{figure*}[t]
    \centering
    \begin{minipage}[t]{.19\hsize}
        \centering
        \includegraphics[height=6cm]{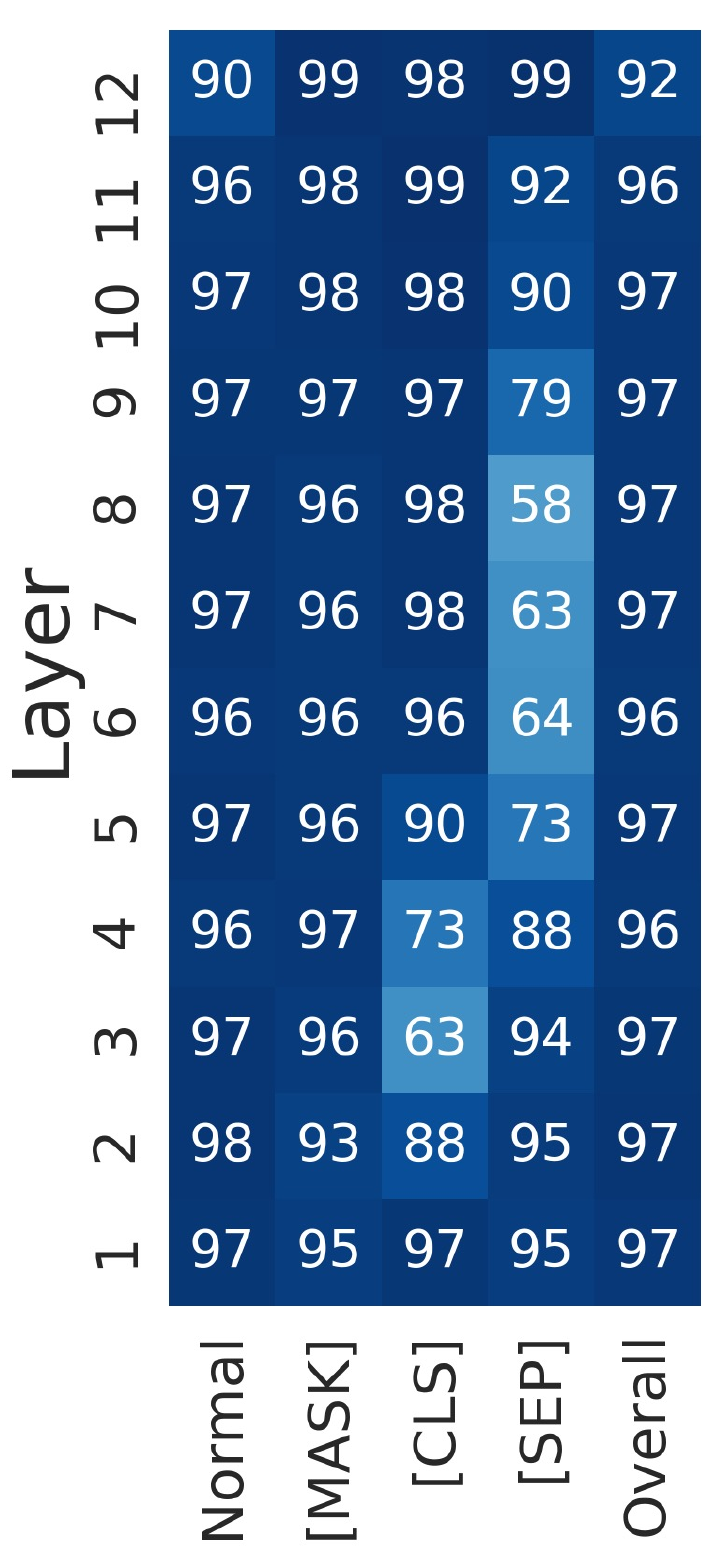}
        \subcaption{
        \textsc{Attn-w}.
        }
        \label{fig:bert_weight_rate_seed0}
    \end{minipage}
    \;
    \begin{minipage}[t]{.17\hsize}
        \centering
        \includegraphics[height=6cm]{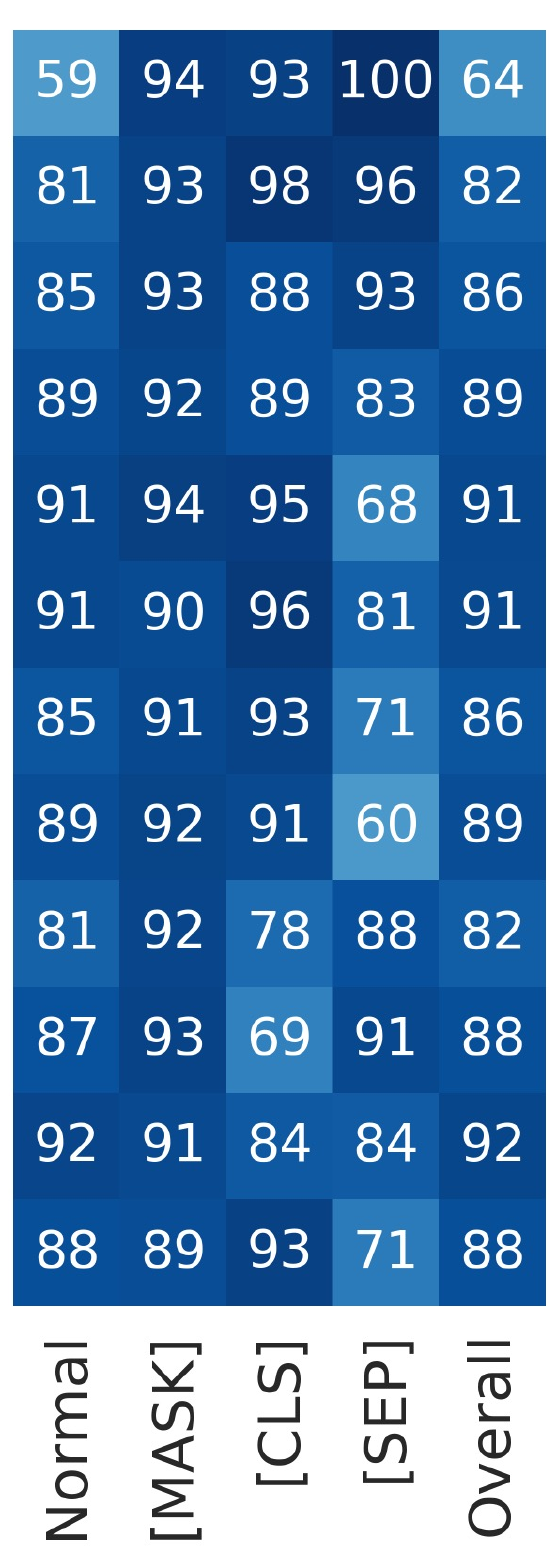}
        \subcaption{
        \textsc{Attn-n} \cite{kobayashi-etal-2020-attention}.
        }
        \label{fig:bert_kobayashi_rate_seed0}
    \end{minipage}
    \;
    \begin{minipage}[t]{.17\hsize}
        \centering
        \includegraphics[height=6cm]{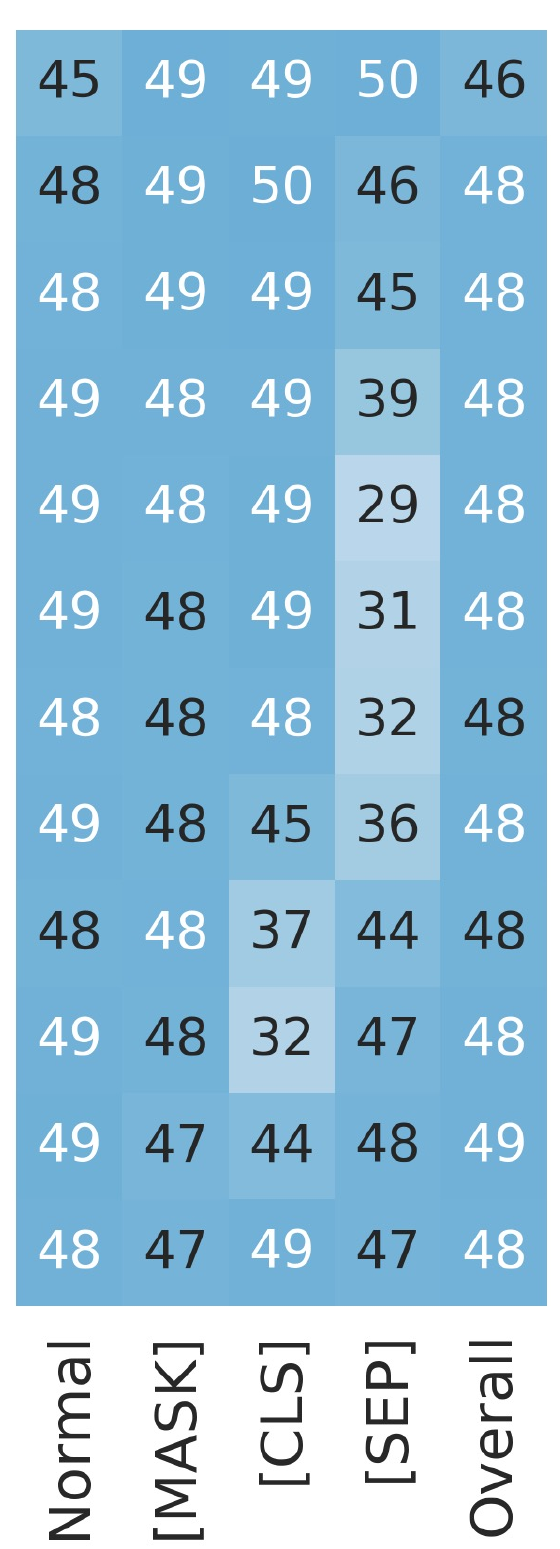}
        \subcaption{
        \textsc{AttnRes-w} \cite{abnar-zuidema-2020-quantifying}.
        }
        \label{fig:bert_abnar_rate_seed0}
    \end{minipage}
    \;
    \begin{minipage}[t]{.17\hsize}
    \centering
    \includegraphics[height=6cm]{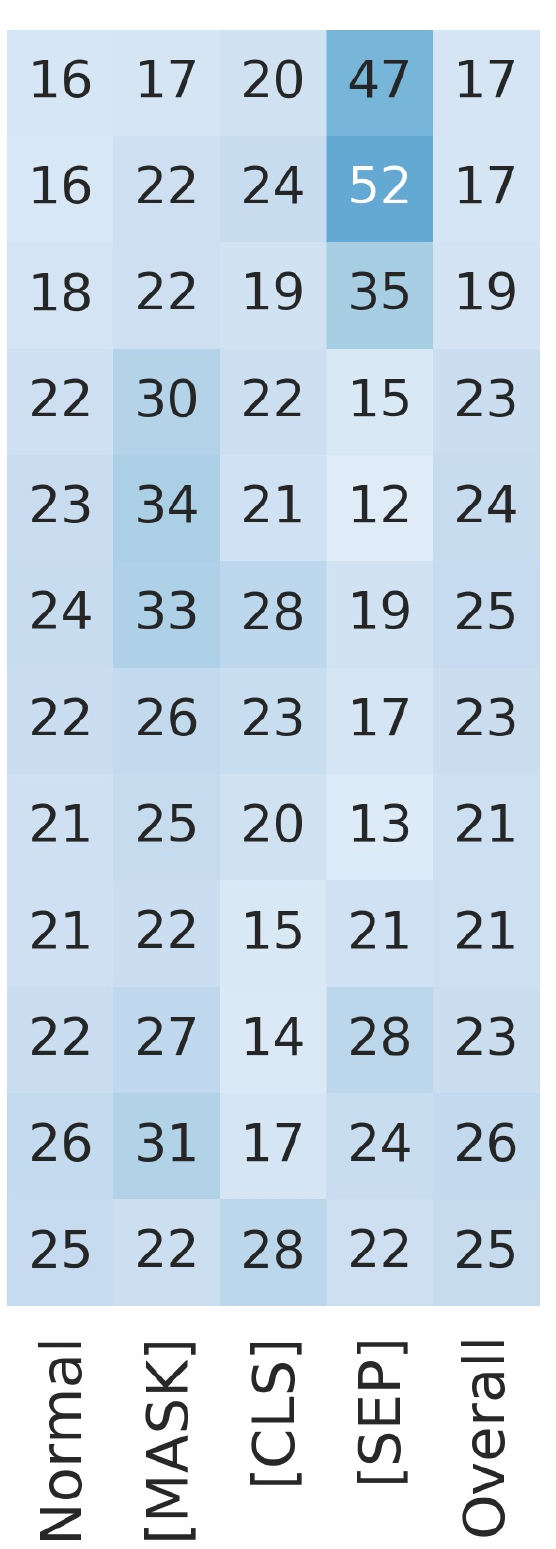}
    \subcaption{
    \textsc{AttnRes-n}.
    }
    \label{fig:before_ln_mixing_rate_seed0}
    \end{minipage}
    \;
    \begin{minipage}[t]{.22\hsize}
    \centering
    \includegraphics[height=6cm]{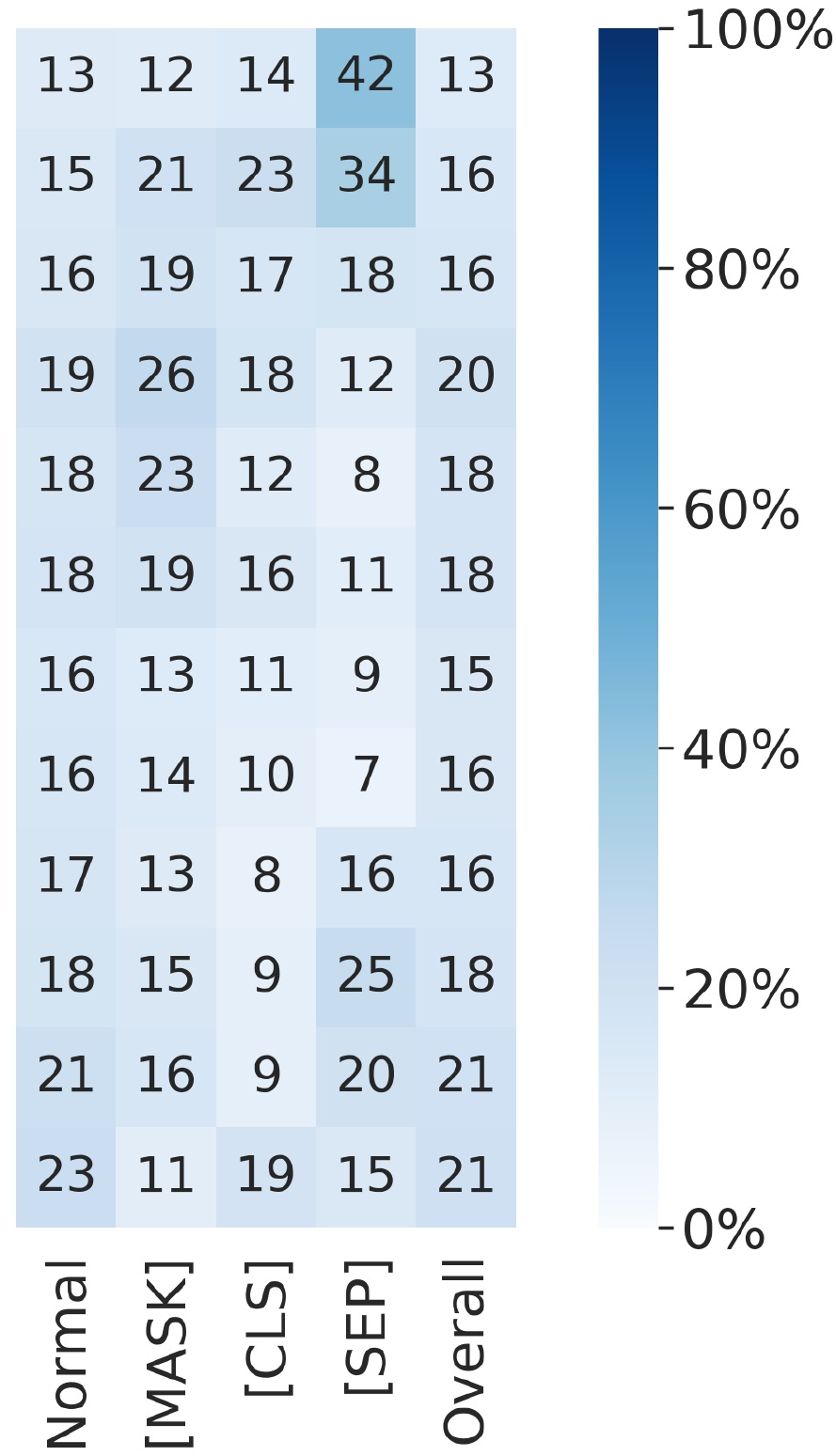}
    \subcaption{
    \textsc{AttnResLn-n}.
    }
    \label{fig:bert_mix_rate_seed0}
    \end{minipage}
    \caption{
    Mixing ratio at each layer of BERT-base trained with $0$th seed. 
    }
    \label{fig:mixing_rate_seed0}
\end{figure*}

\begin{figure*}[t]
    \centering
    \begin{minipage}[t]{.19\hsize}
        \centering
        \includegraphics[height=6cm]{figures/seed0/seed0_attn_weight.pdf}
        \subcaption{
        \textsc{Attn-w}.
        }
        \label{fig:bert_weight_rate_seed10}
    \end{minipage}
    \;
    \begin{minipage}[t]{.17\hsize}
        \centering
        \includegraphics[height=6cm]{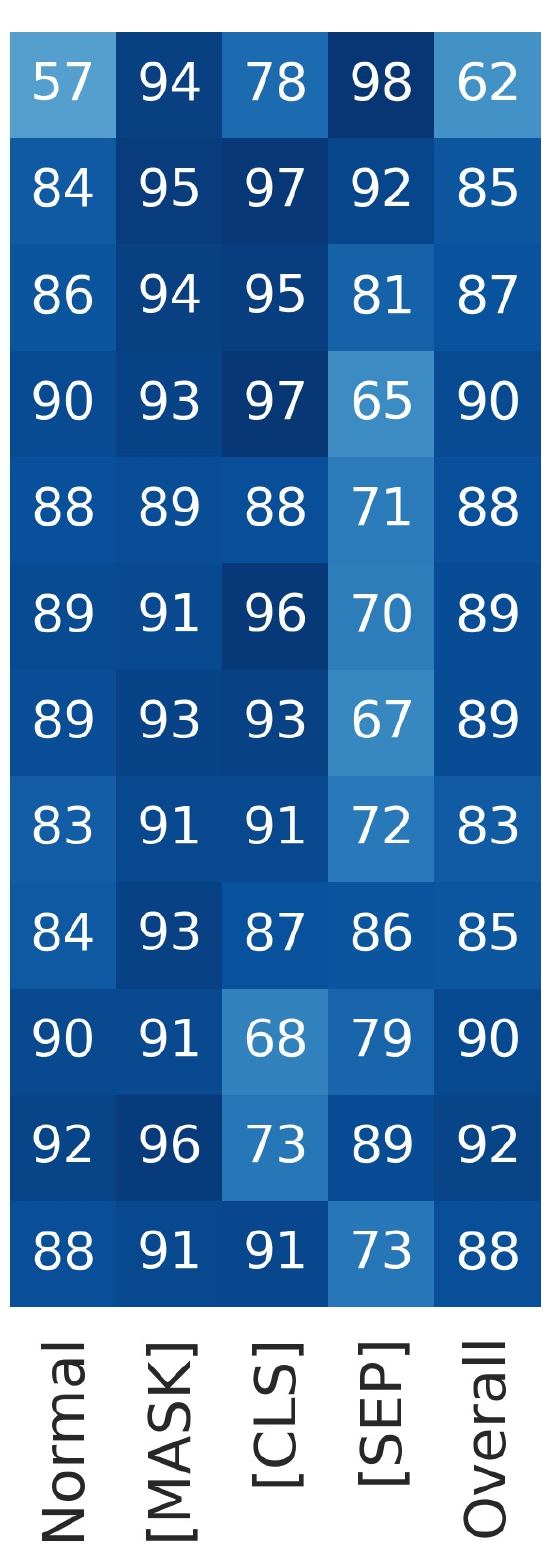}
        \subcaption{
        \textsc{Attn-n} \cite{kobayashi-etal-2020-attention}.
        }
        \label{fig:bert_kobayashi_rate_seed10}
    \end{minipage}
    \;
    \begin{minipage}[t]{.17\hsize}
        \centering
        \includegraphics[height=6cm]{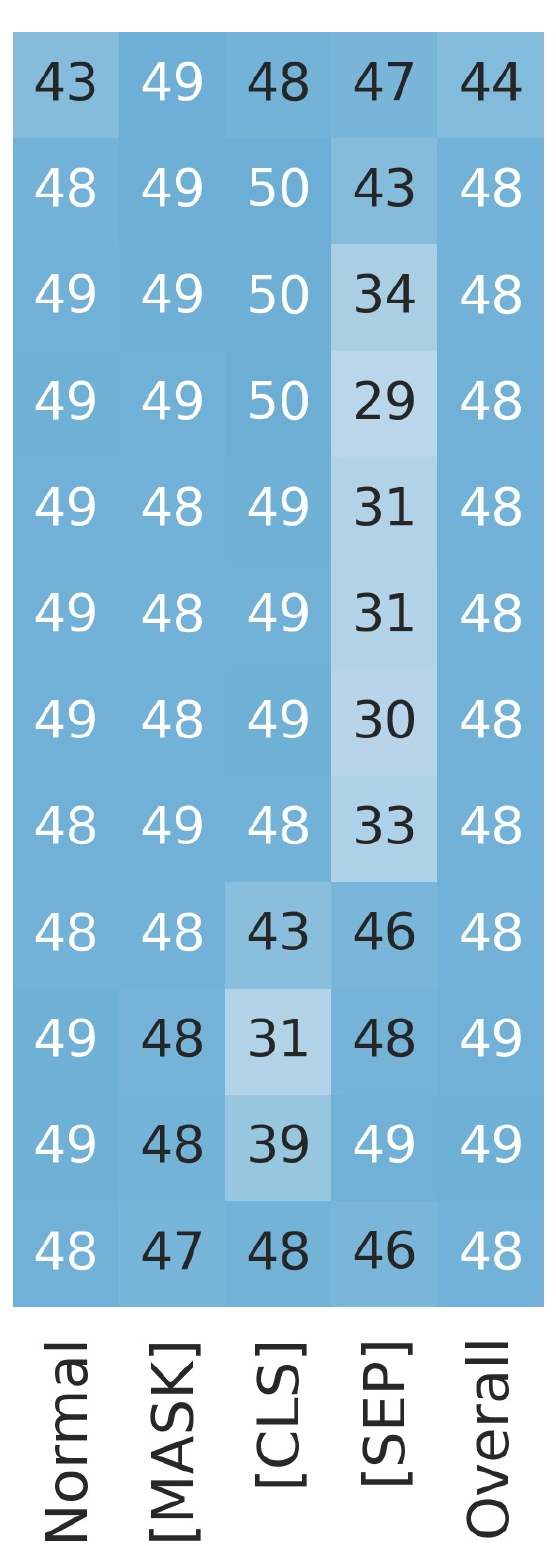}
        \subcaption{
        \textsc{AttnRes-w} \cite{abnar-zuidema-2020-quantifying}.
        }
        \label{fig:bert_abnar_rate_seed10}
    \end{minipage}
    \;
    \begin{minipage}[t]{.17\hsize}
    \centering
    \includegraphics[height=6cm]{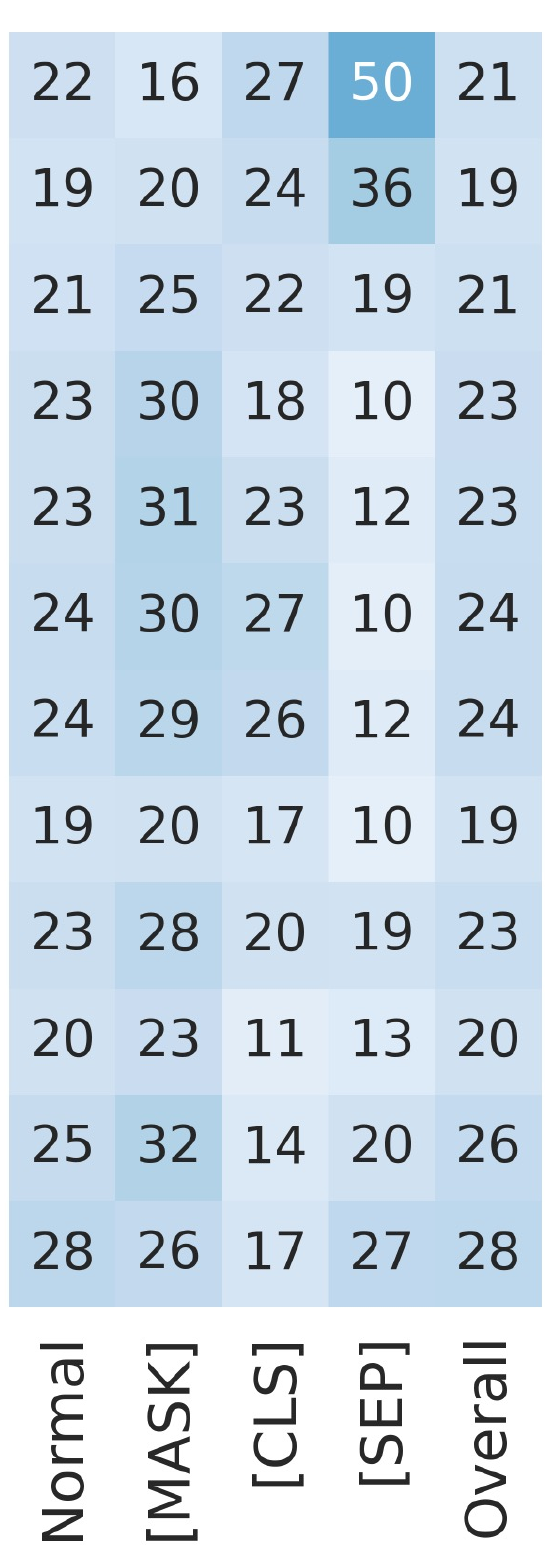}
    \subcaption{
    \textsc{AttnRes-n}.
    }
    \label{fig:before_ln_mixing_rate_seed10}
    \end{minipage}
    \;
    \begin{minipage}[t]{.22\hsize}
    \centering
    \includegraphics[height=6cm]{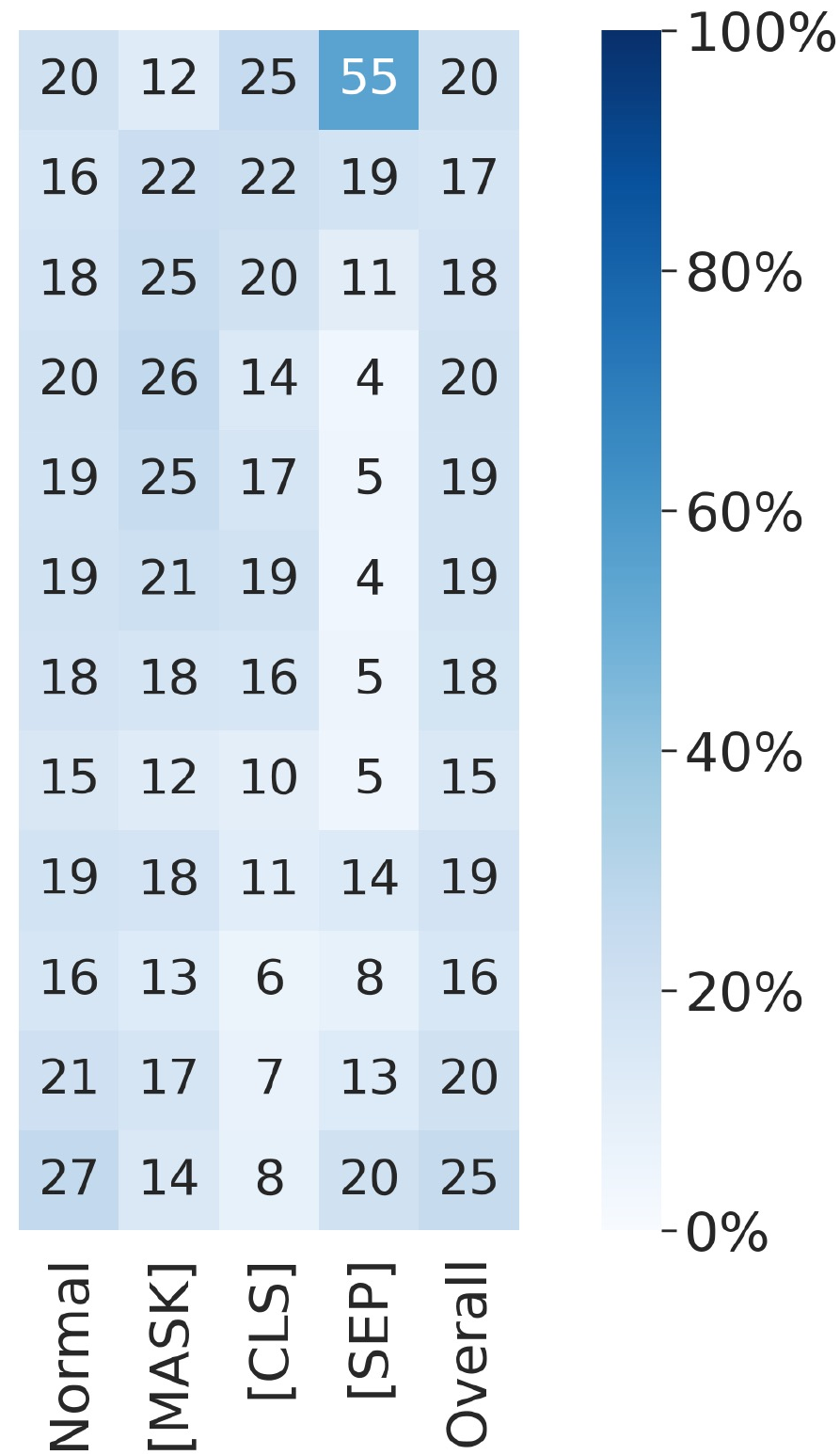}
    \subcaption{
    \textsc{AttnResLn-n}.
    }
    \label{fig:bert_mix_rate_seed10}
    \end{minipage}
    \caption{
    Mixing ratio at each layer of BERT-base trained with $10$th seed. 
    }
    \label{fig:mixing_rate_seed10}
\end{figure*}

\begin{figure*}[t]
    \centering
    \begin{minipage}[t]{.19\hsize}
        \centering
        \includegraphics[height=6cm]{figures/seed0/seed0_attn_weight.pdf}
        \subcaption{
        \textsc{Attn-w}.
        }
        \label{fig:bert_weight_rate_seed20}
    \end{minipage}
    \;
    \begin{minipage}[t]{.17\hsize}
        \centering
        \includegraphics[height=6cm]{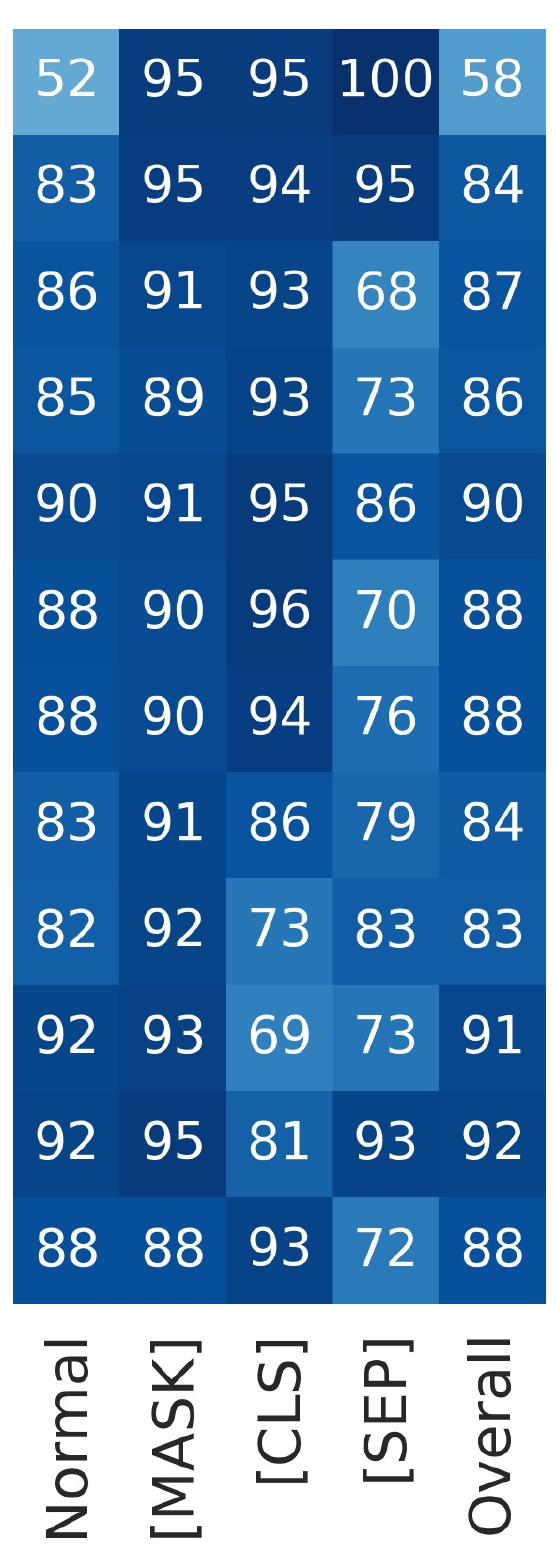}
        \subcaption{
        \textsc{Attn-n} \cite{kobayashi-etal-2020-attention}.
        }
        \label{fig:bert_kobayashi_rate_seed20}
    \end{minipage}
    \;
    \begin{minipage}[t]{.17\hsize}
        \centering
        \includegraphics[height=6cm]{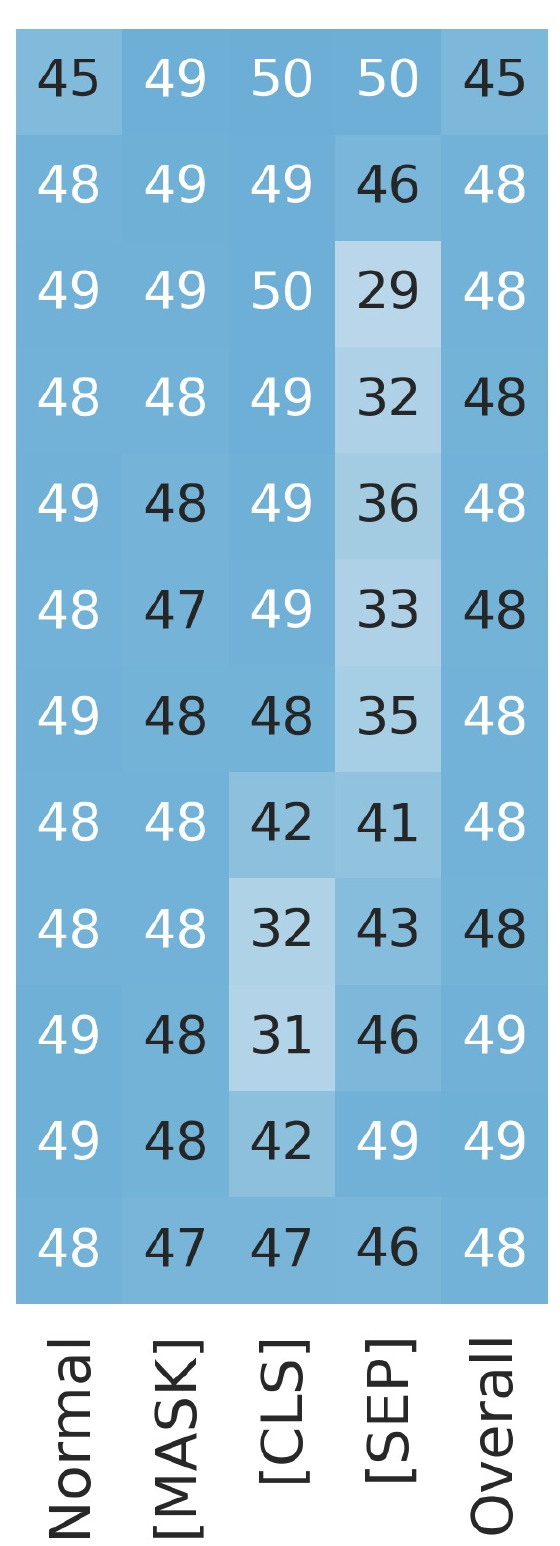}
        \subcaption{
        \textsc{AttnRes-w} \cite{abnar-zuidema-2020-quantifying}.
        }
        \label{fig:bert_abnar_rate_seed20}
    \end{minipage}
    \;
    \begin{minipage}[t]{.17\hsize}
    \centering
    \includegraphics[height=6cm]{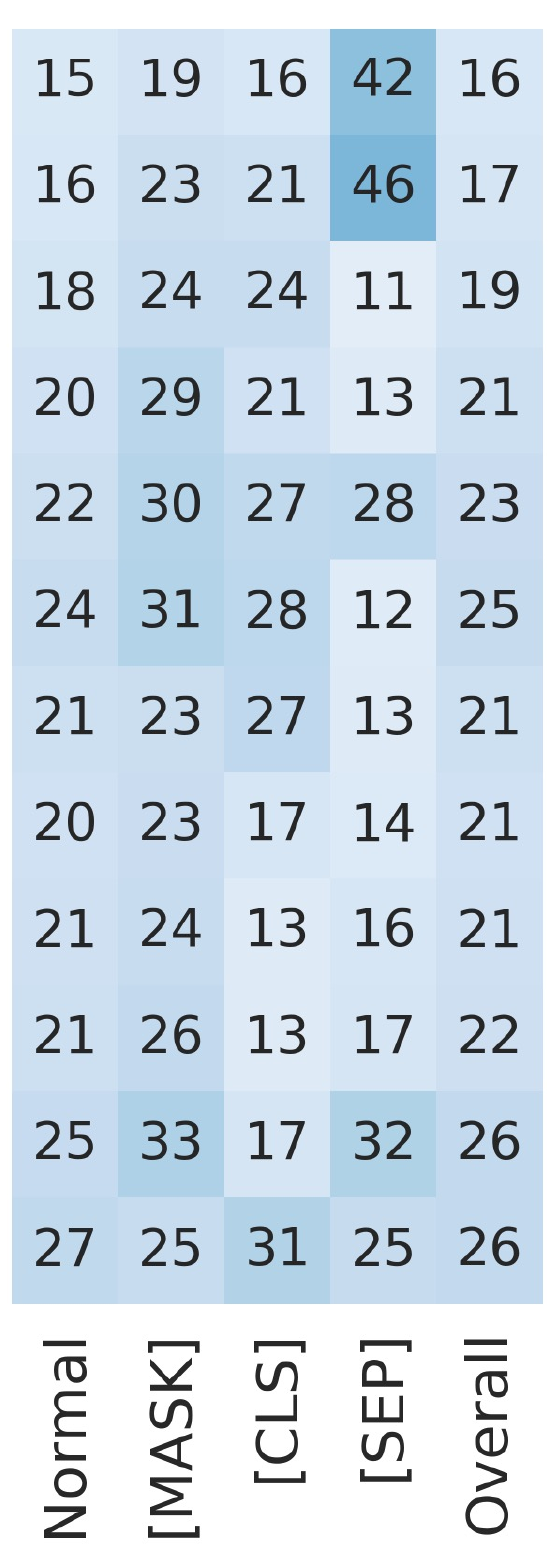}
    \subcaption{
    \textsc{AttnRes-n}.
    }
    \label{fig:before_ln_mixing_rate_seed20}
    \end{minipage}
    \;
    \begin{minipage}[t]{.22\hsize}
    \centering
    \includegraphics[height=6cm]{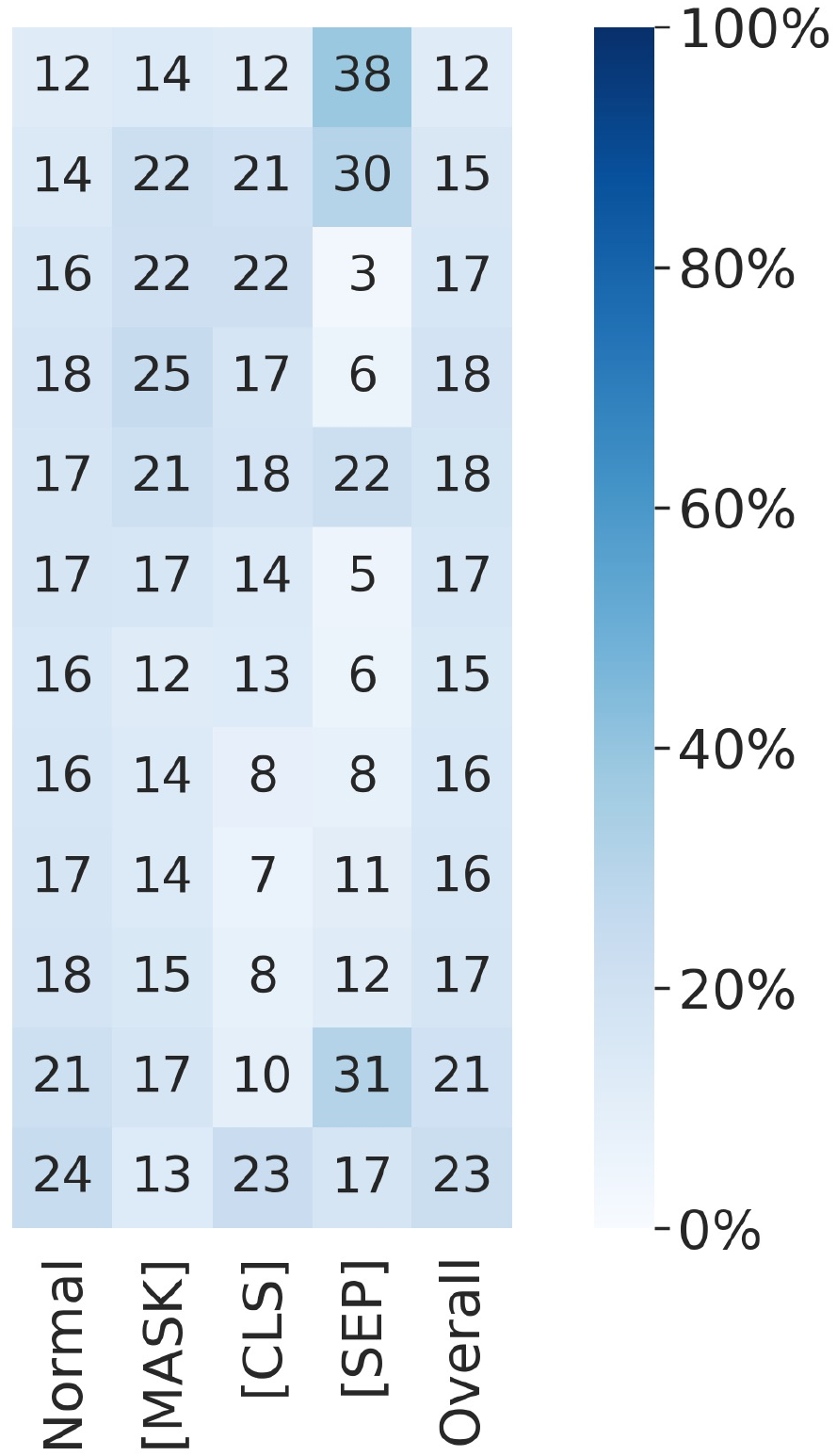}
    \subcaption{
    \textsc{AttnResLn-n}.
    }
    \label{fig:bert_mix_rate_seed20}
    \end{minipage}
    \caption{
    Mixing ratio at each layer of BERT-base trained with $20$th seed. 
    }
    \label{fig:mixing_rate_seed20}
\end{figure*}

Furthermore, we conducted it with 25 BERT-base models trained with different seeds by \citet{sellam2021multiberts}.
Table~\ref{table:overall_mixing_ratio_seeds} shows the statistics of the mixing ratio for the models on the Wikipedia dataset.
Figures~\ref{fig:mixing_rate_seed0} to \ref{fig:mixing_rate_seed20} show the mixing ratio at each layer of three models (trained with $0$th, $5$th, $20$th seeds) from them.

In Section~\ref{subsec:analysis:pipeline}, we showed the distinctive trend for the \texttt{[MASK]} tokens in BERT-base with the Wikipedia dataset.
Even in the other models and with the other datasets, the mixing ratio for the masked tokens was relatively high in the middle and deep layers (Figures~\ref{fig:mixing_rate_large} to \ref{fig:bert_mix_rate_ner}).

Contrary to the results for the masked tokens, the trend for the beginning of sentence token (\texttt{[CLS]} or \texttt{<s>}) was different across these models (Figures~\ref{fig:mixing_rate_large} to \ref{fig:roberta_mixing_rate_large}).
For BERT-large, RoBERTa-large, and RoBERTa-base, the layer with the highest mixing ratio for CLS was the first layer, while for the other models, it was the final or penultimate layer.
Different trends between BERT and RoBERTa can be naturally explained by the fact that RoBERTa is pre-trained without the next sentence prediction.
Although we cannot interpret the difference of trends across BERT models with various sizes, it was consistent among them in that the later layers mix contextual information into \texttt{[CLS]} with a relatively high mixing ratio.
This implies that, in the later layers, BERT conducts some operations specialized to the next sentence prediction task.
Solving such a discourse-level task in the later layers is consistent with the previous report that BERT makes lower-level decisions (e.g., part-of-speech tagging) in the earlier layers and that the later layers have high-level information (e.g., knowledge on co-reference)~\cite{Tenney19pipeline}.

\section{Details on the investigation of the mechanism of ATTN's shrinking}
\label{ap:singular_values}

We describe the details of Section~\ref{subsec:mechanism}.

\subsection{Affine transformation in ATTN}
\label{ap:subsec:detail_transform}

\subsubsection*{Integration of each head's affine transformation}
To consider the scaling effect of the affine transformations in ATTN, we integrate each head's affine transformation $f^{\encircled{h}}$ into one affine transformation $f: \mathbb{R}^d \mapsto \mathbb{R}^d$, under a coarse assumption.
First, for simplicity, we assume that all heads in an ATTN assign the same weights
\begin{align}
    \alpha_{i,j}^{\encircled{1}} = \dots = \alpha_{i,j}^{\encircled{H}} \equiv \alpha_{i,j}
    \text{.}
\end{align}
Then, the computation of ATTN (Equation~\ref{eq:attn_decomposed}) can be rewritten as follows:
\begin{align}
    \mathrm{ATTN}(\bm x_i, \bm X)
    &= \sum_{j=1}^n \sum_{h=1}^H \alpha_{i,j}^{\encircled{h}} f^{\encircled{h}}(\bm x_j)
    \\
    &\approx \sum_{j=1}^n \alpha_{i,j} \sum_{h=1}^H f^{\encircled{h}}(\bm x_j)
    \\
    &= \sum_{j=1}^n \alpha_{i,j} f(\bm x_j)
    \label{eq:ap:attn_decomposed}
    \text{,}
\end{align}
where
\begin{align}
    f(\bm x)
    &:= \sum_{h=1}^H f^{\encircled{h}}(\bm x)
    \text{.}
\end{align}

\subsubsection*{Concrete computation of $f$}
From Equation~\ref{eq:attn_f},
the affine transformation $f$ is
\begin{align}
    f(\bm x)
    = \sum_{h=1}^H \left(\bm x \bm{W}_V^{\encircled{h}}+\bm{b}_V^{\encircled{h}} \right)\bm W_O^{\encircled{h}}
    \text{.}
\end{align}
Following the Transformer implementation, it can be further simplified as follows:
\begin{align}
    &f(\bm x) = \bigl(\bm x \bm{W}_V+\bm{b}_V \bigr)\bm W_O
    \text{,}
    \\
    &\bm W_V :=  \left[
    \begin{array}{ccc}
    \bm W_V^{\encircled{1}} & \ldots & \bm W_V^{\encircled{H}}
    \end{array}
    \right] \in\mathbb{R}^{d\times d}
    \text{,}
    \\
    &\bm b_V := \left[ 
    \begin{array}{ccc}
    \bm b_V^{\encircled{1}} & \ldots & \bm b_V^{\encircled{H}} 
    \end{array}
    \right] \in\mathbb{R}^{d}
    \text{,}
    \\
    &\bm W_O := \left[
    \begin{array}{c}
      \bm W_O^{\encircled{1}} \\
      \vdots \\
      \bm W_O^{\encircled{H}}
    \end{array}
    \right]
     \in\mathbb{R}^{d\times d}
    \text{.}
\end{align}

\subsubsection*{On the difference in arguments of ATTN and $f$}
In Section~\ref{subsec:mechanism}, we considered the scaling effect of ATTN, using the affine transformation $f$.
One may wonder about the difference between arguments of ATTN (i.e., $\bm x_i$) and arguments of $f$ (i.e., $\bm x_j$) in Equation~\ref{eq:ap:attn_decomposed}.
We can give two kinds of justification to this question.

In the estimation of the expansion rate, we consider the expected value.
From the symmetry of $\bm x_i $ and $\bm x_j $, when the expected value for $\bm x_j $ is obtained, the expected value for $\bm x_i $ is obtained.
In the actual BERT model, it has been empirically confirmed that two token vectors $\bm x_i, \bm x_j \in \bm X$ contained in the same context $\bm X$ exist in a fairly close position ($\bm x_i \approx \bm x_j$).
First, \citet{ethayarajh-2019-contextual} found that the cosine similarity between the intra-sentence representations in BERT is much larger than $0$.
Second, the norm of input vectors has just been unified by the layer normalization in the previous layer.
Thus, for our target models, $\bm x_i \approx \bm x_j$ is not a strong assumption.

\subsubsection*{Affine transformation as linear transformation}
\label{ap:subsec:affine_linear}
The affine transformation $f: \mathbb{R}^d \mapsto \mathbb{R}^d$ in ATTN can be viewed as a linear transformation $\widetilde{f}: \mathbb{R}^{d+1} \mapsto \mathbb{R}^{d+1}$.
Given $\widetilde{\bm x} := \left[ \begin{array}{cccc} & \bm x & & 1 \end{array} \right]\in\mathbb{R}^{d+1}$, where $1$ is concatenated to the end of each input vector $\bm x \in \mathbb{R}^d$, the affine transformation $f$ can be viewed as:
\begin{align}
&\widetilde{f}(\widetilde{\bm{x}}) = \widetilde{\bm{x}}\widetilde{\bm{W}}^V\widetilde{\bm{W}}^O \\
&\widetilde{\bm{W}}^V := \left[
\begin{array}{cccc}
  & & & 0 \\
  & \bm{W}^V & & \vdots \\
  & & & 0 \\
  & \bm{b}^V & & 1
\end{array}
\right]\in\mathbb{R}^{(d+1)\times(d+1)} \\
&\widetilde{\bm{W}}^O := \left[
\begin{array}{cccc}
  & & & 0 \\
  & \bm{W}^O & & \vdots \\
  & & & 0 \\
  0 & \ldots & 0 & 1
\end{array}
\right]\in\mathbb{R}^{(d+1)\times(d+1)}
\text{.}
\end{align}
The ``affine transformation'' mentioned in Section~\ref{subsec:mechanism} represent this linear transformation $\widetilde{f}$, and we measured the singular values of $\widetilde{f}$.

\subsection{Expected expansion rate for a random vector}
\label{ap:subsec:expected_exp_rate}
In the following, we introduce the derivation of the expansion rate of the affine transformation $f$, that is,
\begin{align}
    \frac{\norm{f(\mathbf x)}}{\norm{\mathbf x}}
    \approx \frac{\sqrt{\sum_{k=1}^d \sigma_k^2}}{\sqrt{d}}
    \text{.}
\end{align}
We assume that the input vector $\mathbf x$ is a sample from the standard normal distribution: 
\begin{align}
    \mathbf x = (\mathrm x_1,\dots,\mathrm x_d) \sim \mathcal N(\bm 0, I_d)
    \text{.}
\end{align}

First, the expectation value of  $\norm{\mathbf x}^2$ is as follows~\cite{vershynin_2018}:
\begin{align}
    \underset{\mathbf x}{\mathbb E} \norm{\mathbf x}^2
    =
    \underset{\mathbf x}{\mathbb E} \sum_{k=1}^d \mathrm x_k^2
    = \sum_{k=1}^d \underset{\mathbf x}{\mathbb E} \mathrm x_k^2
    = d
    \text{.}
\end{align}
Then, we have $\norm{\mathbf x} \approx \sqrt d$.

Next,
let the singular value decomposition of the linear transformation $f$ is $f = U\Sigma V^\top$, where $\Sigma = \mathrm{diag}(\sigma_1,\dots,\sigma_d)\in\mathbb R^{d\times d}$ is the diagonal matrix of singlar values of $f$.
As the matrix $V$ is orthogonal,
the following random vecotr $\mathbf f$ also follows the standard normal distribution, as does $\mathbf x$:
\begin{align}
    \mathbf y = (\mathrm y_1,\dots,\mathrm y_d) := V^\top \mathbf x \sim \mathcal N(\bm 0, I_d)
    \text{.}
\end{align}
By the orthogonal transformation by $U$ does not change the norm,
we need to estimate $\norm{\Sigma \mathbf y}^2$ in order to estimate $\norm{f(\mathbf x)}^2 = \norm{U\Sigma V^\top \mathbf x}^2$.
\begin{align}
    \underset{\mathbf y}{\mathbb E} \norm{\Sigma \mathbf y}^2
    &= \underset{\mathbf y}{\mathbb E} \sum_{k=1}^d \sigma_k^2 \mathrm y_k^2
    = \sum_{k=1}^d \sigma_k^2 \underset{\mathbf x}{\mathbb E} \mathrm y_k^2
    \\
    &= \sum_{k=1}^d \sigma_k^2
    \text{.}
\end{align}
Then, we have $\norm{f(\mathbf x)} \approx \sqrt{\sum_{k=1}^d \sigma_k^2}$.

To summarize,
\begin{align}
    \frac{\norm{f(\mathbf x)}}{\norm{\mathbf x}}
    \approx \frac{\sqrt{\sum_{k=1}^d \sigma_k^2}}{\sqrt{d}}
    \text{.}
\end{align}

\subsection{Results for other models}
\label{ap:subsec:singular_results_models}

Table~\ref{table:singular_values} shows the expected expansion rate of $f$ for each model.

\section{Relationship between word frequency and mixing ratio in other settings}
\label{ap:relation_with_freq_variant}
We also conducted the experiment shown in Section~\ref{subsec:bert_exp:freq} with the pre-trained BERT-large, BERT-medium, BERT-small, BERT-mini, and BERT-tiny.
However, we didn't do for RoBERTa-large and RoBERTa-base due to the difficulty of reproducing the pre-training dataset to count the word frequency.
Table~\ref{table:relation_with_freq_berts} lists the Spearman's rank correlation $\rho$ between the frequency rank and the mixing ratio for each model.
We discussed the inconsistency of the results across different model sizes in Section~\ref{subsec:bert_exp:freq}.

We also conducted it with the other three datasets.
Table~\ref{table:relation_with_freq_datasets} lists the Spearman's rank correlation $\rho$ between the frequency rank and the mixing ratio for each dataset.

Furthermore, we conducted with 25 BERT-base models trained with different seeds.
Table~\ref{table:relation_with_freq_seeds} lists the Spearman's rank correlation $\rho$ between the frequency rank and the mixing ratio for the models on the Wikipedia dataset.

\begin{table}[t]
\centering
\setlength{\tabcolsep}{2pt} 
\renewcommand{\arraystretch}{1.0}
{\small
\begin{tabular}{lccc}
\toprule
Models & Hidden dim. & \#Layer & \#Head \\
\cmidrule(r){1-1} \cmidrule(lr){2-2} \cmidrule(lr){3-3} \cmidrule(l){4-4}
BERT-large & 1026 & 24 & 16 \\
BERT-base & 768 & 12 & 12 \\
BERT-medium & 512 & 8 & 8 \\
BERT-small & 512 & 4 & 8 \\
BERT-mini & 256 & 4 & 4 \\
BERT-tiny & 128 & 2 & 2 \\
RoBERTa-large & 1026 & 24 & 16 \\
RoBERTa-base & 768 & 12 & 12 \\
\bottomrule
\end{tabular}
}
\caption{
Architecture hyperparameters of each model.
}
\label{table:bert_size}
\end{table}

\begin{table}[t]
\centering
\setlength{\tabcolsep}{4pt} 
\renewcommand{\arraystretch}{1.0}
    {\small
\begin{tabular}{lccc}
\toprule
\multicolumn{1}{c}{Methods} & Mean & Max & Min \\
\cmidrule(r){1-1} \cmidrule(lr){2-2} \cmidrule(lr){3-3} \cmidrule(l){4-4}
\multicolumn{1}{c}{--- BERT-large ---} \\
\textsc{Attn-w} & 97.4 & 100.0 & 15.0 \\
\textsc{Attn-n}  &  87.0  &  100.0 &  5.6 \\
\textsc{AttnRes-w} & 48.7 &  50.0 &  7.5     \\
\textsc{AttnRes-n} & 19.1 & 87.4 & 1.8 \\
\textsc{AttnResLn-n} & 14.9 & 86.6 & 1.6 \\
\multicolumn{1}{c}{--- BERT-base ---} \\
\textsc{Attn-w} & 97.1 & 100.0 & 45.0 \\
\textsc{Attn-n}  &  85.2  &  100.0 &  4.9 \\ 
\textsc{AttnRes-w}  &  48.6  &  50.0 &  22.5 \\ 
\textsc{AttnRes-n} &  22.3 & 65.7 & 2.0 \\
\textsc{AttnResLn-n} &  18.8 & 61.3 & 1.3 \\
\multicolumn{1}{c}{--- BERT-medium ---} \\
\textsc{Attn-w} & 95.6 & 100.0 & 49.5 \\
\textsc{Attn-n}  &  83.4  &  99.9 &  9.7 \\
\textsc{AttnRes-w} & 47.8 &  50.0 &  24.8     \\
\textsc{AttnRes-n} & 20.9 & 49.2 & 3.8 \\
\textsc{AttnResLn-n} & 18.7 & 65.2 & 1.2 \\
\multicolumn{1}{c}{--- BERT-small ---} \\
\textsc{Attn-w} & 96.2 & 100.0 & 57.7 \\
\textsc{Attn-n}  &  85.3  &  100.0 &  10.3 \\
\textsc{AttnRes-w} & 48.1 &  50.0 &  28.9     \\
\textsc{AttnRes-n} & 29.6 & 80.4 & 6.7 \\
\textsc{AttnResLn-n} & 27.2 & 85.5 & 7.3 \\
\multicolumn{1}{c}{--- BERT-mini ---} \\
\textsc{Attn-w} & 95.5 & 100.0 & 50.9 \\
\textsc{Attn-n}  &  85.7  &  100.0 &  10.4 \\
\textsc{AttnRes-w} & 47.8 &  50.0 &  25.4     \\
\textsc{AttnRes-n} & 27.2 & 68.1 & 7.3 \\
\textsc{AttnResLn-n} & 26.4 & 70.7 & 6.6 \\
\multicolumn{1}{c}{--- BERT-tiny ---} \\
\textsc{Attn-w} & 94.1 & 99.9 & 38.6 \\
\textsc{Attn-n}  &  90.4  &  99.9 &  28.3 \\
\textsc{AttnRes-w} & 47.1 &  50.0 &  19.3     \\
\textsc{AttnRes-n} &  37.8 & 77.9 & 18.1 \\
\textsc{AttnResLn-n} &  37.3 & 70.4 & 17.6 \\
\multicolumn{1}{c}{--- RoBERTa-large ---} \\
\textsc{Attn-w} & 96.7 & 100.0 & 10.1 \\
\textsc{Attn-n}  &  87.8  &  99.9 &  15.2 \\
\textsc{AttnRes-w} & 48.4 &  50.0 &  5.0     \\
\textsc{AttnRes-n} &  19.8 & 87.8 & 4.3 \\
\textsc{AttnResLn} &  19.7 & 87.9 & 4.3 \\
\multicolumn{1}{c}{--- RoBERTa-base ---} \\
\textsc{Attn-w} & 95.8 & 100.0 & 3.8 \\
\textsc{Attn-n}  &  84.4  &  100.0 &  13.8 \\ 
\textsc{AttnRes-w}  &  47.9  &  50.0 &  1.9 \\ 
\textsc{AttnRes-n} &  19.6 & 69.9 & 1.8 \\
\textsc{AttnResLn-n} &  16.2 & 73.4 & 1.5 \\
\bottomrule
\end{tabular}
}
\caption{Mean, maximum, and minimum values of the mixing ratio in seven variants of the masked language models, computed with each method.
}
\label{table:overall_mixing_ratio_models}
\end{table}
\begin{table}[t]
\centering
\setlength{\tabcolsep}{4pt} 
\renewcommand{\arraystretch}{0.85}
    {\small
\begin{tabular}{lrrr}
\toprule
\multicolumn{1}{c}{Methods} & Mean & Max & Min \\
\cmidrule(r){1-1} \cmidrule(lr){2-2} \cmidrule(lr){3-3} \cmidrule(l){4-4}
\multicolumn{1}{c}{--- Wikipedia ---} \\
\textsc{Attn-w} & 97.1 & 100.0 & 45.0 \\
\textsc{Attn-n}  &  85.2  &  100.0 &  4.9 \\ 
\textsc{AttnRes-w}  &  48.6  &  50.0 &  22.5 \\ 
\textsc{AttnRes-n} &  22.3 & 65.7 & 2.0 \\
\textsc{AttnResLn-n} &  18.8 & 61.3 & 1.3 \\
\multicolumn{1}{c}{--- SST-2 ---} \\
\textsc{Attn-w} & 92.5 & 100.0 & 2.2 \\
\textsc{Attn-n}  &  80.3  &  99.8 &  6.7 \\ 
\textsc{AttnRes-w}  &  46.3  &  50.0 &  1.1 \\ 
\textsc{AttnRes-n} &  22.5 & 50.4 & 2.4 \\
\textsc{AttnResLn-n} &  18.5 & 44.9 & 1.1 \\
\multicolumn{1}{c}{--- MNLI ---} \\
\textsc{Attn-w} & 94.6 & 100.0 & 10.0 \\
\textsc{Attn-n}  &  83.5  &  99.9 &  6.8 \\ 
\textsc{AttnRes-w}  &  47.3  &  50.0 &  5.0 \\ 
\textsc{AttnRes-n} &  22.4 & 65.4 & 2.8 \\
\textsc{AttnResLn-n} &  18.3 & 60.7 & 1.2 \\
\multicolumn{1}{c}{--- CoNLL'03 NER ---} \\
\textsc{Attn-w} & 91.7 & 100.0 & 1.5 \\
\textsc{Attn-n}  &  79.0  &  99.9 &  7.0 \\ 
\textsc{AttnRes-w}  &  45.8  &  50.0 &  0.8 \\ 
\textsc{AttnRes-n} &  22.4 & 51.5 & 2.7 \\
\textsc{AttnResLn-n} &  18.6 & 45.8 & 0.8 \\

\bottomrule
\end{tabular}
}
\caption{Mean, maximum, and minimum values of the mixing ratio in each method for BERT-base on each data.
}
\label{table:overall_mixing_ratio_data}
\end{table}
\begin{table}[t]
\centering
\setlength{\tabcolsep}{4pt} 
\renewcommand{\arraystretch}{0.85}
    {\small
\begin{tabular}{lrrrr}
\toprule
\multicolumn{1}{c}{Methods} & Mean (SD) & Max & Min \\
\cmidrule(r){1-1} \cmidrule(lr){2-2} \cmidrule(lr){3-3} \cmidrule(l){4-4}
\textsc{Attn-w} & 96.1 (0.1) & 100.0 & 8.8 \\
\textsc{Attn-n}  &  85.2 (0.4) &  100.0 &  7.4 \\ 
\textsc{AttnRes-w}  &  48.1 (0.1)  &  50.0 &  4.4 \\ 
\textsc{AttnRes-n} &  21.9 (0.3) & 64.6 & 3.4 \\
\textsc{AttnResLn-n} &  17.5 (0.4) & 67.7 & 1.4 \\
\bottomrule
\end{tabular}
}
\caption{Mean, maximum, and minimum values of the mixing ratio in each method for 25 BERT-base models trained with different random seeds by \citet{sellam2021multiberts}.
Mean value is the average of the values from 25 models, and the standard deviation (SD) is also listed.
Maximum and minimum values are the maximum and minimum of these values from 25 models, respectively.
}
\label{table:overall_mixing_ratio_seeds}
\end{table}

\begin{table}[t]
\centering
\setlength{\tabcolsep}{4pt} 
\renewcommand{\arraystretch}{1.0}
    {\small
\begin{tabular}{lccc}
\toprule
\multicolumn{1}{c}{Models} & Mean (SD) & Min & Max \\
\cmidrule(r){1-1} \cmidrule(lr){2-2} \cmidrule(lr){3-3} \cmidrule(l){4-4}
BERT-large & 0.80 & 0.61 & 1.08 \\
BERT-base & 0.88  & 0.63 & 1.05 \\
MultiBERTs (base) & 0.88 (0.01) & 0.65 & 1.43 \\
RoBERTa-large & 0.94 & 0.67 & 1.09 \\
\hline
BERT-medium & 0.87 & 0.60 & 1.41 \\
BERT-small & 1.31 & 0.78 & 2.41 \\
BERT-mini & 1.24 & 0.65 & 2.48 \\
BERT-tiny & 1.86 & 1.63 & 2.09 \\
RoBERTa-base & 1.30 & 1.10 & 1.49 \\
\bottomrule
\end{tabular}
}
\caption{Mean, maximum, and minimum values of the scaling magnification in each layer for nine variants of the masked language models.
In the ``MultiBERTs (base)'', results for 25 BERT-base models trained with different random seeds by \citet{sellam2021multiberts} are reported.
Mean value is the average of the values from 25 models, and the standard deviation (SD) is also listed.
Maximum and minimum values are the maximum and minimum of these values from 25 models, respectively.
}
\label{table:singular_values}
\end{table}

\begin{table}[t]
\centering
\setlength{\tabcolsep}{2pt} 
\renewcommand{\arraystretch}{1.0}
{\small
\begin{tabular}{lrr}
\toprule
\multicolumn{1}{c}{\multirow{2}{*}{Methods}} & \multicolumn{2}{c}{Spearman's $\rho$}     \\
\multicolumn{1}{c}{}                         & all tokens & w/o special tokens \\ 
\cmidrule(r){1-1} \cmidrule(lr){2-2} \cmidrule(l){3-3}
\multicolumn{1}{c}{--- BERT-large ---} \\
\textsc{Attn-w} & $0.44$      & $0.44$   \\
\textsc{Attn-n}  &  $-0.53$  &  $-0.56$  \\ 
\textsc{AttnRes-w}  &  $0.44$  &  $0.44$  \\ 
\textsc{AttnRes-n}         & $-0.83$       & $-0.84$   \\
\textsc{AttnResLn-n}      & $-0.71$       & $-0.75$    \\ 
\multicolumn{1}{c}{--- BERT-base ---} \\
\textsc{Attn-w} & $0.16$      & $0.14$   \\
\textsc{Attn-n}  &  $-0.39$  &  $-0.41$  \\ 
\textsc{AttnRes-w}  &  $0.16$  &  $0.14$  \\ 
\textsc{AttnRes-n}         & $-0.84$       & $-0.86$   \\
\textsc{AttnResLn-n}      & $-0.54$       & $-0.58$    \\
\multicolumn{1}{c}{--- BERT-medium ---} \\
\textsc{Attn-w} & $-0.09$      & $-0.11$   \\
\textsc{Attn-n}  &  $-0.13$  &  $-0.14$  \\ 
\textsc{AttnRes-w}  &  $-0.09$  &  $-0.11$  \\ 
\textsc{AttnRes-n}         & $-0.41$       & $-0.43$   \\
\textsc{AttnResLn-n}      & $-0.02$       & $-0.03$    \\ 
\multicolumn{1}{c}{--- BERT-small ---} \\
\textsc{Attn-w} & $-0.05$      & $-0.07$   \\
\textsc{Attn-n}  &  $0.26$  &  $0.26$  \\ 
\textsc{AttnRes-w}  &  $-0.05$  &  $-0.07$  \\ 
\textsc{AttnRes-n}         & $-0.22$       & $-0.20$   \\
\textsc{AttnResLn-n}      & $0.19$       & $0.21$    \\ 
\multicolumn{1}{c}{--- BERT-mini ---} \\
\textsc{Attn-w} & $-0.52$      & $-0.55$   \\
\textsc{Attn-n}  &  $-0.15$  &  $-0.17$  \\ 
\textsc{AttnRes-w}  &  $-0.52$  &  $-0.55$  \\ 
\textsc{AttnRes-n}         & $0.23$       & $0.25$   \\
\textsc{AttnResLn-n}      & $0.42$       & $0.44$    \\ 
\multicolumn{1}{c}{--- BERT-tiny ---} \\
\textsc{Attn-w} & $-0.75$      & $-0.77$   \\
\textsc{Attn-n}  &  $-0.62$  &  $-0.64$  \\ 
\textsc{AttnRes-w}  &  $-0.75$  &  $-0.77$  \\ 
\textsc{AttnRes-n}         & $0.26$       & $0.27$   \\
\textsc{AttnResLn-n}      & $0.24$       & $0.25$    \\ 
\bottomrule
\end{tabular}
}
\caption{
The Spearman's $\rho$ between the frequency rank and the mixing ratio calculated by each method for five variants of pre-trained BERT.
In the ``w/o special tokens'' setting, it was calculated without \texttt{[CLS]} and \texttt{[SEP]}.
}
\label{table:relation_with_freq_berts}
\end{table}

\begin{table}[t]
\centering
\setlength{\tabcolsep}{2pt} 
\renewcommand{\arraystretch}{1.0}
{\small
\begin{tabular}{lrr}
\toprule
\multicolumn{1}{c}{\multirow{2}{*}{Methods}} & \multicolumn{2}{c}{Spearman's $\rho$}     \\
\multicolumn{1}{c}{}                         & all tokens & w/o special tokens \\ 
\cmidrule(r){1-1} \cmidrule(lr){2-2} \cmidrule(l){3-3}
\multicolumn{1}{c}{--- Wikipedia ---} \\
\textsc{Attn-w} & $0.16$      & $0.14$   \\
\textsc{Attn-n}  &  $-0.39$  &  $-0.41$  \\ 
\textsc{AttnRes-w}  &  $0.16$  &  $0.14$  \\ 
\textsc{AttnRes-n}         & $-0.84$       & $-0.86$   \\
\textsc{AttnResLn-n}      & $-0.54$       & $-0.58$    \\ 
\multicolumn{1}{c}{--- SST-2 ---} \\
\textsc{Attn-w} & $0.22$      & $0.19$   \\
\textsc{Attn-n}  &  $-0.24$  &  $-0.33$  \\ 
\textsc{AttnRes-w}  &  $0.22$  &  $0.19$  \\ 
\textsc{AttnRes-n}         & $-0.81$       & $-0.84$   \\
\textsc{AttnResLn-n}      & $-0.42$       & $-0.54$    \\
\multicolumn{1}{c}{--- MNLI ---} \\
\textsc{Attn-w} & $0.22$      & $0.19$   \\
\textsc{Attn-n}  &  $-0.31$  &  $-0.40$  \\ 
\textsc{AttnRes-w}  &  $0.22$  &  $0.19$  \\ 
\textsc{AttnRes-n}         & $-0.77$       & $-0.84$   \\
\textsc{AttnResLn-n}      & $-0.40$       & $-0.50$    \\ 
\multicolumn{1}{c}{--- NER ---} \\
\textsc{Attn-w} & $0.16$      & $0.09$   \\
\textsc{Attn-n}  &  $-0.22$  &  $-0.34$  \\ 
\textsc{AttnRes-w}  &  $0.16$  &  $0.09$  \\ 
\textsc{AttnRes-n}         & $-0.79$       & $-0.85$   \\
\textsc{AttnResLn-n}      & $-0.41$       & $-0.57$    \\ 
\bottomrule
\end{tabular}
}
\caption{
The Spearman's $\rho$ between the frequency rank and the mixing ratio calculated by each method for the four variants of datasets.
In the ``w/o special tokens'' setting, it was calculated without \texttt{[CLS]} and \texttt{[SEP]}.
}
\label{table:relation_with_freq_datasets}
\end{table}

\begin{table}[t]
\centering
\setlength{\tabcolsep}{2pt} 
\renewcommand{\arraystretch}{1.0}
{\small
\begin{tabular}{lrr}
\toprule
\multicolumn{1}{c}{\multirow{2}{*}{Methods}} & \multicolumn{2}{c}{Spearman's $\rho$}     \\
\multicolumn{1}{c}{}                         & all tokens (SD) & w/o special tokens (SD) \\ 
\cmidrule(r){1-1} \cmidrule(lr){2-2} \cmidrule(l){3-3}
\textsc{Attn-w} & $0.35$ ($0.01$)      & $0.35$ ($0.07$)   \\
\textsc{Attn-n}  &  $-0.23$ ($0.01$)  &  $-0.25$ ($0.09$)  \\ 
\textsc{AttnRes-w}  &  $0.35$ ($0.01$)  &  $0.35$ ($0.07$)  \\ 
\textsc{AttnRes-n}         & $-0.79$ ($0.02$)       & $-0.80$ ($0.02$)   \\
\textsc{AttnResLn-n}      & $-0.36$ ($0.10$)       & $-0.38$ ($0.11$)    \\ 
\bottomrule
\end{tabular}
}
\caption{
Spearman's $\rho$ between the frequency rank and the mixing ratio calculated by each method for for 25 BERT-base models trained with different random seeds.
In the ``w/o special tokens'' setting, it was calculated without \texttt{[CLS]} and \texttt{[SEP]}.
Both of the values are the mean of the values from 25 models, and the standard deviation (SD) is also listed.
}
\label{table:relation_with_freq_seeds}
\end{table}

\end{document}